%% file: TPAMI.tex
\documentclass[10pt,journal,twoside,compsoc]{IEEEtran}

\ifCLASSOPTIONcompsoc
  \usepackage[nocompress]{cite}
\else
  \usepackage{cite}
\fi

\usepackage{amsmath,amsfonts,amssymb,bm}
\usepackage{eqnarray}
\usepackage{xr,verbatim,color}
\usepackage{graphicx}

\usepackage{algorithm,algorithmic}

\ifCLASSOPTIONcompsoc
  \usepackage[caption=false,font=footnotesize,labelfont=sf,textfont=sf]{subfig}
\else
  \usepackage[caption=false,font=footnotesize]{subfig}
\fi

\newtheorem{representer}{Representer Theorem}
\newtheorem{lemma}{Lemma}
\newtheorem{definition}{Definition}

\begin{document}
\title{Efficient Training for Positive Unlabeled Learning}

\author{Emanuele~Sansone,
        Francesco~G.~B.~De~Natale,~\IEEEmembership{Senior~Member,~IEEE}
        and~Zhi-Hua~Zhou,~\IEEEmembership{Fellow,~IEEE}
\IEEEcompsocitemizethanks{\IEEEcompsocthanksitem Emanuele Sansone and Francesco De Natale are with the Department of Information Engineering and
Computer Science (DISI), University of Trento, Trento 38123, Italy.\protect\\
E-mail: e.sansone@unitn.it
\IEEEcompsocthanksitem Zhi-Hua Zhou is with the National Key Laboratory for Novel Software Technology, Nanjing University, Nanjing 210023, China.}}


\IEEEtitleabstractindextext{%
\begin{abstract}
Positive unlabeled (PU) learning is useful in various practical situations, where there is a need to learn a classifier for a class of interest from an unlabeled data set, which may contain anomalies as well as samples from unknown classes. The learning task can be formulated as an optimization problem under the framework of statistical learning theory. Recent studies have theoretically analyzed its properties and generalization performance, nevertheless, little effort has been made to consider the problem of scalability, especially when large sets of 
unlabeled data are available. In this work we propose a novel scalable PU learning algorithm that is theoretically proven to provide the optimal solution, while showing superior computational and memory performance. Experimental evaluation confirms the theoretical evidence and shows that the proposed method can be successfully applied to a large variety of real-world problems involving PU learning.
\end{abstract}

}

\maketitle

\IEEEdisplaynontitleabstractindextext

\IEEEpeerreviewmaketitle

\IEEEraisesectionheading{\section{Introduction}\label{sec:intro}}
\vspace{-0.3cm}
\input{introduction.tex}

\vspace{-0.5cm}
{\section{Literature Review}\label{sec:related}}
\vspace{-0.1cm}
\input{related.tex}

\vspace{-0.5cm}
{\section{PU Learning Formulation}\label{sec:pu}}
\input{puformulation.tex}

\vspace{-0.8cm}
{\section{USMO Algorithm}\label{sec:algorithm}}
\vspace{-0.3cm}
\input{algorithm.tex}

{\section{Theoretical Analysis}\label{sec:theory}}
\input{theory.tex}

\vspace{-0.4cm}
{\section{Experimental Results}\label{sec:results}}
\input{results.tex}

\vspace{-0.5cm}
\section{Conclusions}
In this work an efficient algorithm for PU learning is proposed. Theoretical analysis is 
provided to ensure that the obtained solution recovers the optimum of the objective function. 
Experimental validation confirms that the proposed solution
can be applied to real-world problems. 





\vspace{-0.4cm}
\ifCLASSOPTIONcompsoc
  \section*{Acknowledgments}
\else
  \section*{Acknowledgment}
\fi
This research was partially supported by NSFC (61333014) and the Collaborative Innovation Center of Novel Software Technology and Industrialization. Part of this research was conducted when E. Sansone was visiting the LAMDA Group, Nanjing University. We gratefully acknowledge the support of NVIDIA Corporation with the donation of a Titan X Pascal machine to support this research.

\ifCLASSOPTIONcaptionsoff
  \newpage
\fi

\vspace{-0.3cm}
\bibliographystyle{IEEEtran}
\bibliography{IEEEabrv,reference}

\end{document}

%% file: introduction.tex
\IEEEPARstart{P}{ositive} unlabeled (PU) learning refers to the task of learning a binary classifier from only positive and unlabeled data~\cite{elkan2008learning}. This classification problem arises in various practical situations, such as:
\begin{itemize}
\item Retrieval~\cite{onoda2005one}, where the goal is to find samples in an unlabeled data set similar to user-provided ones.
\item Inlier-based outlier detection~\cite{hido2008inlier}, where the goal is to detect outliers from an unlabeled data set, based on inlier samples.
\item One-vs-rest classification~\cite{li2011positive}, where the negative class is too diverse, thus being difficult to collect and label enough negative samples.
\item Open set recognition~\cite{scheirer2013toward}, where testing classes are unknown at training time and the exploitation of unlabeled data may help learning more robust concepts.
\end{itemize}

Naive approaches are proposed to address PU learning. In particular, it is possible to distinguish between solutions that heuristically identify reliable negative samples from unlabeled data and use them to train a standard binary classifier, and solutions based on binary classifiers using all unlabeled data as negative. The former are heavily dependent on heuristics, while the latter suffer the problem of wrong label assignment.

The recent works in~\cite{plessis2014analysis} and~\cite{plessis2015convex} formulate PU learning as an optimization problem under the framework of statistical learning theory~\cite{vapnik1999overview}. Both works theoretically analyse the problem, deriving generalization error bounds and studying the optimality of the obtained solutions. Even though these methods are theoretically grounded, they are not scalable. In this work we present a method that provides better scalability, while maintaining the optimality of the above approaches for what concerns the generalization. In particular, starting from the formulation of the convex optimization problem in~\cite{plessis2015convex}, we derive an algorithm that requires significantly lower memory and computation, while being proven to converge to the optimal solution.



The rest of the paper is organized as follows. Section~\ref{sec:related} reviews related works, starting from the comparison of PU learning with one-class classification and semi-supervised learning and describing the main theoretical results achieved in PU learning. Section~\ref{sec:pu} provides the formulation of the optimization problem under the framework of statistical learning theory and enunciates the representer theorem for PU learning. Section~\ref{sec:algorithm} and Section~\ref{sec:theory} describe USMO (Unlabeled data in Sequential Minimal Optimization) algorithm and prove its convergence, respectively. Section~\ref{sec:results} provides a comprehensive evaluation of the proposed algorithm on a large collection of real-world datasets. 

%% file: related.tex
PU learning is well known in the machine learning community, being used in a variety of tasks such as matrix completion~\cite{hsieh2015pu}, multi-view learning~\cite{zhou2012multi}, and semi-supervised learning~\cite{sakai2016beyond}. It is also applied in data mining to classify data streams~\cite{li2009positive} or time series~\cite{nguyen2011positive} and to detect events, like co-occurrences, in graphs~\cite{silva2009hypergraph}.

PU learning approaches can be classified in two broad categories, according to the use of unlabeled data: two-stage and single stage approaches. The former extract a set of reliable negative samples from unlabeled data and use them, together with the available positive data, to train a binary classifier~\cite{liu2002partially,yu2002pebl,li2003learning,yu2005single}. These first step is heuristic and strongly influences the final result. The latter regard \textbf{all unlabeled data as negative samples}. Positive and negative data are then used to train different classifiers based on SVM~\cite{liu2003building,elkan2008learning}, neural networks~\cite{skabar2003single} or kernel density estimators~\cite{blanchard2010semi}. These approaches suffer the problem of \textbf{wrong label assignment}, whose effect depends on the proportion of positive samples in the unlabeled dataset. We will see later, in the discussion about theoretical studies of PU learning, how critical this issue is. For the moment, we focus on analyzing the relations of PU learning with one-class classification (OCC) and semi-supervised learning, which allows us drawing some clear boundaries between these tasks and highlighting the novelty of this work.

\vspace{0.1cm}
\subsection{Comparison with one-class classification}
The main goal of OCC is to estimate the support of data distribution, which is extremely useful in unsupervised learning, especially in high-dimensional feature spaces, where it is very difficult to perform density estimation. OCC is applied to many real-world problems, such as anomaly/novelty detection (see~\cite{kittler2014domain} for a recent survey and definition of anomaly detection and ~\cite{markou2003novelty1,markou2003novelty2} for novelty detection). Other possible applications of OCC include author verification in text documents~\cite{koppel2004authorship}, document retrieval~\cite{onoda2005one}, and collaborative filtering in social networks~\cite{pan2008one}.

Authors is~\cite{scholkopf1999sv,tax1999support} are among the first to develop OCC algorithms.\footnote{More precisely, the term OCC was coined in 1996~\cite{moya1996network}.} In particular, ~\cite{scholkopf1999sv} proposes a classifier that finds the hyperplane separating data from the origin with the maximum margin, while ~\cite{tax1999support} proposes a classifier that finds the mimimum radius hypersphere enclosing data. Despite the difference between these two approaches, ~\cite{scholkopf1999sv} proves that, for translation-invariant kernels such as the Gaussian kernel, they obtain the same solutions. Extensions of these two pioneering works, falling in the category of kernel methods, are proposed few years later. ~\cite{scholkopf2000kernel} modifies the model of~\cite{scholkopf1999sv} by incorporating a small training set of anomalies and using the centroid of such set as the reference point to find the hyperplane. ~\cite{tax2001uniform} proposes a strategy to automatically select the hyperparameters defined in~\cite{tax1999support} to increase the usability of the framework. Rather than repelling samples from a specific point, as in~\cite{scholkopf1999sv,scholkopf2000kernel}, ~\cite{bennett2001linear} suggests a strategy to attract samples towards the centroid, solving a linear programming problem that minimizes the average output of the target function computed on the training samples. Authors in~\cite{pekalska2002one} propose a similar strategy based on linear programming, where data are represented in a similarity/dissimilarity space. The framework is well suited for OCC applications involving strings, graphs or shapes. Solutions different from kernel methods are also proposed. To mention a few, ~\cite{manevitz2001one} proposes a neural network-based approach, where the goal is to learn the identity function. New samples are fed into the network and tested against their corresponding outputs. The test sample is considered as part of the class of interest only when the input and the output of the network are similar. ~\cite{tax2001one} proposes a one-class nearest neighbour, where a test sample is accepted as a member of the target class only when the distance from its neighbours is comparable to their local density. \textbf{It is worth noting that most of the works in OCC focus on increasing classification performance, rather than improving scalability}. This is arguably motivated by the fact that it is usually difficult to collect large amounts of training samples for the concept/class of interest. Solutions to improve classification performance rely on classical strategies such as ensemble methods~\cite{tax2001combining}, bagging~\cite{shieh2009ensembles} or boosting~\cite{ratsch2002constructing}. Authors in~\cite{jumutc2014multi} argue that existing one-class classifiers fail when dealing with mixture distributions. Accordingly, they propose a multi-class classifier exploiting the supervised information of all classes of interest to refine support estimation.

A promising solution to improve OCC consists of exploiting unlabeled data, which are in general largely available. As discussed in~\cite{blanchard2010semi}, standard OCC algorithms are not designed to use unlabeled data, thus making the implicit assumption that they are uniformly distributed over the support of nominal distribution, which does not hold in general. The recent work in~\cite{niu2016theoretical} proves that, under some simple conditions,\footnote{The conditions are based on class prior and size of positive (class of interest) and negative (the rest) data.} large amounts of unlabeled data can boost OCC even compared to fully supervised approaches. Furthermore, unlabeled data allow building OCC classifiers in the context of open set recognition~\cite{scheirer2013toward}, where it is essential to learn robust concepts/functions. Since the primary goal of PU learning is to exploit this unsupervised information, \textbf{it can be regarded as a generalization of OCC}~\cite{khan2014one}, in the sense that it can manage unlabeled data coming from more general distributions than the uniform one.

\vspace{-0.4cm}
\subsection{Comparison with semi-supervised learning}
The idea of exploiting unlabeled data in semi-supervised learning was originally proposed in~\cite{shahshahani1994effect}. Earlier studies do not thoroughly explain why unlabeled data can be beneficial. Authors of~\cite{miller1997mixture} are among the first to analyze this aspect from a generative perspective. In particular, they assume that data are distributed according to a mixture of Gaussians and show that the class posterior distribution can be decomposed in two terms, depending one on the class labels and the other on the mixture components. The two terms can be estimated using labeled and unlabeled data, respectively, thus improving the performance of the learnt classifier. ~\cite{zhang2000value} extends this analysis assuming that data can be correctly described by a more general class of parametric models. The authors shows that if both the class posterior distribution and the data prior distribution are dependent on model parameters, unlabeled examples can be exploited to learn a better set of parameters. Thus, the key idea of semi-supervised learning is to exploit the distributional information contained in unlabeled samples.

Many approaches have been proposed. The work in~\cite{chapelle2009semi} provides a taxonomy of semi-supervised learning algorithms. In particular, it is possible to distinguish five types of approaches: generative approaches (see, e.g., ~\cite{sansone16classtering}), exploit unlabeled data to better estimate the class-conditional densities and infer the unknown labels based on the learnt model; low-density separation methods (see, e.g., ~\cite{chapelle2005semi}, look for decision boundaries that correctly classify labeled data and are placed in regions with few unlabeled samples (the so called low-density regions); 
graph-based methods (see, e.g., ~\cite{belkin2006manifold}), exploit unlabeled data to build a similarity graph and then propagate labels based on smoothness assumptions; 
dimensionality reduction methods (see, e.g., ~\cite{kingma2014semi}), use unlabeled samples for representation learning and then perform classification on the learnt feature representation; and disagreement-based methods (discussed in~\cite{zhou2010semi}), exploit the disagreement among multiple base learners to learn more robust ensemble classifiers.

The scalability issue is largely studied in the context of semi-supervised learning. For example, the work in~\cite{li2013convex} proposes a framework to solve a mixed-integer programming problem, which runs multiple times the SVM algorithm. State-of-art implementations of SVM (see, e.g., LIBSVM~\cite{chang2011libsvm}) are based mainly on decomposition methods~\cite{chen2006study}, like our proposed approach. Other semi-supervised approaches use approximations of the fitness function to simplify the optimization problem (see ~\cite{fergus2009semi,talwalkar2008large}).

Both semi-supervised and PU learning exploit unlabeled data to learn better classifiers. Nevertheless, substantial differences hold that make semi-supervised learning not applicable to PU learning tasks. An important aspect is that most semi-supervised learning algorithms assume that unlabeled data are originated from a set of known classes (close set environment), thus not coping with the presence of unknown classes in training and test sets (open set environment). To the best of our knowledge, only few works like (see ~\cite{da2014learning}) propose semi-supervised methods able to handle such situation. Another even more relevant aspect is that \textbf{semi-supervised learning cannot learn a classifier when only one known class is present}, since it requires at least two known classes to calculate the decision boundary. On the contrary, recent works show that it is possible to apply PU learning to solve semi-supervised learning tasks, even in the case of open set environment ~\cite{sakai2016beyond,niu2016theoretical}.

\vspace{-0.4cm}
\subsection{Theoretical studies about PU learning}
\vspace{-0.1cm}
Inspired by the seminal work ~\cite{kearns1998efficient} and by the first studies on OCC~\cite{tax1999support,scholkopf1999sv}, the work in ~\cite{de1999positive} and its successive extension ~\cite{letouzey2000learning} are the first to define and theoretically analyze the problem of PU learning. In particular, they propose a framework based on the statistical query model~\cite{kearns1998efficient} to theoretically assess the classification performance and to derive algorithms based on decision trees. The authors study the problem of learning functions characterized by monotonic conjuctions, which are particularly useful in text classification, where documents can be represented as binary vectors that model the presence/absence of words from an available dictionary. Instead of considering binary features, ~\cite{he2010naive} proposes a Naive-Bayes classifier for categorical features in noisy environments. The work assumes the attribute independence, which eases the estimate of class-conditional densities in high-dimensional spaces, but is limiting as compared to discriminative approaches, directly focusing on classification and not requiring density estimations~\cite{jordan2002discriminative}.

As already mentioned at the beginning of this section, PU learning studies can be roughly classified in two-stage and single stage approaches. The former are based on heuristics to select a set of reliable negative samples from the unlabeled data and are not theoretically grounded, while the latter are subject to the problem of wrong label assignment. In order to understand how this issue is critical, let consider the theoretical result of consistency presented in~\cite{blanchard2010semi}.\footnote{It is rewritten to be more consistent with the notation in this paper.} For any set of classifiers $\mathcal{F}$ of Vapnik-Chervonenkis (VC) dimension $V$ and any $\delta>0$, there exists a constant $c$ such that the following bounds hold with probability $1-\delta$:
\vspace{-0.2cm}
{\small\begin{align}
&FNR(f)-FNR(f^*)\leq c\epsilon_n,\nonumber\\
&FPR(f)-FPR(f^*)\leq \frac{c}{1-\pi}(\epsilon_n+\epsilon_p)\nonumber,
\end{align}}
\vspace{-0.1cm}
where $FPR, FNR$ are the false positive/negative rates, $f\in\mathcal{F}$ is the function obtained by the above-mentioned strategy, $f^*\in\mathcal{F}$ is the optimal function having access to the ground truth, $\pi$ is the positive class prior, $\epsilon_{\cdot}=\sqrt[]{\frac{V\log(\cdot)-\log(\delta)}{\cdot}}$ and $p$ and $n$ are the number of positive and unlabeled samples, respectively. In particular, if one considers a simple scenario where the feature space is $\mathbb{R}^{100}$  and $V=101$ (in the case of linear classifiers), it is possible to learn a classifier such that, with probability of $90\%$, the performance does not deviate from the optimal values for more than $10\%$ (which is equivalent to setting $\delta,\epsilon_p,\epsilon_n=0.1$). This is guaranteed when both positive and unlabeled sets consist of at least $10^5$ training samples each. This is impractical in real world applications, since collecting and labelling so many data is very expensive. The effect of wrong label assignment is even more evident for larger values of positive class prior.

Recently, ~\cite{plessis2014analysis,plessis2015convex} proposed two frameworks based on the statistical learning theory~\cite{vapnik1999overview}. These works are free from heuristics, do not suffer the problem of wrong label assignment, and provide theoretical bounds on the generalization error. Another theoretical work is the one in~\cite{hsieh2015pu}, which specifically addresses the matrix completion problem, motivated by applications like recovering friendship relations in social networks based on few observed connections. This work however is unable to deal with the more general problem of binary PU learning, and formulates the optimization problem using the squared loss, which is known to be subobtimal according to the theoretical findings in ~\cite{plessis2014analysis,plessis2015convex}.\footnote{In fact, the best convex choice is the double Hinge loss, see Section~\ref{sec:pu} for further details.} Overall, the analysis of the literature in the field makes evident the lack of \textbf{theoretically-grounded PU learning approaches with good scalability properties}.

%% file: puformulation.tex
Assume that we are given a training dataset $D_b=\{(\mathbf{x}_i,y_i):\mathbf{x}_i\in X,y_i\in Y\}_{i=1}^{m}$, where $X\subseteq\mathbb{R}^d$, $Y=\{-1,1\}$ and each pair of samples in $D_b$ is drawn independently from the same unknown joint distribution $\mathcal{P}$ defined over $X$ and $Y$. The goal is to learn a function $f$ that maps the input space $X$ into the class set $Y$, known as the binary classification problem. According to statistical learning theory~\cite{vapnik1999overview}, the function $f$ can be learnt by minimizing the risk functional $\mathcal{R}$, namely
\vspace{-0.3cm}
{\small\begin{align}
\label{eq:birisk}
\mathcal{R}(f)=&\sum_{y\in Y}\int \ell(f(\mathbf{x}),y)\mathcal{P}(\mathbf{x},y)d\mathbf{x}\nonumber\\
=&\pi\int\ell(f(\mathbf{x}),1)\mathcal{P}(\mathbf{x}|y=1)d\mathbf{x}\nonumber\\
&{+}\:(1-\pi)\int\ell(f(\mathbf{x}),-1)\mathcal{P}(\mathbf{x}|y=-1)d\mathbf{x}
\end{align}}
where $\pi$ is the positive class prior and $\ell$ is a loss function measuring the disagreement between the prediction and the ground truth for sample $x$, viz. $f(\mathbf{x})$ and $y$, respectively.

In PU learning, the training set is split into two parts: a set of samples $D_p=\{\mathbf{x}_i\in X\}_{i=1}^p$ drawn from the positive class and a set of unlabeled samples $D_n=\{\mathbf{x}_i\in X\}_{i=1}^n$ drawn from both positive and negative classes. The goal is the same of the binary classification problem, but this time the supervised information is available only for one class. The learning problem can be still formulated as a risk minimization. In fact, since $\mathcal{P}(\mathbf{x})=\pi\mathcal{P}(\mathbf{x}|y=1)+(1-\pi)\mathcal{P}(\mathbf{x}|y=-1)$, (\ref{eq:birisk}) can be rewritten in the following way:
\vspace{-0.2cm}
{\small\begin{align}
\label{eq:purisk}
\mathcal{R}(f){=}&{\pi}{\int}\ell(f(\mathbf{x}),1)\mathcal{P}(\mathbf{x}|y=1)d\mathbf{x}\nonumber\\
&{+}\:{(1-\pi)}{\int}\ell(f(\mathbf{x}),-1)\frac{\mathcal{P}(\mathbf{x})-\pi\mathcal{P}(\mathbf{x}|y=1)}{1-\pi}d\mathbf{x}\nonumber\\
=&\:{\pi}{\int}\tilde{\ell}(f(\mathbf{x}),1)\mathcal{P}(\mathbf{x}|y=1)d\mathbf{x}{+}{\int}\ell(f(\mathbf{x}),-1)\mathcal{P}(\mathbf{x})d\mathbf{x}
\end{align}}
\vspace{-0.1cm}
where $\tilde{\ell}(f(\mathbf{x}),1)=\ell(f(\mathbf{x}),1)-\ell(f(\mathbf{x}),-1)$ is called the \textbf{composite loss}~\cite{plessis2015convex}.

The risk functional in~(\ref{eq:purisk}) cannot be minimized since the distributions are unknown. In practice, one considers the empirical risk functional in place of~(\ref{eq:purisk}), where expectation integrals are replaced with the empirical mean estimates computed over the available training data, namely
\vspace{-0.1cm}
{\small\begin{equation}
\label{eq:emprisk}
\mathcal{R}_{emp}(f)=\frac{\pi}{p}\sum_{\mathbf{x}_i\in D_p}\tilde{\ell}(f(\mathbf{x}_i),1)+\frac{1}{n}\sum_{\mathbf{x}_i\in D_n}\ell(f(\mathbf{x}_i),-1)
\end{equation}}
The minimization of $\mathcal{R}_{emp}$ is in general an ill-posed problem. A regularization term is usually added to $\mathcal{R}_{emp}$ to restrict the solution space and to penalize complex solutions. The learning problem is then stated as an optimization task:
\vspace{-0.2cm}
{\small\begin{equation}
\label{eq:puopt}
f^*=\arg \min_{f\in\mathcal{H}_k} \bigg\{\mathcal{R}_{emp}(f)+\lambda\|f\|_{\mathcal{H}_k}^2\bigg\}
\end{equation}}
where $\lambda$ is a positive real parameter weighting the relative importance of the regularizer with respect to the empirical risk  and $\|\cdot\|_{\mathcal{H}_k}$ is the norm associated with the function space $\mathcal{H}_k$. In this case, $\mathcal{H}_k$ refers to the Reproducing Kernel Hilbert Space (RKHS) associated with its Mercer kernel {\small$k:X\times X\rightarrow\mathbb{R}$}.\footnote{For an overview of RKHS and their properties, see the work in \cite{aronszajn1950theory}} We can then enunciate the representer theorem for PU learning (proof in 
Supplementary Material):
\begin{representer}
Given the training set $D=D_p\cup D_n$ and the Mercer kernel $k$ associated with the RKHS $\mathcal{H}_k$, any minimizer $f^*\in\mathcal{H}_k$ of~(\ref{eq:puopt}) admits the following representation \vspace{-0.2cm}{\small$$f^*(\mathbf{x})=\sum_{i:\mathbf{x}_i\in D}\alpha_i k(\mathbf{x},\mathbf{x}_i)$$}\vspace{-0.3cm} where $\alpha_i\in\mathbb{R}$ for all $i$.
\end{representer}
It is worth mentioning that many types of representer theorem have been proposed, but none of them can be applied to the PU learning problem. Just to mention a few, ~\cite{scholkopf2001generalized} provides a generalized version of the classical representer theorem for classification and regression tasks, while ~\cite{belkin2006manifold} derives the representer theorem for semi-supervised learning. Recently, ~\cite{argyriou2014unifying} proposed a unified view of existing representer theorems, identifying the relations between these theorems and certain classes of regularization penalties. Nevertheless, the proposed theory does not apply to ~(\ref{eq:puopt}) due to the hypotheses made on the empirical risk functional.

This theorem shows that it is possible to learn functions defined on possibly-infinite dimensional spaces, i.e., the ones induced by the kernel $k$, but depending only on a finite number of parameters $\alpha_i$. Thus, training focuses on learning such restricted set of parameters. Another important aspect is that the representer theorem does not say anything about the uniqueness of the solution, as it only says that every minimum solution has the same parametric form. In other words, different solutions have different values of parameters. The uniqueness of the solution is guaranteed only when the empirical risk functional in~(\ref{eq:puopt}) is convex. A proper selection of the loss function is then necessary to fulfill this condition. Authors in~\cite{plessis2015convex} analysed the properties of loss functions for the PU learning problem, showing that a necessary condition for convexity is that the composite loss function in~(\ref{eq:emprisk}) is affine. This requirement is satisfied by some loss functions, such as the squared loss, the modified Huber loss, the logistic loss, the perceptron loss, and the double Hinge loss. The latter ensures the best generalization performance~\cite{plessis2015convex}.\footnote{Double Hinge loss can be considered as the equivalent of Hinge loss function for the binary classification problem.} Even, the comparison with non-convex loss functions~\cite{plessis2014analysis,plessis2015convex} suggests to use the double Hinge loss for the PU learning problem, with a twofold advantage: ensuring that the obtained solution is globally optimal, and allowing the use of convex optimization theory to perform a more efficient training.

These considerations, together with the result stated by the representer theorem, allow us rewriting problem~(\ref{eq:puopt}) in an equivalent parametric form. In particular, defining $\bm{\alpha}\in\mathbb{R}^{(p+n)}$ as the vector of alpha values, $\bm{\xi}\in\mathbb{R}^n$ as the vector of slack variables, $\mathbf{K}\in\mathbb{R}^{(p+n)\times(p+n)}$ as the Gram matrix computed using the training set $D$, and considering, without loss of generality, target functions in the form $f(\mathbf{x})=\sum_i\alpha_i k(\mathbf{x},\mathbf{x}_i)+\beta$, where $\beta$ is the bias of $f$, it is possible to derive the following optimization problem (derivation in 
Supplementary Material):
\vspace{-0.2cm}
{\small\begin{align}
\label{eq:doubleprimal}
\min_{\bm{\alpha},\bm{\xi},\beta}&\Big\{{-}c_1\tilde{\mathbf{1}}^T\mathbf{K}\bm{\alpha}-c_1\tilde{\mathbf{1}}^T\mathbf{1}\beta+c_2\mathbf{1}_n^T\mathbf{\xi}+\frac{1}{2}\bm{\alpha}^T\mathbf{K}\bm{\alpha}\Big\}\nonumber\\
\text{s.t.}&\;\bm{\xi}\succeq\mathbf{0}_n,\nonumber\\
&\;\bm{\xi}\succeq\mathbf{U}\mathbf{K}\bm{\alpha}+\beta\mathbf{1}_n,\nonumber\\
&\;\bm{\xi}\succeq\frac{1}{2}\mathbf{1}_n+\frac{1}{2}\mathbf{U}\mathbf{K}\bm{\alpha}+\frac{\beta}{2}\mathbf{1}_n,
\end{align}}
where $\tilde{\mathbf{1}}=[1,\dots,1,0,\dots,0]^T$ is a vector of size $p+n$ with $p$ non-zero entries, $\mathbf{1}$ and $\mathbf{1}_n$ are unitary vectors of size $p+n$ and $n$, respectively, $\mathbf{U}$ is a $n\times(p+n)$ matrix obtained through the concatenation of a $n\times p$ null matrix and an identity matrix of size $n$, $\succeq$ is an element-wise operator, $c_1=\frac{\pi}{2\lambda p}$ and $c_2=\frac{1}{2\lambda n}$.

The equivalent dual problem of~(\ref{eq:doubleprimal}) is more compactly expressed as:
\vspace{-0.2cm}
{\small\begin{align}
\label{eq:doubledual}
\min_{\bm{\sigma},\bm{\delta}}&\:\Big\{\frac{1}{2}\bm{\sigma}^T\mathbf{UKU}^T\bm{\sigma}-c_1\tilde{\mathbf{1}}^T\mathbf{KU}^T\bm{\sigma}-\frac{1}{2}\mathbf{1}_n^T\bm{\delta}\Big\}\nonumber\\
\text{s.t.}&\:\mathbf{1}^T\bigg[c_1\tilde{\mathbf{1}}-\mathbf{U}^T\bm{\sigma}\bigg]=0,\nonumber\\
&\:\bm{\sigma}+\frac{1}{2}\bm{\delta}\preceq c_2\mathbf{1}_n,\nonumber\\
&\:\bm{\sigma}-\frac{1}{2}\bm{\delta}\succeq\mathbf{0}_n,\nonumber\\
&\:\mathbf{0}_n\preceq\bm{\delta}\preceq c_2\mathbf{1}_n,
\end{align}}
where $\bm{\sigma},\bm{\delta}\in\mathbb{R}^n$ and are related to the Lagrange multipliers introduced during the derivation of the dual formulation (see 
Supplementary Material for details).

Due to linearity of constraints in~(\ref{eq:doubleprimal}), Slater's condition is trivially satisfied\footnote{See, e.g., \cite{boyd2004convex}}, thus strong duality holds. This means that~(\ref{eq:doubledual}) can be solved in place of~(\ref{eq:doubleprimal}) to get the primal solution. The optimal $\bm{\alpha}$ can be obtained from one of the stationarity conditions used during the Lagrangian formulation (details in 
Supplementary Material), namely using the following relation
\vspace{-0.3cm}
\begin{equation}
\label{eq:alpha}
\bm{\alpha}=c_1\tilde{\mathbf{1}}-\mathbf{U}^T\bm{\sigma}
\end{equation}
\vspace{-0.1cm}
Note that the bias $\beta$ has to be considered separately, since problem~(\ref{eq:doubledual}) does not give any information on how to compute it (this will be discussed in the next section).

It is to be pointed out that (\ref{eq:doubledual}) is a quadratic programming (QP) problem that can be solved by existing numerical QP optimization libraries. Nevertheless, it is memory inefficient, as it requires storing the Gram matrix, which scales quadratically with the number of training samples. Thus, modern computers cannot manage to solve (\ref{eq:doubledual}) even for a few thousands samples. A question therefore arises: \textbf{is it possible to efficiently find an exact solution to problem~(\ref{eq:doubledual}) without storing the whole Gram matrix?} 

%% file: algorithm.tex
In order to avoid the storage problem, we propose an iterative algorithm that converts problem~(\ref{eq:doubledual}) into a sequence of smaller QP subproblems associated to subsets of training samples (working sets), which require the computation and temporary storage of much smaller Gram matrices.
\begin{algorithm}
\caption{General USMO algorithm}
\label{alg:usmo}
\begin{algorithmic}[1]
\STATE $k\leftarrow 1$.
\STATE Initialize $(\bm{\sigma}^1,\bm{\delta}^1)$.
\WHILE{$(\bm{\sigma}^k,\bm{\delta}^k)$ is not a stationary point of~(\ref{eq:doubledual})}
\STATE Select the working set $S\subset U=\{u:\bm{x}_u\in D_n\}$ with $|S|=2$.
\STATE Compute $\bm{K}_{SS}$, $\bm{K}_{SP}$ and $\bm{K}_{S\bar{S}}$ where $P=\{u:\bm{x}_u\in D_p\}$ and $\bar{S}=U\backslash S$.
\STATE Solve
{\small\begin{align}
\label{eq:qpsub}
\min_{\bm{\sigma}_S^k,\bm{\delta}_S^k}\:&\Big\{\frac{1}{2}(\bm{\sigma}_S^k)^T\bm{K}_{SS}\bm{\sigma}_S^k+\bm{e}^T\bm{\sigma}_S^k-\frac{1}{2}\bm{1}^T\bm{\delta}_S^k\Big\} \nonumber\\
\text{s.t.}\:&\bm{1}^T\bm{\sigma}_S^k=c_1p-\bm{1}^T\bm{\sigma}_{\bar{S}}^k \nonumber\\
&\bm{\sigma}_S^k+\frac{1}{2}\bm{\delta}_S^k\preceq c_2\bm{1} \nonumber\\
&\bm{\sigma}_S^k-\frac{1}{2}\bm{\delta}_S^k\succeq \bm{0} \nonumber\\
&\bm{0}\preceq\bm{\delta}_S^k\preceq c_2\bm{1}
\end{align}}
where {\small$$\bm{e}=\bm{K}_{S\bar{S}}\bm{\sigma}_{\bar{S}}^k-c_1\bm{K}_{SP}\bm{1}_p$$} and 
{\small$$\tilde{\bm{K}}=\begin{bmatrix}
\bm{K}_{PP} & \bm{K}_{PS} & \bm{K}_{P\bar{S}} \\
\bm{K}_{SP} & \bm{K}_{SS} & \bm{K}_{S\bar{S}} \\
\bm{K}_{\bar{S}P} & \bm{K}_{\bar{S}S} & \bm{K}_{\bar{S}\bar{S}} 
\end{bmatrix},\;
\tilde{\bm{\sigma}}^k=\begin{bmatrix}
\bm{\sigma}_S^k \\
\bm{\sigma}_{\bar{S}}^k
\end{bmatrix},\;
\tilde{\bm{\delta}}^k=\begin{bmatrix}
\bm{\delta}_S^k \\
\bm{\delta}_{\bar{S}}^k
\end{bmatrix}$$}
$\tilde{\bm{K}}, \tilde{\bm{\sigma}}^k$ and $\tilde{\bm{\delta}}^k$ are permutations of $\bm{K}, \bm{\sigma}^k$ and $\bm{\delta}^k$, respectively. In general, $\bm{K}_{VW}$ is used to denote a matrix containing rows of $\bm{K}$ indexed by elements in set $V$ and columns of $\bm{K}$ indexed by elements in set $W$.
\STATE $(\bm{\sigma}_S^{k+1},\bm{\delta}_S^{k+1})\leftarrow(\bm{\sigma}_S^k,\bm{\delta}_S^k)$.
\STATE $k\leftarrow k+1$.
\ENDWHILE
\end{algorithmic}
\end{algorithm}

Each iteration of USMO is made of three steps: selection of the working set $S$, computation of the Gram matrix for samples associated to indices in $S$, solution of a QP subproblem, where only terms depending on $S$ are considered. Details are provided in Algorithm~\ref{alg:usmo}. It is to be mentioned that in principle this strategy allows decreasing the storage requirement at the expense of a heavier computation. In fact, the same samples may be selected multiple times over iterations, thus requiring recomputing matrices $\bm{K}_{SS}, \bm{K}_{SP}$ and $\bm{K}_{S\bar{S}}$. We will see later how to deal with this inefficiency. Another important aspect is that at each iteration only few parameters are updated (those indexed by the working set $S$, namely $\bm{\sigma}_S^k$ and $\bm{\delta}_S^k$), while the others are kept fixed. Here, we consider a working set of size two, as this allows solving the QP subproblem~(\ref{eq:qpsub}) analytically, without the need for further optimization. This is discussed in the next subsection.

\vspace{-0.7cm}
{\subsection{QP Subproblem}\label{sec:subproblem}}
\vspace{-0.1cm}
We start by considering the following Lemma (proof in 
Supplementary Material):
\begin{lemma}
\label{th:twovars}
Given $S=\{i,j\}$, any optimal solution $\bm{\sigma}_S^*=[\sigma_i^*\:\sigma_j^*]^T$, $\bm{\delta}_S^*=[\delta_i^*\:\delta_j^*]^T$ of the QP subproblem~(\ref{eq:qpsub}) has to satisfy the following condition: $\forall u:\bm{x}_u\in S\wedge0\leq\delta_u^*\leq c_2$ either $\sigma_u^*=c_2-\frac{\delta_u^*}{2}$ or $\sigma_u^*=\frac{\delta_u^*}{2}$.
\end{lemma}
\vspace{-0.1cm}
This tells us that the optimal solution $(\bm{\sigma}_S^*,\bm{\delta}_S^*)$ assumes a specific form and can be calculated by searching in a smaller space. In particular, four subspaces can be identified for search :
{\small\begin{align}
\label{eq:qpcases}
&\text{Case 1:}\:\sigma_i^k=c_2-\frac{\delta_i^k}{2}\:\wedge\:\sigma_j^k=c_2-\frac{\delta_j^k}{2}\:\wedge\:0\leq\delta_i^k,\delta_j^k\leq c_2,\nonumber\\
&\text{Case 2:}\:\sigma_i^k=c_2-\frac{\delta_i^k}{2}\:\wedge\:\sigma_j^k=\frac{\delta_j^k}{2}\:\wedge\:0\leq\delta_i^k,\delta_j^k\leq c_2,\nonumber\\
&\text{Case 3:}\:\sigma_i^k=\frac{\delta_i^k}{2}\:\wedge\:\sigma_j^k=c_2-\frac{\delta_j^k}{2}\:\wedge\:0\leq\delta_i^k,\delta_j^k\leq c_2,\nonumber\\
&\text{Case 4:}\:\sigma_i^k=\frac{\delta_i^k}{2}\:\wedge\:\sigma_j^k=\frac{\delta_j^k}{2}\:\wedge\:0\leq\delta_i^k,\delta_j^k\leq c_2,
\end{align}}
Then, in order to solve the QP subproblem~(\ref{eq:qpsub}), one can solve four optimization subproblems, where the objective function is the same as~(\ref{eq:qpsub}), but the inequality constraints of~(\ref{eq:qpsub}) are simplified to~(\ref{eq:qpcases}). Each of these subproblems can be expressed as an optimization of \textbf{just one variable}, by exploiting the equality constraints of both~(\ref{eq:qpsub}) and~(\ref{eq:qpcases}). It can therefore be solved analytically, without the need for further optimization. Table~\ref{tab:update} reports the equations used to solve the four subproblems (we omit the derivation, which is straightforward),
\begin{table}
\caption{Equations and Conditions Used to Solve the Four QP Subproblems.}
\label{tab:update}
\vspace{-2em}
\centering
\resizebox{0.35\textwidth}{!}{
\begin{tabular}[t]{cl}
\hline
\textbf{Case}&\textbf{Equations}\\
\hline
1&
$\sigma_j^k{=}(a^k(k(\bm{x}_i,\bm{x}_i)-k(\bm{x}_i,\bm{x}_j))+e_1-e_2)\eta$\\
2&
$\sigma_j^k{=}(a^k(k(\bm{x}_i,\bm{x}_i)-k(\bm{x}_i,\bm{x}_j))+e_1-e_2+2)/\eta$\\
3&
$\sigma_j^k{=}(a^k(k(\bm{x}_i,\bm{x}_i)-k(\bm{x}_i,\bm{x}_j))+e_1-e_2-2)/\eta$\\
4&
$\sigma_j^k{=}(a^k(k(\bm{x}_i,\bm{x}_i)-k(\bm{x}_i,\bm{x}_j))+e_1-e_2)/\eta$\\
\hline
\textbf{Case}&\textbf{Conditions}\\
\hline
1&
{\small$\max\{c_2/2,a^k{-}c_2\}{\leq}\sigma_j^k{\leq}\min\{c_2,a^k{-}c_2/2\}$}\\
2&
{\small$\max\{0,a^k{-}c_2\}{\leq}\sigma_j^k{\leq}\min\{c_2/2,a^k{-}c_2/2\}$}\\
3&
{\small$\max\{c_2/2,a^k{-}c_2/2\}{\leq}\sigma_j^k{\leq}\min\{c_2,a^k\}$}\\
4&
{\small$\max\{0,a^k{-}c_2/2\}{\leq}\sigma_j^k{\leq}\min\{c_2/2,a^k\}$}\\\hline
\textbf{Note:}&$a^k=c_1p-\bm{1}^T\bm{\sigma}_{\bar{S}}^k$, $\bm{e}=[e_1,e_2]^T$ and\\
&$\eta=k(\bm{x}_i,\bm{x}_i)+k(\bm{x}_j,\bm{x}_j)-2k(\bm{x}_i,\bm{x}_j).$
\end{tabular}}
\end{table}
where $\sigma_j^k$ is computed for all four cases. All other variables, namely $\sigma_i^k$, $\delta_i^k$ and $\delta_j^k$, can be obtained in a second phase by simply exploiting the equality constraints in~(\ref{eq:qpsub}) and~(\ref{eq:qpcases}). 

These equations do not guarantee that the inequalities in~(\ref{eq:qpcases}) are satisfied. To verify this, one can rewrite these inequalities as equivalent conditions of only $\sigma_j^k$ (by exploiting the equality constraints in~(\ref{eq:qpsub}) and~(\ref{eq:qpcases})), and check $\sigma_j^k$ against them, as soon as all $\sigma_j^k$ are available. If these conditions are violated, a proper clipping is applied to $\sigma_j^k$ to restore feasibility. Table~\ref{tab:update} summarizes the checking conditions. 

Finally, the minimizer of the QP subproblem~(\ref{eq:qpsub}) can be obtained by retaining only the solution with the lowest level of objective. At each iteration, the output of the algorithm is both optimal and feasible for the QP subproblem~(\ref{eq:qpsub}). The question now is: when is it also optimal for the problem~(\ref{eq:doubledual})?
%
%
%
%
\vspace{-0.9cm}
{\subsection{Optimality Conditions}\label{sec:optimality}}
A problem of any optimization algorithm is to determine the stop condition. In Algorithm~\ref{alg:usmo}, the search of the solution is stopped as soon as some stationarity conditions are met. These conditions, called Karush-Kuhn-Tucker (KKT) conditions, represent the certificates of optimality for the obtained solution. In case of~(\ref{eq:doubledual}) they are both necessary and sufficient conditions, since the objective is convex and the constraints are affine functions \cite{hanson1999invexity}. More in detail, an optimal solution has to satisfy the following relations:
{\small\begin{align}
\label{eq:kkt}
&\frac{\partial F(\bm{\sigma},\bm{\delta})}{\partial \sigma_u}-\beta=-\lambda_u+\mu_u,\nonumber\\
&\frac{\partial F(\bm{\sigma},\bm{\delta})}{\partial \delta_u}=-\frac{\lambda_u}{2}-\frac{\mu_u}{2}-\xi_u+\eta_u,\nonumber\\
&\lambda_u\big(\sigma_u+\frac{\delta_u}{2}-c_2\big)=0,\nonumber\\
&\mu_u\big(\frac{\delta_u}{2}-\sigma_u\big)=0,\nonumber\\
&\xi_u\big(\delta_u-c_2\big)=0,\nonumber\\
&\eta_u\delta_u=0,\nonumber\\
&\lambda_u,\mu_u,\xi_u,\eta_u\geq 0,
\end{align}}
and this is valid for any component of the optimal solution, namely $\forall u:\bm{x}_u\in D_n$. In~(\ref{eq:kkt}) $F(\bm{\sigma},\bm{\delta})$ is used as an abbreviation of the objective function of~(\ref{eq:doubledual}), while $\beta,\lambda_u,\mu_u,\xi_u,\eta_u$ are the Lagrange multipliers introduced to deal with the constraints in~(\ref{eq:doubledual}). These conditions can be rewritten more compactly as:
{\small\begin{align}
\label{eq:opt}
0\leq\delta_u{<}c_2\:\wedge\:\sigma_u{=}\frac{\delta_u}{2}&\:\Rightarrow\:
\frac{\partial F(\bm{\sigma},\bm{\delta})}{\partial \sigma_u}{-}\beta\geq1\nonumber\\
&\:\Rightarrow\:f(\bm{x}_u)\leq{-}1,\nonumber\\
0\leq\delta_u{<}c_2\:\wedge\:\sigma_u{=}c_2{-}\frac{\delta_u}{2}&\:\Rightarrow\:
\frac{\partial F(\bm{\sigma},\bm{\delta})}{\partial \sigma_u}{-}\beta\leq{-}1\nonumber\\
&\:\Rightarrow\:f(\bm{x}_u)\geq1,\nonumber\\
\delta_u{=}c_2\:\wedge\:\sigma_u{=}\frac{c_2}{2}&\:\Rightarrow\:
-1\leq\frac{\partial F(\bm{\sigma},\bm{\delta})}{\partial \sigma_u}{-}\beta\leq 1\nonumber\\
&\:\Rightarrow\:{-}1\leq f(\bm{x}_u)\leq1,
\end{align}}
In order to derive both~(\ref{eq:kkt}) and (\ref{eq:opt}), one can follow a strategy similar to the proof of Lemma~\ref{th:twovars}. It is easy to verify that $\frac{\partial F(\bm{\sigma},\bm{\delta})}{\partial \sigma_u}=-f(\bm{x}_u)+\beta$. Thus,~(\ref{eq:opt}) provides also conditions on the target function $f$. From now on, we will refer to~(\ref{eq:opt}) as the \textbf{optimality conditions}, to distinguish from approximate conditions used to deal with numerical approximations of calculators. To this aim, the $\bm{\tau}-$\textbf{optimality conditions} are introduced, namely:
{\small\begin{align}
\label{eq:tauopt}
0\leq\delta_u{<}c_2\:\wedge\:\sigma_u{=}\frac{\delta_u}{2}&\:\Rightarrow\:\frac{\partial F(\bm{\sigma},\bm{\delta})}{\partial \sigma_u}{-}\beta\geq1-\frac{\tau}{2},\nonumber\\
&\:\Rightarrow\:f(\bm{x}_u)\leq{-}1+\frac{\tau}{2},\nonumber\\
0\leq\delta_u{<}c_2\wedge\sigma_u{=}c_2{-}\frac{\delta_u}{2}&\:\Rightarrow\:\frac{\partial F(\bm{\sigma},\bm{\delta})}{\partial \sigma_u}{-}\beta\leq{-}1+\frac{\tau}{2},\nonumber\\
&\:\Rightarrow\:f(\bm{x}_u)\geq1-\frac{\tau}{2},\nonumber\\
\delta_u{=}c_2\:\wedge\:\sigma_u{=}\frac{c_2}{2}&\:\Rightarrow\:-1{-}\frac{\tau}{2}{\leq}\frac{\partial F(\bm{\sigma},\bm{\delta})}{\partial \sigma_u}{-}\beta{\leq} 1{+}\frac{\tau}{2},\nonumber\\
&\:\Rightarrow\:{-}1{-}\frac{\tau}{2}\leq f(\bm{x}_u)\leq1{+}\frac{\tau}{2},
\end{align}}
where $\tau$ is a real-positive scalar used to perturb the optimality conditions.

%
%
%
%

By introducing the sets {\small$D_1(\bm{\sigma},\bm{\delta})=\big\{\bm{x}_u\in D_n: 0\leq\delta_u<c_2\:\wedge\:\sigma_u=\frac{\delta_u}{2}\big\}$}, {\small$D_2(\bm{\sigma},\bm{\delta})=\big\{\bm{x}_u\in D_n: 0\leq\delta_u<c_2\:\wedge\:\sigma_u=c_2-\frac{\delta_u}{2}\big\}$} and {\small$D_3(\bm{\sigma},\bm{\delta})=\big\{\bm{x}_u\in D_n: 0<\delta_u\leq c_2\:\wedge\:\big(\sigma_u=\frac{\delta_u}{2}\:\vee\:\sigma_u=c_2-\frac{\delta_u}{2}\big)\big\}$} and the quantities {\small$m_1^{max}(\bm{\sigma},\bm{\delta})=\max_{\bm{x}_u\in D_1}f(\bm{x}_u)$}, {\small$m_2^{min}(\bm{\sigma},\bm{\delta})=\min_{\bm{x}_u\in D_2}f(\bm{x}_u)$}, {\small$m_3^{min}(\bm{\sigma},\bm{\delta})=\min_{\bm{x}_u\in D_3}f(\bm{x}_U)$} and  {\small$m_3^{max}(\bm{\sigma},\bm{\delta})=\max_{\bm{x}_u\in D_3}f(\bm{x}_u)$}, called the  \textbf{most critical values}, it is possible to rewrite conditions~(\ref{eq:tauopt}) in the following equivalent way:
{\small\begin{align}
\label{eq:maxtauopt}
&m_1^{max}(\bm{\sigma},\bm{\delta})-m_3^{min}(\bm{\sigma},\bm{\delta})\leq\tau,\nonumber\\
&m_3^{max}(\bm{\sigma},\bm{\delta})-m_2^{min}(\bm{\sigma},\bm{\delta})\leq\tau,\nonumber\\
&m_1^{max}(\bm{\sigma},\bm{\delta})-m_2^{min}(\bm{\sigma},\bm{\delta})+2\leq\tau,
\end{align}}
Apart from being written more compactly than~(\ref{eq:tauopt}), conditions~(\ref{eq:maxtauopt}) have the advantage that they can be computed without knowing the bias $\beta$. Due to the dependence on $\bm{\sigma}$ and $\bm{\delta}$, $m_1^{max}$, $m_2^{min}$, $m_3^{min}$ $m_3^{max}$ need to be tracked and computed at each iteration in order to check $\tau-$optimality and to decide whether to stop the algorithm.

\vspace{0.1cm}
{\subsection{Working Set Selection}\label{sec:working}}
A natural choice for selecting the working set is to look for pairs violating the $\tau-$optimality conditions. In particular,
\begin{definition}
Any pair $(\bm{x}_i,\bm{x}_j)$ from $D_n$ is a \textbf{violating pair}, if and only if it satisfies the following relations:
{\small\begin{align}
\label{eq:violating}
& \bm{x}_i\in D_1, \bm{x}_j\in D_3\Rightarrow f(\bm{x}_i)-f(\bm{x}_j)>\tau,\nonumber\\
& \bm{x}_i\in D_3, \bm{x}_j\in D_1\Rightarrow f(\bm{x}_i)-f(\bm{x}_j)<-\tau,\nonumber\\
& \bm{x}_i\in D_2, \bm{x}_j\in D_3\Rightarrow f(\bm{x}_i)-f(\bm{x}_j)<-\tau,\nonumber\\
& \bm{x}_i\in D_3, \bm{x}_j\in D_2\Rightarrow f(\bm{x}_i)-f(\bm{x}_j)>\tau,\nonumber\\
& \bm{x}_i\in D_1, \bm{x}_j\in D_2\Rightarrow f(\bm{x}_i)-f(\bm{x}_j)+2>\tau,\nonumber\\
& \bm{x}_i\in D_2, \bm{x}_j\in D_1\Rightarrow f(\bm{x}_i)-f(\bm{x}_j)-2<-\tau,
\end{align}}
\end{definition}
Conditions~(\ref{eq:maxtauopt}) are not satisfied as long as violating pairs are found. Therefore, the algorithm keeps looking for violating pairs and use them to improve the objective function until $\tau-$optimality is reached.

The search of violating pairs as well as the computation of the most critical values go hand in hand in the optimization and follow two different approaches. The former consists of finding violating pairs and computing the most critical values based only on a subset of samples called the \textbf{non-bound} set, namely $D_n^{-}=(D_1\cap D_3)\cup(D_2\cap D_3)$,\footnote{The term \textit{non-bound} comes from the fact that $0<\delta_u<c_2$ for all $\bm{x}_u\in D_n$.} while the latter consists of looking for violating pairs based on the whole set $D_n$ by scanning all samples one by one. In this second approach, the most critical values are updated using the non-bound set together with the current examined sample. Only when all samples are examined, it is possible to check conditions~(\ref{eq:maxtauopt}), since the most critical values correspond to the original definition. The algorithm keeps using the first approach until $\tau-$optimality for the non-bound set is reached, after that the second approach is used. This process is repeated until the $\tau-$optimality for the whole set $D_n$ is achieved. On the one hand, checking these conditions only on the non-bound set is very efficient but does not ensure the global $\tau$ optimality; on the other hand, the use of the whole unlabeled set is much more expensive, while ensuring the global $\tau$ optimality

The motivation of having two different approaches for selecting the violating pairs and computing the most critical values is to enhance efficiency in computation. This will be clarified in the next subsection.

\vspace{-0.5cm}
{\subsection{Function Cache and Bias Update}\label{sec:function}}
Recall that each iteration of the USMO algorithm is composed by three main operations, namely: the working set selection, the resolution of the associated QP subproblem, and the update of the most critical values based on the obtained solution. It is interesting to note that all these operations require to compute the target function $f(\bm{x}_u)$ for all $\bm{x}_u$ belonging either to the non-bound set or to the whole set $D_n$, depending on the approach selected by that iteration. In fact, the stage of working set selection requires to evaluate conditions in~(\ref{eq:violating}) for pairs of samples depending on $f$; the equations used to solve the QP subproblem, shown in Table~\ref{tab:update}, depend on vector $\bm{e}=\bm{K}_{S\bar{S}}\bm{\sigma}_{\bar{S}}^k-c_1\bm{K}_{SP}\bm{1}_p+\bm{K}_{SS}\bm{\sigma}_S^{k-1}-\bm{K}_{SS}\bm{\sigma}_{S}^{k-1}=-[f(\bm{x}_i)-\beta,f(\bm{x}_j)-\beta]^T-\bm{K}_{SS}\bm{\sigma}_{S}^{k-1}$, which is also influenced by $f$; finally, the computation of the most critical values requires to evaluate the target function $f$. Therefore, it is necessary to define a strategy that limits the number of times the target function is evaluated at each iteration. This can be achieved by considering the fact that the algorithm performs most of the iterations on samples in the non-bound set, while the whole set is used mainly to check if $\tau-$optimality is reached, and then the values of the target function for those samples can be stored in a cache, called the \textbf{function cache}. Since usually $|D_n^{-}|\ll|D_n|$, storing $f(\bm{x}_u)$ for all $\bm{x}_u\in D_n^{-}$ is a cheap operation, which allows to save a huge amount of computation, thus increasing the computational efficiency.

At each iteration the function cache has to be updated in order to take into account the changes occured at some of the entries of vectors $\bm{\sigma}$ and $\bm{\delta}$, or equivalently at some of the entries of $\bm{\alpha}$ . By defining $\mathcal{F}^k(\bm{x}_u)$ as the function cache for sample $\bm{x}_u$ at iteration $k$, it is possible to perform the update operation using the following relation:
{\small\begin{align}
\label{eq:update}
\mathcal{F}^k(\bm{x}_u)=&\mathcal{F}^{k-1}(\bm{x}_u)+(\alpha_i^k-\alpha_u^{k-1})k(\bm{x}_i,\bm{x}_u)\nonumber\\
&+(\alpha_j^k-\alpha_j^{k-1})k(\bm{x}_j,\bm{x}_u)
\end{align}}
\vspace{-0.05cm}
Since all operations at each iteration are invariant with respect to $\beta$ (because they require to evaluate differences between target function values), $\beta$ can be computed at the end of the algorithm, namely when $\tau-$optimality is reached. By exploiting the fact that the inequalities in~(\ref{eq:opt}) become simply equalities for samples in the non-bound set, meaning that the target function evaluated at those samples can assume only two values, 1 or -1,\footnote{The same principle holds for conditions~(\ref{eq:tauopt}), but in this case the inequalities are defined over arbitrary small intervals centered at 1 and -1 rather than being equalities.} it is possible to compute $\beta$ for each of these samples in the following way:
\vspace{-0.1cm}
{\small\begin{equation}
\label{eq:betasample}
\beta_u=\Big\{\begin{array}{ll}
-1-\mathcal{F}(\bm{x}_u),\:&\forall\bm{x}_u\in D_1\cap D_3\\
1-\mathcal{F}(\bm{x}_u),\:&\forall\bm{x}_u\in D_2\cap D_3
\end{array}
\end{equation}}
The final $\beta$ can be computed by averaging of~(\ref{eq:betasample}) over all samples in the non-bound set, in order to reduce the effect of wrong label assignment.

\begin{figure}[t!]
\centering
\vspace{-1.5em}
\subfloat[]{\includegraphics[width=0.24\textwidth]{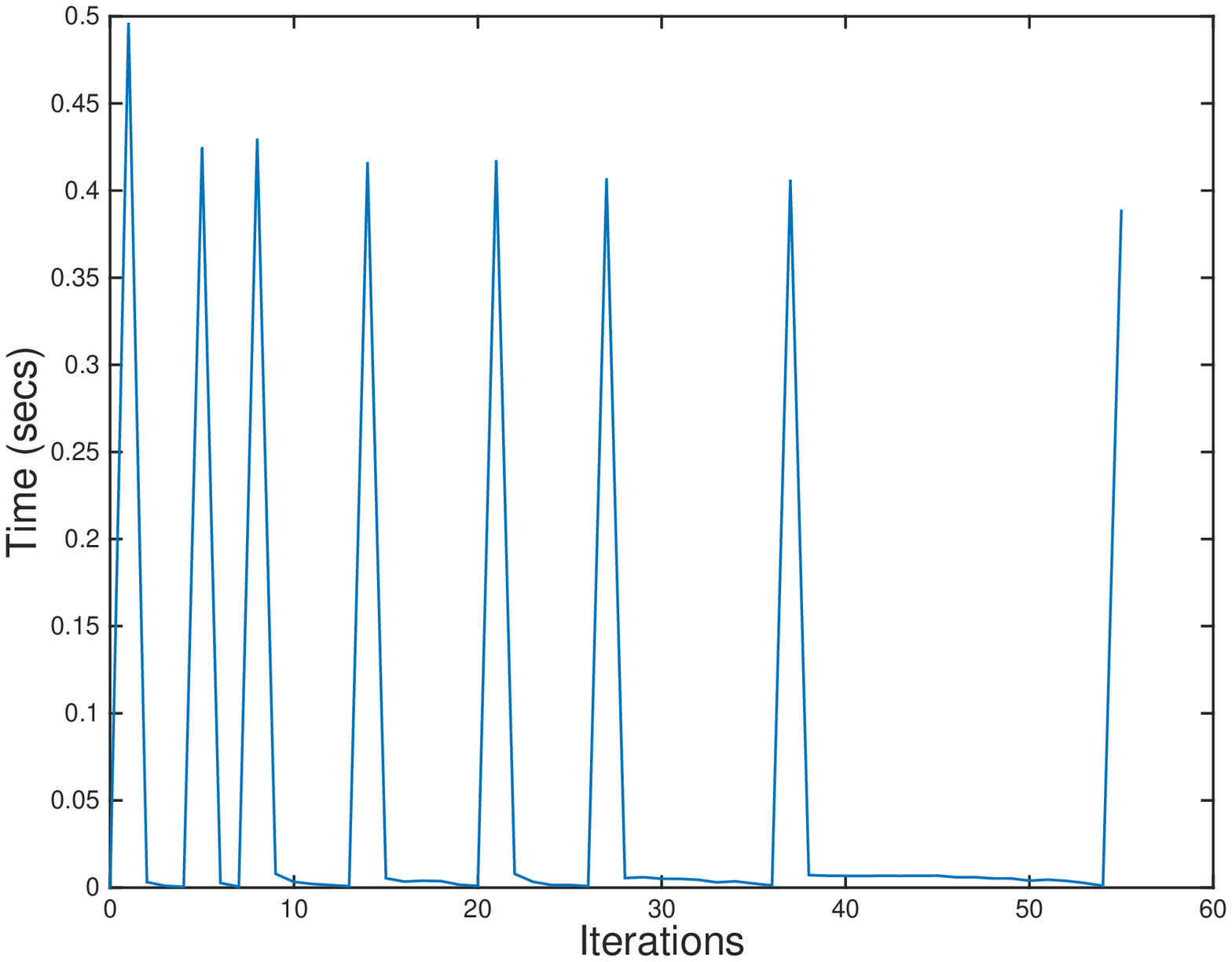}}
\subfloat[]{\includegraphics[width=0.24\textwidth]{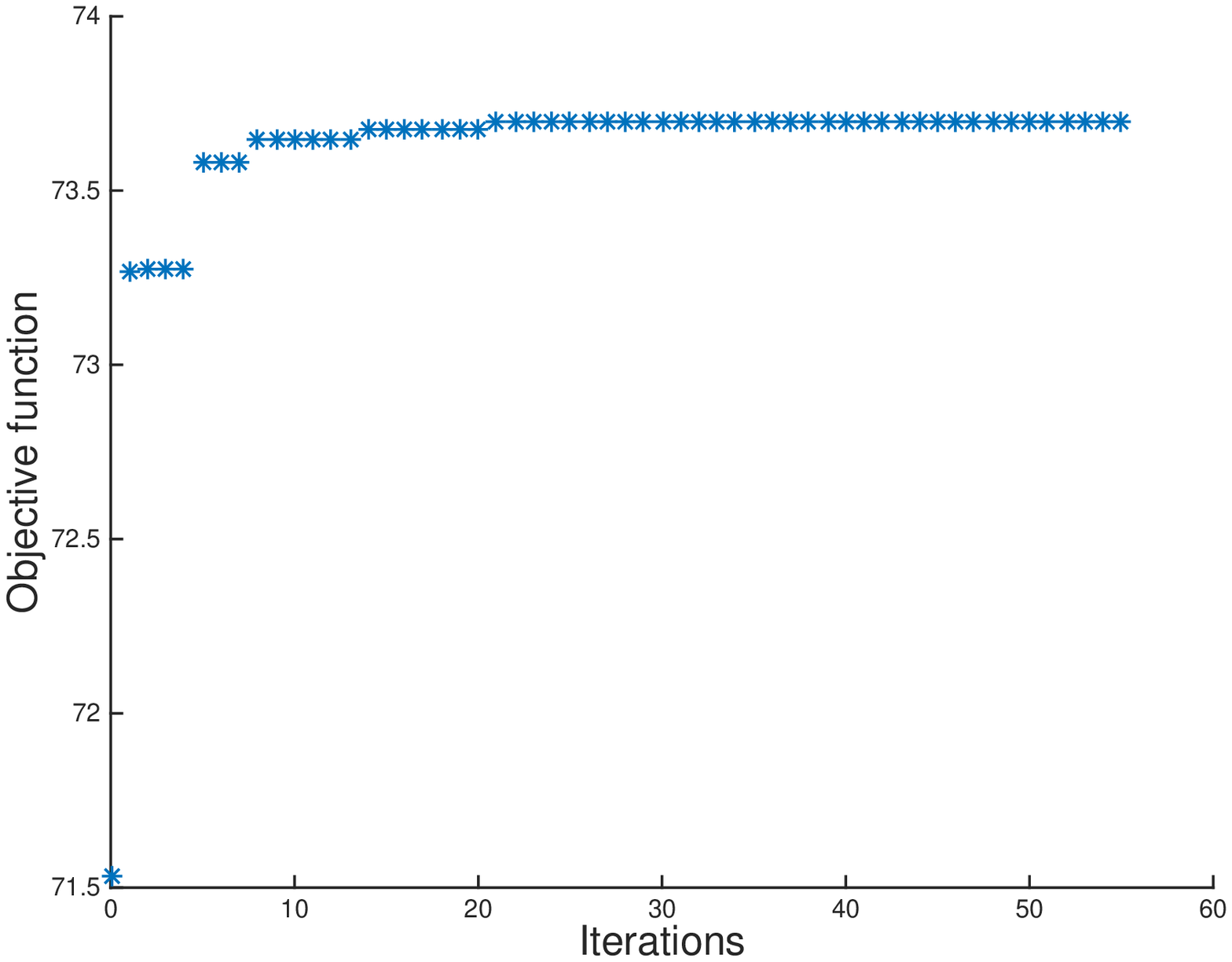}}
\vspace{-0.5em}
\caption{(a) plot of training time over iterations, (b) learning curve expressed in terms of objective function.}
\label{fig:optimization}
\end{figure}

{\subsection{Initialization}\label{sec:initialization}}
As previously mentioned, the proposed algorithm is characterized by iterations focusing either on 
the whole training data set or on a smaller set of non-bound samples. The formers are computationally 
more expensive, not only because a larger amount of samples is involved in the optimization, but 
also because the algorithm has to perform many evaluation operations, which can be skipped in the 
latter case, thanks to the exploitation of the function cache. An example of this is provided in 
Figure~\ref{fig:optimization}: the chart on the left plots the time required by the algorithm to 
complete the corresponding set of iterations. Peaks correspond to cases where the whole training 
data set is considered, while valleys represent the cases where iterations are performed over the 
non-bound samples. The chart on the right plots the overall objective score vs. the different sets 
of iterations. Jumps are associated with cases where all training samples are involved in the 
optimization. As soon as the algorithm approaches convergence, the contribution of the first kind 
of iterations becomes less and less relevant.
Therefore, the selection of a good starting point is important
to limit the number of iterations requested over the whole
unlabeled dataset.
Given the convex nature of the problem, any suboptimal solution
achievable with a low complexity can be used as a starting point.
In our method, we propose the following heuristic procedure.
Labeled samples are used to train a one-class SVM~\cite{scholkopf2001estimating}, that is in turn used to rank the unlabeled samples according to their value of estimated target function. From this ordered list it is possible to identify groups of samples that can be associated with the cases in~(\ref{eq:opt}). In particular, we identify five groups of samples corresponding to the following five cases:
{\small\begin{align}
\label{eq:cases}
&\delta_u^{(1)}=0\:\wedge\:\sigma_u^{(1)}=0,\nonumber\\
&0<\delta_u^{(2)}<c_2\:\wedge\:\sigma_u=\frac{\delta_u^{(2)}}{2},\nonumber\\
&\delta_u^{(3)}=c_2\:\wedge\:\sigma_u^{(3)}=\frac{c_2}{2},\nonumber\\
&0<\delta_u^{(4)}<c_2\:\wedge\:\sigma_u^{(4)}=c_2-\frac{\delta_u}{2},\nonumber\\
&\delta_u^{(5)}=0\:\wedge\:\sigma_u^{(5)}=c_2,
\end{align}}
The size of each group as well as the initial parameters for cases in~(\ref{eq:cases}) can be computed by solving an optimization problem, whose objective function is defined starting from the equality constraint in~(\ref{eq:doubledual}). In particular, by defining $n_1,n_2,n_3,n_4,n_5$ as the sizes of the groups for the different cases and by assuming that $n_1=(1-\pi)an$, $n_2=bn$, $n_3=(1-a-b-c)n$, $n_4=cn$, $n_5=\pi an$, where $a,b,c\in\mathbb{R}^+$, and that the parameters for the second and the fourth cases in~(\ref{eq:cases}), namely $\sigma_u^{(2)}$ and $\sigma_u^{(4)}$, are the same, the optimization problem can be formulated in the following way:
\vspace{-0.1cm}
{\small\begin{align}
\label{eq:initial}
\min_{\substack{\sigma_u^{(2)},\sigma_u^{(4)},\\ a,b,c}}&\:\Big\{c_1p-bn\sigma_u^{(2)}-cn\sigma_u^{(4)}-c_2n\Big[\pi a +\frac{1-a-b-c}{2}\Big]\Big\}^2\nonumber\\
\text{s.t.}&\:0\leq a+b+c\leq 1-\frac{1}{n},\nonumber\\
&\:\max\Big\{\frac{1}{\pi n},\frac{1}{(1-\pi)n}\Big\}\leq a\leq\min\Big\{\frac{1}{1-\pi},\frac{1}{\pi}\Big\},\nonumber\\
&\:\frac{1}{n}\leq b,c\leq \frac{\log(n)}{n},\nonumber\\
&\:0<\sigma_u^{(2)}<\frac{c_2}{2}\:\wedge\:\frac{c_2}{2}<\sigma_u^{(4)}<c_2,
\end{align}}
where the constraints in~(\ref{eq:initial}) can be obtained by imposing $n_1+n_2+n_3+n_4+n_5=n$ and $1\leq n_2,n_3,n_4,n_5\leq n$. Furthermore, we decide to have some upper bounds for $b$ and $c$ to limit the size of the initial non-bound set.

In practice, after ranking the unlabeled samples through the one-class SVM and solving the optimization problem in~(\ref{eq:initial}), the initial solution is obtained by assigning to each sample the value of parameters corresponding to the case that sample belongs to. For example, if the samples are ranked in ascending order, then the first $n_1$ samples in the list have $\sigma_u=0$ and $\delta_u=0$, the next $n_2$ samples have $\sigma_u=\sigma_u^{(2)}$ and $\delta_u=2\sigma_u$ and the others follow the same strategy.

%% file: theory.tex
\externaldocument{src/algorithm}
\begin{figure}[!t]
\centering
\vspace{-1em}
\includegraphics[width=0.3\textwidth]{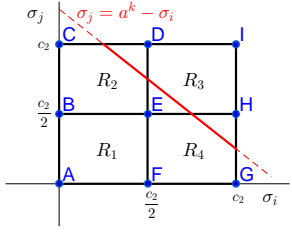}
\vspace{-3em}
\caption{Subdivision of the feasible region in the plane defined by the variables $\sigma_i$ and $\sigma_j$. The red solid line represents the feasible region including the equality constraint in~(\ref{eq:qpsub}).}
\label{fig:feasible}
\end{figure}
In this section, we present the main theoretical result, namely, we prove that Algorithm~\ref{alg:usmo} converges to a $\tau-$optimal solution.

It is important to recall that each iteration of USMO requires to solve an optimization subproblem, that depends on a single variable. In particular, if $\bm{x}_i$ and $\bm{x}_j$ correspond to the selected pair of points at one iteration, then the solution space corresponds to a line lying in the two-dimensional plane defined by the variables $\sigma_i$ and $\sigma_j$. The feasible region in that plane can be subdivided into four parts, as defined according to Figure~\ref{fig:feasible}. These regions are considered closed sets, therefore including boundary points, like edges and corners. To consider only the interior of any set $U$, we use the notation $\text{int }U$. Based on these considerations, it is possible to prove the following lemma.
\begin{lemma}
\label{th:minimization}
Let the vector $\bm{z}'=[\bm{\sigma}';\bm{\delta}']$ be in the feasible set of~(\ref{eq:doubledual}) and $(\bm{x}_i,\bm{x}_j)$ be a violating pair at point $\bm{z}'$. Let also $\bm{z}^*=[\bm{\sigma}^*;\bm{\delta}^*]$ be the solution obtained after applying one iteration of the Algorithm~\ref{alg:usmo} using the working set $S=\{i,j\}$ and starting from $\bm{z}'$. Then, the following results hold:
\begin{itemize}
\item[(a)] $\bm{z}^*\neq\bm{z}'$,
\item[(b)] After the minimization step, $(\bm{x}_i,\bm{x}_j)$ is no more a violating pair,
\item[(c)] $(\sigma_i^*{,}\sigma_j^*)\in \text{int }R_1\cup \text{int }R_3\:\Rightarrow\:f_{\bm{z}^*}(\bm{x}_j){-}f_{\bm{z}^*}(\bm{x}_i)=0$,
\item[(d)] $(\sigma_i^*{,}\sigma_j^*)\in \text{int }R_2\:\Rightarrow\:f_{\bm{z}*}(\bm{x}_j){-}f_{\bm{z}^*}(\bm{x}_i)=2$,
\item[(e)] $(\sigma_i^*{,}\sigma_j^*)\in \text{int }R_4\:\Rightarrow\:f_{\bm{z}^*}(\bm{x}_j){-}f_{\bm{z}^*}(\bm{x}_i)=-2$,
\item[(f)] $(\sigma_i^*{,}\sigma_j^*)\in BE\Rightarrow 0\leq f_{\bm{z}^*}(\bm{x}_j){-}f_{\bm{z}^*}(\bm{x}_i)\leq 2$,
\item[(g)] $(\sigma_i^*{,}\sigma_j^*)\in DE\Rightarrow 0\leq f_{\bm{z}^*}(\bm{x}_j){-}f_{\bm{z}^*}(\bm{x}_i)\leq 2$,
\item[(h)] $(\sigma_i^*{,}\sigma_j^*)\in EF\Rightarrow -2\leq f_{\bm{z}^*}(\bm{x}_j){-}f_{\bm{z}^*}(\bm{x}_i)\leq 0$,
\item[(i)] $(\sigma_i^*{,}\sigma_j^*)\in EH\Rightarrow -2\leq f_{\bm{z}^*}(\bm{x}_j){-}f_{\bm{z}^*}(\bm{x}_i)\leq 0$,
\item[(l)] $(\sigma_i^*{,}\sigma_j^*)\in AB\cup DI\Rightarrow f_{\bm{z}^*}(\bm{x}_j){-}f_{\bm{z}^*}(\bm{x}_i)\geq 0$,
\item[(m)] $(\sigma_i^*{,}\sigma_j^*)\in AF\cup HI\Rightarrow f_{\bm{z}^*}(\bm{x}_j){-}f_{\bm{z}^*}(\bm{x}_i)\leq 0$,
\item[(n)] $(\sigma_i^*{,}\sigma_j^*)\in BC\cup CD\Rightarrow f_{\bm{z}^*}(\bm{x}_j){-}f_{\bm{z}^*}(\bm{x}_i)\geq 2$,
\item[(o)] $(\sigma_i^*{,}\sigma_j^*)\in FG\cup GH\Rightarrow f_{\bm{z}^*}(\bm{x}_j){-}f_{\bm{z}^*}(\bm{x}_i)\leq -2$,
\item[(p)] $F(\bm{\sigma}',\bm{\delta}')-F(\bm{\sigma}^*,\bm{\delta}^*)>\frac{\tau}{2\sqrt[]{2}}\|\bm{\sigma}'-\bm{\sigma}^*\|_2$,
\end{itemize}
where $f_{\bm{z}^*}$ represents the target function with coefficients $\alpha_i$ computed according to~(\ref{eq:alpha}) using $\bm{z}^*$ (see Figure~\ref{fig:feasible}).
\end{lemma}
\begin{IEEEproof}
Note that the feasible region for the QP subproblem~(\ref{eq:qpsub}) is a portion of line with negative slope lying on the plane defined by variables $\sigma_i$ and $\sigma_j$ (see Figure~\ref{fig:feasible}). Thus, any point $(\sigma_i,\sigma_j)$ on this line can be expressed using the following relationship:
\vspace{-0.1cm}
{\small\begin{align}
\label{eq:line}
&\sigma_i=\sigma_i'+t,\nonumber\\
&\sigma_j=\sigma_j'-t,
\end{align}}
where $t\in\mathbb{R}$. In particular, if $t=0$, then $(\sigma_i,\sigma_j)\equiv(\sigma_i',\sigma_j')$ and, if $t=t^*$, then $(\sigma_i,\sigma_j)\equiv(\sigma_i^*,\sigma_j^*)$.

Considering~(\ref{eq:line}) and the fact that $\delta_{\bm{x_i}}{=}2\sigma_i\:\wedge\:\delta_j{=}2\sigma_j$ when $(\sigma_i,\sigma_j)\in R_1$, $\delta_{\bm{x_i}}{=}2\sigma_i\:\wedge\:\delta_j{=}2c_2-2\sigma_j$ when $(\sigma_i,\sigma_j)\in R_2$, $\delta_{\bm{x_i}}{=}c_2-2\sigma_i\:\wedge\:\delta_j{=}c_2-2\sigma_j$ when $(\sigma_i,\sigma_j)\in R_3$ and $\delta_{\bm{x_i}}{=}c_2-2\sigma_i\:\wedge\:\delta_j{=}2\sigma_j$ when $(\sigma_i,\sigma_j)\in R_4$, it is possible to rewrite the objective in~(\ref{eq:qpsub}) as a function of $t$, namely:
{\small\begin{align}
\label{eq:monoobj}
\phi(t)=&\frac{1}{2}(\sigma_i'+t)^2k(\bm{x}_i,\bm{x}_i)+
\frac{1}{2}(\sigma_j'-t)^2k(\bm{x}_j,\bm{x}_j)+\nonumber\\
&(\sigma_i'+t)(\sigma_j'-t)k(\bm{x}_i,\bm{x}_j)+h_{\bm{z}}(t)
\end{align}}
where $h_{\bm{z}}$ is a function defined in the following way:
{\small\begin{equation}
h_{\bm{z}}(t){=}\left\{{\arraycolsep=0.5pt\begin{array}{ll}
(e_1{-}1)(\sigma_i'{+}t){+}(e_2{-}1)(\sigma_j'{-}t),&(\sigma_i{,}\sigma_j){\in}R_1,\\
(e_1{-}1)(\sigma_i'{+}t){+}(e_2{+}1)(\sigma_j'{-}t){-}c_2,&(\sigma_i{,}\sigma_j){\in}R_2,\\
(e_1{+}1)(\sigma_i'{+}t){+}(e_2{+}1)(\sigma_j'{-}t){-}2c_2,&(\sigma_i{,}\sigma_j){\in}R_3,\\
(e_1{+}1)(\sigma_i'{+}t){+}(e_2{-}1)(\sigma_j'{-}t){-}c_2,&(\sigma_i{,}\sigma_j){\in}R_4,\\
\end{array}}\right.\nonumber
\end{equation}}
Note that $\frac{d^2\phi(t)}{dt^2}=k(\bm{x}_i,\bm{x}_i)+k(\bm{x}_j,\bm{x}_j)-2k(\bm{x}_i,\bm{x}_j)\geq 0$ ($k$ is a Mercer kernel), meaning that~(\ref{eq:monoobj}) is convex. 

If $(\sigma_i^*,\sigma_j^*)\in\text{int }R_1$, then $(\sigma_i^*,\sigma_j^*)$ is the minimum and $\frac{d\phi(t^*)}{dt}=0$. Since $\frac{d\phi(t^*)}{dt}=f_{\bm{z}^*}(\bm{x}_j)-f_{\bm{z}^*}(\bm{x}_i)=0$, the first and the second conditions in~(\ref{eq:violating}), which are the only possibilities to have a violating pair, are not satisfied. Therefore, $(\bm{x}_i,\bm{x}_j)$ is not violating at point $\bm{z}^*$, but it is violating at $\bm{z}'$, implying that $\bm{z}^*\neq\bm{z}'$. The same situation holds for $(\sigma_i^*,\sigma_j^*)\in\text{int }R_3$ and this proves statement (c).\footnote{In this case, the admissible conditions for violation are the second and the third conditions in~(\ref{eq:violating}).} Statements (d) and (e) can be proven in the same way, considering that the admissible conditions to have a violating pair are the first, the fourth and the fifth conditions for the former case and the second, the third and the sixth ones for the latter case.

If $(\sigma_i^*{,}\sigma_j^*)\in BE$, there are two possibilities to compute the derivative depending on the position of $(\sigma_i'{,}\sigma_j')$, namely approaching $(\sigma_i^*{,}\sigma_j^*)$ from the bottom or from the top of the constraint line. In the first case, the derivative is identified by $\frac{d\phi(t^*)}{dt^-}$, while in the second case by $\frac{d\phi(t^*)}{dt^+}$. Since $(\sigma_i^*{,}\sigma_j^*)$ is the minimum and due to the convexity of function $\phi(t)$, $\frac{d\phi(t^*)}{dt^-}\geq 0$ and $\frac{d\phi(t^*)}{dt^+}\leq 0$. Furthermore, it is easy to verify that $\frac{d\phi(t^*)}{dt^-}=f_{\bm{z}^*}(\bm{x}_j){-}f_{\bm{z}^*}(\bm{x}_i)$ and $\frac{d\phi(t^*)}{dt^+}=f_{\bm{z}^*}(\bm{x}_j){-}f_{\bm{z}^*}(\bm{x}_i)-2$. By combining these results, we obtain that $0\leq f_{\bm{z}^*}(\bm{x}_j){-}f_{\bm{z}^*}(\bm{x}_i)\leq 2$. This, compared with the first condition in~(\ref{eq:violating}), guarantees that $(\bm{x}_i,\bm{x}_j)$ is not a violating pair at $\bm{z}^*$ and therefore that $\bm{z}^*\neq\bm{z}'$. The same strategy can be applied to derive statements (g)-(o).

For the sake of notation compactness, we use $\phi'(t)$ to identify both the classical and the directional derivatives of $\phi(t)$, viz. $\frac{d\phi(t)}{dt}$, $\frac{d\phi(t^*)}{dt^-}$ and $\frac{d\phi(t^*)}{dt^+}$, respectively.  Therefore, it is possible to show that $\phi(t)=\phi(0)+\phi'(0)t+\frac{\phi''(0)}{2}t^2$. Furthermore, due to the convexity of $\phi(t)$, we have that
{\small\begin{align}
\label{eq:derivative}
\phi'(0)<0\Rightarrow t_q\geq t^*>0,\nonumber\\
\phi'(0)>0\Rightarrow t_q\leq t^*<0,
\end{align}}
where $t_q=-\frac{\phi'(0)}{\phi''(0)}$ is the unconstrained minimum of $\phi(t)$. From all these considerations, we can derive the following relation:
\vspace{-0.1cm}
{\small\begin{equation}
\label{eq:inequality}
\phi(t^*)\leq\phi(0)+\frac{\phi'(0)}{2}t^*
\end{equation}}
In fact, if $\phi''(0)=0$, then~(\ref{eq:inequality}) trivially holds. If $\phi''(0)>0$, then 
\vspace{-0.1cm}
{\small\begin{equation}
\label{eq:inequality2}
\phi(t^*)-\phi(0)=\frac{\phi'(0)}{2}t^*\bigg(\frac{2t_q-t^*}{t_q}\bigg)\leq\frac{\phi'(0)}{2}t^*
\end{equation}} 
where the last inequality of~(\ref{eq:inequality2}) is valid because $\Big(\frac{2t_q-t^*}{t_q}\Big)\geq 1$, by simply applying~(\ref{eq:derivative}).

Note also that~(\ref{eq:line}) can be used to derive the following result, namely:
\begin{equation}
\label{eq:distance}
\|\bm{\sigma}'-\bm{\sigma}^*\|_2=|t^*|\sqrt[]{2}
\end{equation}
By combining~(\ref{eq:inequality2}) and~(\ref{eq:distance}) and considering that conditions~(\ref{eq:violating}) can be compactly rewritten as $|\phi'(0)|>\tau$, we obtain that
\vspace{-0.1cm}
{\small\begin{align}
\label{eq:final}
\phi(0)-\phi(t^*)&\geq-\frac{\phi'(0)}{2}t^*=\frac{|\phi'(0)|}{2}|t^*|\nonumber\\
&>\frac{\tau}{2}|t^*|=\frac{\tau}{2\sqrt[]{2}}\|\bm{\sigma}'-\bm{\sigma}^*\|_2,
\end{align}}
Finally, statement (p) is obtained from~(\ref{eq:final}), by taking into account that $\phi(t^*)=F(\bm{\sigma}^*,\bm{\delta}^*)$ and $\phi(0)=F(\bm{\sigma}',\bm{\delta}')$.
\end{IEEEproof}

Lemma~\ref{th:minimization} states that each iteration of Algorithm~\ref{alg:usmo} generates a solution that is $\tau-$optimal for the indices in the working set $S$.

The convergence of USMO to a $\tau-$optimal solution can be proven by contradiction by assuming that the algorithm proceeds indefinitely. This is equivalent to assume that $(\bm{x}_{i^k},\bm{x}_{j^k})$ is violating $\forall k\geq0$, where $(i^k,j^k)$ represents the pair of indices selected at iteration $k$.

Since $\{F(\bm{\sigma}^k,\bm{\delta}^k)\}$ is a decreasing sequence (due to the fact that $\bm{z}^k\neq\bm{z}^{k+1}$ $\forall k\geq0$\footnote{Statement (a) of Lemma~\ref{th:minimization}.} and that the algorithm minimizes the objective function at each iteration) and bounded below (due to the existence of an unknown global optimum), it is convergent. By exploiting this fact and by considering that {\small$\frac{2\sqrt[]{2}}{\tau}[F(\bm{\sigma}^k,\bm{\delta}^k){-}F(\bm{\sigma}^{k+l},\bm{\delta}^{k+l})]>\|\bm{\sigma}^k{-}\bm{\sigma}^{k+l}\|_2,\:\forall k,l\geq 0$}, which can be obtained from~(p) of Lemma~\ref{th:minimization} by applying $l$ times the triangle inequality, it is possible to conclude that $\{\bm{\sigma}^k\}$ is a Cauchy sequence. Therefore, since the sequence lies also in a closed feasible set, it is convergent. In other words, we have that $\bm{\sigma}^k\rightarrow\bar{\bm{\sigma}}$ for $k\rightarrow\infty$, meaning that Algorithm~\ref{alg:usmo} produces \textbf{a convergent sequence of points}. Now, it is important to understand if this sequence converges to a $\tau-$optimal solution. 

First of all, let us define the set of indices that are encountered/selected by the algorithm infinitely many times:
{\small\begin{equation}
\label{eq:index}
I_\infty=\{(\mu,\nu):\exists\{k_t\}\subset\{k\},(i^{k_t},j^{k_t})=(\mu,\nu), \forall t\in\mathbb{N}\}
\end{equation}}
$\{k_t\}$ is therefore a subsequence of $\{k\}$. It is also important to mention that since the number of iterations is infinite and the number of samples is finite, $I_\infty$ cannot be an empty set. Based on this consideration, we define $\bm{v}_{\mu\nu}$ as the vector, whose elements are the entries at position $\mu$ and $\nu$ of a general vector $\bm{v}$, and provide the following lemma.
\begin{lemma}
\label{th:ball}
Assume $(\mu,\nu)\in I_\infty$ and let $\{k_t\}$ be the sequence of indices  for which $(i^{k_t},j^{k_t})=(\mu,\nu)$. Then, 
\begin{itemize}
\item[(a)] $\forall \epsilon>0$, $\exists \hat{t}>0$: $\forall t\geq\hat{t}$, $\|\bm{\sigma}_{\mu\nu}^{k_t}-\bar{\bm{\sigma}}_{\mu\nu}\|<\epsilon$ and $\|\bm{\sigma}_{\mu\nu}^{k_t+1}-\bar{\bm{\sigma}}_{\mu\nu}\|<\epsilon$
\item[(b)] {\small$f_{\bm{\sigma}^{k_t}}(\bm{x}_\mu){-}f_{\bm{\sigma}^{k_t}}(\bm{x}_\nu)>\tau\Rightarrow f_{\bar{\bm{\sigma}}}(\bm{x}_\mu){-} f_{\bar{\bm{\sigma}}}(\bm{x}_\nu)\geq\tau$} 
\item[(c)] {\small$f_{\bm{\sigma}^{k_t}}(\bm{x}_\mu){-}f_{\bm{\sigma}^{k_t}}(\bm{x}_\nu)<{-}\tau\Rightarrow f_{\bar{\bm{\sigma}}}(\bm{x}_\mu){-} f_{\bar{\bm{\sigma}}}(\bm{x}_\nu)\leq{-}\tau$} 
\item[(d)] {\small$f_{\bm{\sigma}^{k_t}}(\bm{x}_\mu){-}f_{\bm{\sigma}^{k_t}}(\bm{x}_\nu)>\tau{-}2\Rightarrow f_{\bar{\bm{\sigma}}}(\bm{x}_\mu){-} f_{\bar{\bm{\sigma}}}(\bm{x}_\nu)\geq\tau{-}2$} 
\item[(e)] {\small$f_{\bm{\sigma}^{k_t}}(\bm{x}_\mu){-}f_{\bm{\sigma}^{k_t}}(\bm{x}_\nu){<}{-}\tau{+}2\Rightarrow f_{\bar{\bm{\sigma}}}(\bm{x}_\mu){-} f_{\bar{\bm{\sigma}}}(\bm{x}_\nu){\leq}{-}\tau{+}2$} 
\end{itemize}
where $f_{\bm{\sigma}^{k_t}}$, $f_{\bar{\bm{\sigma}}}$ represent the target function with coefficients $\alpha_i$ computed according to~(\ref{eq:alpha}) using $\bm{\sigma}^{k_t}$ and $\bar{\bm{\sigma}}$, respectively.
\end{lemma}
\begin{IEEEproof}
Since $\{\bm{\sigma}^k\}$ is convergent and $\{k_t\}$, $\{k_t+1\}$ are subsequences of $\{k\}$, $\{\bm{\sigma}^{k_t}\}$ and $\{\bm{\sigma}^{k_t+1}\}$ are also convergent sequences. In other words, $\exists \hat{t}{>}0$ such that $\|\bm{\sigma}^{k_t}-\bar{\bm{\sigma}}\|{<}\epsilon$ and $\|\bm{\sigma}^{k_t+1}-\bar{\bm{\sigma}}\|{<}\epsilon$. Furthemore, $\|\bm{\sigma}^{k_t}-\bar{\bm{\sigma}}\|{\geq}\|\bm{\sigma}_{\mu\nu}^{k_t}-\bar{\bm{\sigma}}_{\mu\nu}\|$ and $\|\bm{\sigma}^{k_t+1}-\bar{\bm{\sigma}}\|{\geq}\|\bm{\sigma}_{\mu\nu}^{k_t+1}-\bar{\bm{\sigma}}_{\mu\nu}\|$. By combining these two results, we obtain statement (a).

Concerning statement (b), we have that {\small$f_{\bm{\sigma}^{k_t}}(\bm{x}_\mu){-}f_{\bm{\sigma}^{k_t}}(\bm{x}_\nu){>}\tau$}. Furthermore, from convergence of $\{\bm{\sigma}^{k_t}\}$ and continuity of $f$, we obtain that $\forall\epsilon{>}0$, $\exists\tilde{t}{\geq}\hat{t}$: $\forall t{\geq}\tilde{t}$, {\small$-\epsilon{\leq}f_{\bm{\sigma}^{k_t}}(\bm{x}_\mu){-}f_{\bar{\bm{\sigma}}}(\bm{x}_\mu){\leq}\epsilon$} and {\small$-\epsilon{\leq}f_{\bm{\sigma}^{k_t}}(\bm{x}_\nu){-}f_{\bar{\bm{\sigma}}}(\bm{x}_\nu){\leq}\epsilon$}, meaning that both $\{f_{\bm{\sigma}^{k_t}}(\bm{x}_\mu)\}$ and $\{f_{\bm{\sigma}^{k_t}}(\bm{x}_\nu)\}$ are convergent. Therefore, {\small$f_{\bm{\sigma}^{k_t}}(\bm{x}_\mu){-}f_{\bm{\sigma}^{k_t}}(\bm{x}_\nu){>}\tau$} can be rewritten as
{\small\begin{equation}
f_{\bm{\sigma}^{k_t}}(\bm{x}_\mu){-}f_{\bm{\sigma}^{k_t}}(\bm{x}_\nu){+}f_{\bar{\bm{\sigma}}}(\bm{x}_\mu){-}f_{\bar{\bm{\sigma}}}(\bm{x}_\mu){+}f_{\bar{\bm{\sigma}}}(\bm{x}_\nu){-}f_{\bar{\bm{\sigma}}}(\bm{x}_\nu){>}\tau\nonumber
\end{equation}}
and by applying the information about the convergence of both $\{f_{\bm{\sigma}^{k_t}}(\bm{x}_\mu)\}$ and $\{f_{\bm{\sigma}^{k_t}}(\bm{x}_\nu)\}$, we get that 
{\small\begin{equation}
f_{\bar{\bm{\sigma}}}(\bm{x}_\mu)-f_{\bar{\bm{\sigma}}}(\bm{x}_\nu)>\tau-2\epsilon\nonumber
\end{equation}}
which is valid $\forall\epsilon>0$ and therefore proves statement (b). All other statements, namely (c)-(e), can be proven using the same approach.
\end{IEEEproof}
\begin{figure}[!t]
\centering
\vspace{-1em}
\includegraphics[width=0.3\textwidth]{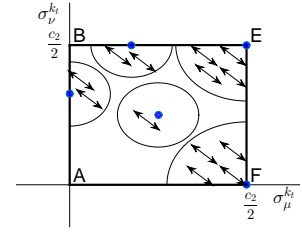}
\vspace{-3em}
\caption{Example of transitions performed by a minimization step of Algorithm~\ref{alg:usmo} for different locations of $\bar{\bm{\sigma}}_{\mu,\nu}$ (highlighted by blue points) and for sufficiently large number of iterations.}
\label{fig:transitions}
\end{figure}
Lemma~\ref{th:ball} states some conditions about the final target function and also states that the sequence output by Algorithm~\ref{alg:usmo}, after a sufficiently large number of iterations, is enclosed in a ball centered at $\bar{\bm{\sigma}}$. This aspect is shown in Figure~\ref{fig:transitions} for $R_1$ and for different possible locations of $\bar{\bm{\sigma}}_{\mu,\nu}$. The same picture shows also the possible transitions that may happen at each iteration. In particular, we see that for $\bar{\bm{\sigma}}_{\mu,\nu}$ lying on corners and edges, different kinds of transitions exist. In fact, we find transitions from border to border, transitions from border to inner points and viceversa, and transitions from inner points to inner points. These are indetified as $bd\rightarrow bd$, $bd\rightarrow int$, $int\rightarrow bd$ and $int\rightarrow int$, respectively. Note that for $\bar{\bm{\sigma}}_{\mu,\nu}$ not lying on borders, $int\rightarrow int$ is the only available kind of transition. Based on these considerations, it is possible to prove the following lemma.
\begin{lemma}
\label{th:transitions}
Let $(\mu,\nu)$, $\{k_t\}$, $\hat{t}$ and $\epsilon$ be defined according to Lemma~\ref{th:ball}. Then, $\exists\bar{t}\geq\hat{t}$ such that $\forall t\geq\bar{t}$ and for sequence $\{k_t\}$ the only allowed transitions are $int\rightarrow bd$ and $bd\rightarrow bd$.
\end{lemma}
\vspace{-0.2cm}
\begin{IEEEproof}
Consider region $R_1$ and $(\bar{\sigma}_\mu,\bar{\sigma}_\nu)\in\text{int }R_1$. Then, the only admissible type of transitions for this case is $int{\rightarrow}int$. Therefore, based on statement (c) of Lemma~\ref{th:minimization} (and thanks also to statement (a) of Lemma~\ref{th:ball}), we obtain that $\forall t{\geq}\bar{t}$, $f_{\bm{\sigma}^{k_t+1}}(\bm{x}_\mu){-}f_{\bm{\sigma}^{k_t+1}}(\bm{x}_\nu){=}0$. By exploiting this fact, the continuity of $f$ and the convergence of $\{\bm{\sigma}^{k_t+1}\}$, it is possible to show that 
{\small\begin{equation}
\label{eq:fin}
f_{\bar{\bm{\sigma}}}(\bm{x}_\mu){-}f_{\bar{\bm{\sigma}}}(\bm{x}_\nu)=0
\end{equation}}
Furthermore, since $(\bm{x}_\mu,\bm{x}_\nu)$ is a violating pair at all iterations and $\forall t{\geq}\bar{t}$, $\bm{\sigma}_{\mu\nu}^{k_t}\in\text{int }R_1$ (due to statement (a) of Lemma~\ref{th:ball}), $(\bm{x}_\mu,\bm{x}_\nu)$ has to satisfy conditions (b) or (c) of Lemma~\ref{th:ball}. These conditions are in contradiction with~(\ref{eq:fin}), meaning that the $int{\rightarrow}int$ transition is not allowed in this case.

Consider now region $R_1$ and $(\bar{\sigma}_\mu,\bar{\sigma}_\nu)\in E$, or equivalently $(\bar{\sigma}_\mu,\bar{\sigma}_\nu)\in A$. This time, the potential transitions are $bd\rightarrow bd$, $bd\rightarrow int$, $int\rightarrow bd$ and $int\rightarrow int$. Nevertheless, it is always possible to define a subsequence containing only either $int\rightarrow int$ or $bd\rightarrow int$ and obtain therefore conclusions similar to the previous case. In fact, both  $int{\rightarrow}int$ and $bd\rightarrow int$ are not allowed transitions.

The same results can be obtained in a similar way for other edges, corners of $R_1$ as well as for points in $R_3$, upon selection of the proper conditions in Lemma~\ref{th:minimization}.

Consider now region $R_2$ and $(\bar{\sigma}_\mu,\bar{\sigma}_\nu)\in\text{int }R_2$. The only admissible transition in this case is $int\rightarrow int$. From statement (d) of Lemma~\ref{th:minimization}, we have that $\forall t{\geq}\bar{t}$, $f_{\bm{\sigma}^{k_t+1}}(\bm{x}_\mu){-}f_{\bm{\sigma}^{k_t+1}}(\bm{x}_\nu){=}-2$ and, from the continuity of $f$ and the convergence of $\{\bm{\sigma}^{k_t+1}\}$, it is possible to show that
{\small\begin{equation}
\label{eq:fin2}
f_{\bar{\bm{\sigma}}}(\bm{x}_\mu){-}f_{\bar{\bm{\sigma}}}(\bm{x}_\nu)=-2
\end{equation}}
Furthermore, since $(\bm{x}_\mu,\bm{x}_\nu)$ is a violating pair at all iterations and $\forall t{\geq}\bar{t}$, $\bm{\sigma}_{\mu\nu}^{k_t}\in\text{int }R_2$, $(\bm{x}_\mu,\bm{x}_\nu)$ has to satisfy conditions (b) or (d) of Lemma~\ref{th:ball}. These conditions are in contradiction with~(\ref{eq:fin2}), meaning that the $int{\rightarrow}int$ transition is not valid.

For all corners and edges of $R_2$, as well as for all points in $R_4$, it is possible to show that $int{\rightarrow}int$ and $bd\rightarrow int$ are not valid transitions. The proof is similar to the previous cases. Therefore, the only admissible transitions after a sufficiently large number of iterations are $int\rightarrow bd$ and $bd\rightarrow bd$.
\end{IEEEproof}
It is interesting to note that each transition $int\rightarrow bd$ increases the number of components of $\bm{\sigma}$ belonging to borders of the four regions, by one or two, while each transition $bd\rightarrow bd$ leaves it unchanged. Since this number is bounded by $n$, transition $int\rightarrow bd$ cannot appear infinitely many times. Therefore, $\exists t^*\geq\bar{t}$, $\forall t\geq t^*$, $bd\rightarrow bd$ is the only valid transition. 

Note that $bd\rightarrow bd$ may happen only when $(\bar{\sigma}_\mu,\bar{\sigma}_\nu)$ is located at some specific corners of the feasible region, namely corners $A$ or $E$ for region $R_1$, corners $B$ or $C$ for region $R_2$, corners $E$ or $I$ for region $R_3$ and corners $F$ or $H$ for region $R_4$. For all cases, it is possible to define a subsequence that goes only from a vertical to a horizontal border and a subsequence that goes only from a horizontal to a vertical border. Without loss of generality, we can consider a specific case, namely $(\bar{\sigma}_\mu,\bar{\sigma}_\nu)\in A$. Note that for the first subsequence, {\small$f_{\bm{\sigma}^{k_t+1}}(\bm{x}_\mu){-}f_{\bm{\sigma}^{k_t+1}}(\bm{x}_\nu){<}-\tau$}, since $(\bm{x}_\mu,\bm{x}_\nu)$ has to be a violating pair in order not to stop the iterations and therefore, from statement (c) of Lemma~\ref{th:ball}, {\small$f_{\bar{\bm{\sigma}}}(\bm{x}_\mu){-}f_{\bar{\bm{\sigma}}}(\bm{x}_\nu){<}-\tau$}. For the second subsequence, {\small$f_{\bm{\sigma}^{k_t+1}}(\bm{x}_\mu){-}f_{\bm{\sigma}^{k_t+1}}(\bm{x}_\nu){>}\tau$} and consequently {\small$f_{\bar{\bm{\sigma}}}(\bm{x}_\mu){-}f_{\bar{\bm{\sigma}}}(\bm{x}_\nu){>}\tau$}. This leads to a contradiction which holds $\forall(\mu,\nu)\in I_\infty$. Therefore, the assumption that Algorithm~\ref{alg:usmo} proceeds indefinitely is not verified. In other words, there exists an iteration at which the algorithm stops because a $\tau-$optimal solution is obtained.

%% file: results.tex
\vspace{-0.3cm}
\begin{table}[h!]
\caption{Characteristics of data sets.}
\label{table:datasets}
\vspace{-1em}
\centering
\resizebox{0.3\textwidth}{!}{\begin{tabular}{ccc}
\hline
Data&\# Instances&\# Features\\
\hline
\textit{Australian} & 690 & 42 \\
\textit{Clean1} & 476 & 166 \\
\textit{Diabetes} & 768 & 8 \\
\textit{Heart} & 270 & 9 \\
\textit{Heart-statlog} & 270 & 13 \\
\textit{House} & 232 & 16 \\
\textit{House-votes} & 435 & 16 \\
\textit{Ionosphere} & 351 & 33 \\
\textit{Isolet} & 600 & 51 \\
\textit{Krvskp} & 3196 & 36 \\
\textit{Liverdisorders} & 345 & 6 \\
\textit{Spectf} & 349 & 44 \\
\hline
\textit{Bank-marketing} & 28000 & 20 \\
\textit{Adult} & 32562 & 123 \\
\textit{Statlog (shuttle)} & 43500 & 9 \\
\textit{Mnist} & 60000 & 784 \\
\textit{Poker-hand} & 1000000 & 10 \\
\hline
\end{tabular}}
\end{table}
In this section, comprehensive evaluations are presented to demonstrate the effectiveness of the 
proposed approach. USMO is compared with~\cite{plessis2015convex} and~\cite{plessis2014analysis}. 
The three methods have been implemented in MATLAB, to ensure fair comparison. 
\footnote{Code available at \url{https://github.com/emsansone/USMO}.} The method in~\cite{plessis2015convex} solves 
problem~(\ref{eq:doubledual}) using the MATLAB built-in function \textit{quadprog}, combined 
with the second-order primal-dual interior point algorithm~\cite{mehrotra1992implementation}, 
while the method in~\cite{plessis2014analysis} solves problem~(\ref{eq:puopt}) with the ramp 
loss function using the \textit{quadprog} function combined with the concave-convex 
procedure~\cite{yuille2002concave}. Experiments were run on a PC with 16 2.40 GHz cores and 
64GB RAM.
\begin{table*}[h!]
\vspace{-0.1cm}
\caption{Comparative results (F-measure) on different small-scale datasets and on different values of 
hyperparameters using the linear kernel. 20\% of positive examples are labeled, while the remaining 
are unlabeled.}
\label{table:results_f_linear}
\vspace{-1.5em}
\centering
\resizebox{0.7\textwidth}{!}{\begin{tabular}{c|cccc|cccc|cccc|cccc}
\hline
&
\multicolumn{4}{c|}{$\lambda=0.0001$} &
\multicolumn{4}{c|}{$\lambda=0.001$} &
\multicolumn{4}{c|}{$\lambda=0.01$} &
\multicolumn{4}{c}{$\lambda=0.1$}\\
Data &
Init* & \cite{plessis2014analysis} & \cite{plessis2015convex} & USMO & 
Init* & \cite{plessis2014analysis} & \cite{plessis2015convex} & USMO &
Init* & \cite{plessis2014analysis} & \cite{plessis2015convex} & USMO &
Init* & \cite{plessis2014analysis} & \cite{plessis2015convex} & USMO \\
\hline

\textit{Australian} & 61.8 & 61.5 & 67.9 & 68.3 & 61.8 & 68.8 & 67.7 & 67.6 & 61.8 & 63.4 & 69.0 & 69.3 & 61.8 & 58.5 & 70.0 & 70.2 \\

\textit{Clean1} & 60.6 & 81.6 & 70.5 & 70.2 & 63.8 & 76.2 & 65.5 & 65.6 & 64.6 & 81.9 & 73.3 & 73.0 & 66.2 & 80.4 & 77.9 & 75.6 \\

\textit{Diabetes} & 41.8 & 73.3 & 70.0 & 70.1 & 41.4 & 71.5 & 71.2 & 71.1 & 37.6 & 77.9 & 78.0 & 79.3 & 6.5 & 82.3 & 82.3 & 82.3 \\ 
       
\textit{Heart} & 46.8 & 60.8 & 60.1 & 59.4 & 49.0 & 63.4 & 60.7 & 60.0 & 57.2 & 64.9 & 66.0 & 66.0 & 75.6 & 75.6 & 75.6 & 75.6 \\
       
\textit{Heart-statlog} & 63.3 & 65.1 & 54.5 & 54.5 & 63.8 & 58.1 & 55.4 & 55.2 & 66.9 & 61.2 & 53.8 & 53.8 & 75.6 & 67.5 & 61.7 & 60.2 \\
       
\textit{House} & 49.0 & 57.9 & 59.9 & 59.9 & 49.8 & 66.7 & 59.0 & 59.0 & 49.8 & 57.9 & 57.4 & 56.8 & 58.4 & 51.8 & 64.6 & 64.0 \\
       
\textit{House-votes} & 59.3 & 51.6 & 59.6 & 59.6 & 60.0 & 56.6 & 60.0 & 59.9 & 66.9 & 59.3 & 57.1 & 57.2 & 71.8 & 51.3 & 61.7 & 62.0 \\
       
\textit{Ionosphere} & 19.3 & 59.9 & 65.1 & 65.1 & 19.3 & 74.9 & 71.5 & 72.0 & 19.3 & 71.9 & 72.6 & 73.7 & 22.3 & 75.5 & 75.2 & 75.2 \\
       
\textit{Isolet} & 98.3 & 77.1 & 91.8 & 93.7 & 98.1 & 78.0 & 93.3 & 93.3 & 98.1 & 81.1 & 94.7 & 94.7 & 98.0 & 86.8 & 95.5 & 95.5 \\
       
\textit{Krvskp} & 57.0 & 78.4 & 81.1 & 81.1 & 57.3 & 78.7 & 79.6 & 79.9 & 61.1 & 75.8 & 82.9 & 82.9 & 68.8 & 75.0 & 80.6 & 80.6 \\
       
\textit{Liverdisorders} & 54.5 & 56.0 & 55.7 & 56.0 & 60.0 & 58.1 & 63.9 & 63.4 & 69.0 & 68.8 & 68.8 & 68.8 & 68.8 & 68.8 & 68.8 & 68.8 \\
       
\textit{Spectf} & 56.4 & 58.0 & 73.5 & 73.5 & 60.4 & 66.3 & 72.9 & 72.3 & 58.1 & 79.9 & 80.6 & 80.6 & 79.2 & 81.1 & 81.1 & 81.1 \\

\hline
Avg. & 55.7$\pm$18.1 & 65.1$\pm$10.0 & \textbf{67.5$\pm$10.9} & \textbf{67.6$\pm$11.3} & 57.0$\pm$18.1 & \textbf{68.1$\pm$7.9} & \textbf{68.4$\pm$10.4} & \textbf{68.3$\pm$10.6} & 59.2$\pm$18.9 & 70.3$\pm$8.9 & \textbf{71.2$\pm$11.9} & \textbf{71.3$\pm$12.1} & 62.7$\pm$24.9 & 71.2$\pm$11.9 & \textbf{74.6$\pm$9.9} & \textbf{74.3$\pm$10.0} \\
\hline
\multicolumn{17}{l}{* Results obtained using only our proposed initialization}
\end{tabular}}
\end{table*}
\begin{table*}[h!]
\vspace{-0.1cm}
\caption{Comparative results (F-measure) on different small-scale datasets and on different 
values of hyperparameters using the Gaussian kernel (scale parameter equal to 1). 20\% of positive 
examples are labeled, while the remaining are unlabeled.}
\label{table:results_f_gaussian}
\vspace{-1.5em}
\centering
\resizebox{0.7\textwidth}{!}{\begin{tabular}{c|cccc|cccc|cccc|cccc}
\hline
&
\multicolumn{4}{c|}{$\lambda=0.0001$} &
\multicolumn{4}{c|}{$\lambda=0.001$} &
\multicolumn{4}{c|}{$\lambda=0.01$} &
\multicolumn{4}{c}{$\lambda=0.1$}\\
Data &
Init* & \cite{plessis2014analysis} & \cite{plessis2015convex} & USMO & 
Init* & \cite{plessis2014analysis} & \cite{plessis2015convex} & USMO &
Init* & \cite{plessis2014analysis} & \cite{plessis2015convex} & USMO &
Init* & \cite{plessis2014analysis} & \cite{plessis2015convex} & USMO \\
\hline

\textit{Australian} & 67.0 & 64.2 & 64.2 & 64.7 & 67.0 & 64.2 & 70.6 & 70.6 & 67.0 & 57.2 & 59.4 & 59.4 & 67.0 & 0.0 & 0.0 & 0.0 \\    
       
\textit{Clean1} & 28.7 & 80.1 & 77.6 & 77.6 & 44.0 & 80.6 & 81.7 & 81.7 & 78.3 & 78.0 & 76.6 & 76.6 & 76.4 & 76.4 & 76.4 & 76.4 \\
       
\textit{Diabetes} & 58.8 & 74.1 & 70.0 & 70.0 & 58.4 & 71.3 & 71.2 & 70.6 & 52.6 & 80.7 & 80.1 & 80.1 & 0.0 & 82.3 & 82.3 & 82.3 \\ 
       
\textit{Heart} & 46.5 & 66.7 & 58.9 & 58.9 & 47.7 & 64.1 & 59.8 & 59.8 & 63.8 & 70.3 & 70.9 & 70.9 & 75.6 & 75.6 & 75.6 & 75.6 \\

\textit{Heart-statlog} & 30.7 & 70.3 & 53.5 & 53.5 & 31.5 & 65.1 & 56.7 & 56.7 & 49.6 & 60.6 & 56.4 & 56.4 & 75.6 & 75.6 & 75.6 & 75.6 \\

\textit{House} & 38.5 & 63.9 & 56.5 & 56.5 & 37.1 & 68.8 & 61.1 & 60.1 & 28.2 & 59.1 & 56.8 & 54.2 & 74.0 & 74.0 & 74.0 & 74.0 \\

\textit{House-votes} & 64.2 & 29.7 & 61.2 & 61.2 & 64.1 & 54.3 & 59.2 & 59.4 & 67.4 & 61.1 & 57.5 & 57.5 & 71.8 & 71.8 & 71.8 & 71.8 \\
       
\textit{Ionosphere} & 55.6 & 52.6 & 65.3 & 65.3 & 56.3 & 68.2 & 74.3 & 74.1 & 58.4 & 58.7 & 65.7 & 65.7 & 72.7 & 74.2 & 74.2 & 74.2 \\

\textit{Isolet} & 0.0 & 73.3 & 75.3 & 75.3 & 0.0 & 73.8 & 75.7 & 75.7 & 0.0 & 74.1 & 76.4 & 76.4 & 0.0 & 73.3 & 75.9 & 75.9 \\    

\textit{Krvskp} & 52.0 & 73.9 & 82.4 & 82.5 & 59.4 & 77.2 & 84.3 & 84.1 & 44.2 & 73.2 & 73.2 & 73.2 & 73.2 & 73.2 & 73.2 & 73.2 \\

\textit{Liverdisorders} & 44.3 & 60.6 & 57.0 & 56.2 & 49.5 & 63.7 & 62.2 & 62.5 & 68.8 & 68.8 & 68.8 & 68.8 & 68.8 & 68.8 & 68.8 & 68.8 \\

\textit{Spectf} & 61.5 & 38.0 & 74.0 & 74.0 & 80.0 & 62.4 & 72.9 & 72.1 & 69.0 & 81.1 & 81.1 & 81.1 & 81.1 & 81.1 & 81.1 & 81.1 \\

\hline
Avg. & 45.7$\pm$19.1 & 62.3$\pm$15.2 & \textbf{66.3$\pm$9.4} & \textbf{66.3$\pm$9.5} & 49.6$\pm$20.5 & 67.8$\pm$7.2 & \textbf{69.2$\pm$9.2} & \textbf{68.9$\pm$9.2} & 53.9$\pm$21.7 & \textbf{68.6$\pm$9.0} & \textbf{68.6$\pm$9.3} & \textbf{68.4$\pm$9.6} & 61.4$\pm$28.9 & \textbf{68.9$\pm$22.0} & \textbf{69.1$\pm$22.1} & \textbf{69.1$\pm$22.1} \\
\hline
\multicolumn{17}{l}{* Results obtained using only our proposed initialization}
\end{tabular}}
\end{table*}
A collection of 17 real-world datasets from the UCI repository was used, 12 of which contain few 
hundreds/thousands of samples, while the remaining 5 are significantly bigger. 
Table~\ref{table:datasets} shows some of their statistics.

Since USMO and~\cite{plessis2015convex} solve the same optimization problem, we first verify 
that \textbf{both achieve the same solution}. We consider the F-measure in a transductive 
setting, to assess the generalization performance on all small-scale datasets and under different 
configurations of hyper-parameters and kernel functions. In particular, we consider different 
values of $\lambda$, viz. {0.0001, 0.001, 0.01, 0.1}, using linear and Gaussian kernels.
\footnote{The positive class prior $\pi$ is set to the class proportion in the training data sets. 
Methods like~\cite{lee2003learning,blanchard2010semi,du2015class} can be used to estimate it.} 
In these experiments, only 20\% of positive samples are labeled. 
Tables~\ref{table:results_f_linear}-\ref{table:results_f_gaussian} show the results for linear 
and Gaussian kernels, respectively. Both algorithms achieve almost identical performance, with 
small differences due to numerical approximations. This fact confirms the theory proven in 
Section~\ref{sec:theory}, according to which USMO is guaranteed to converge to the same value of 
objective function obtained by~\cite{plessis2015convex}. 
Note that ramp loss~\cite{plessis2014analysis} never achieves the best average performance. Furthermore, it is influenced by the starting point due to a non-convex objective function, thus making double Hinge loss preferable in practical applications.

\begin{figure*}[h!]
\centering
\subfloat[Statlog 1 vs. all]{\includegraphics[width=0.16\linewidth]{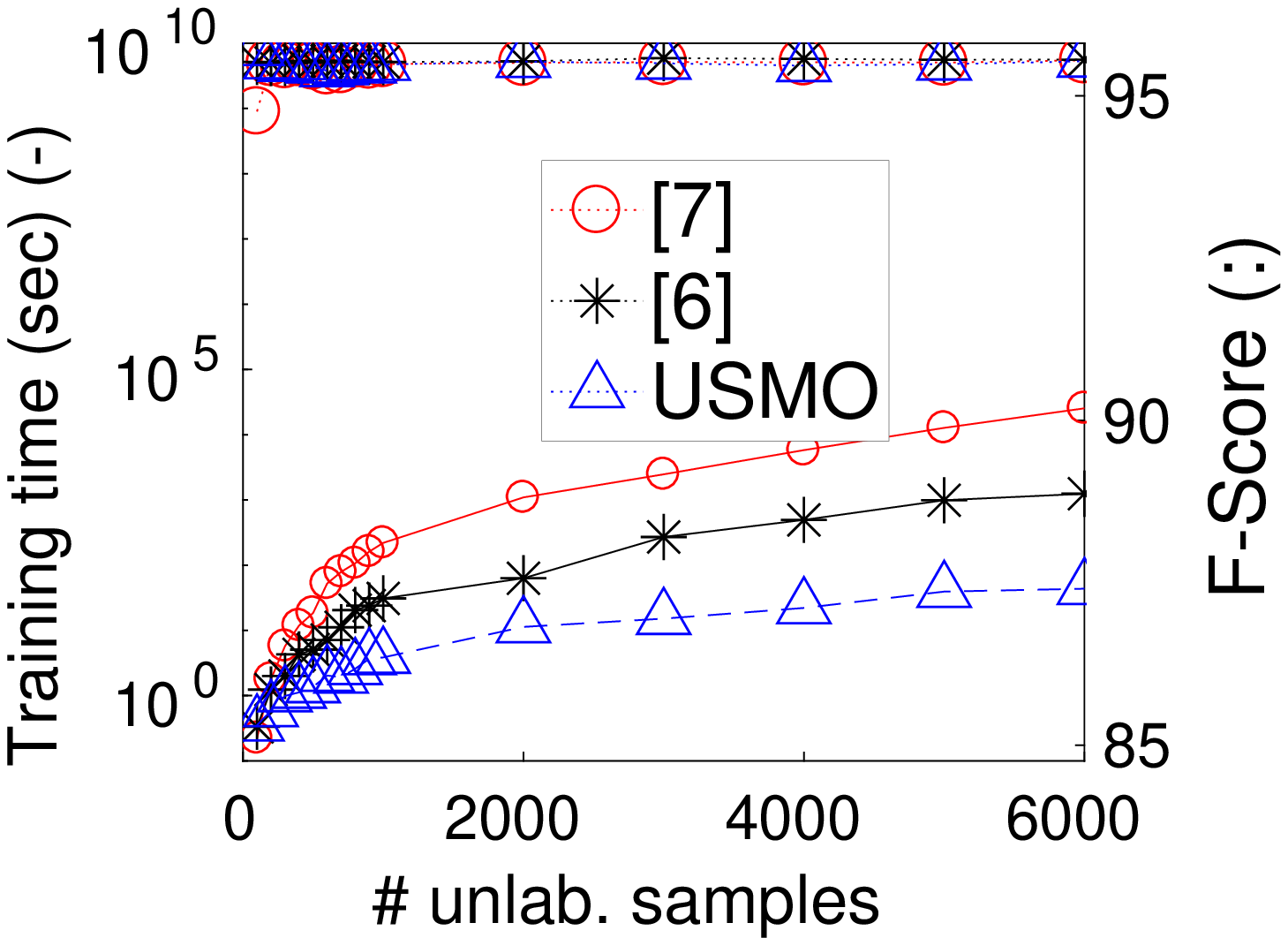}}
\subfloat[Statlog 2 vs. all]{\includegraphics[width=0.16\linewidth]{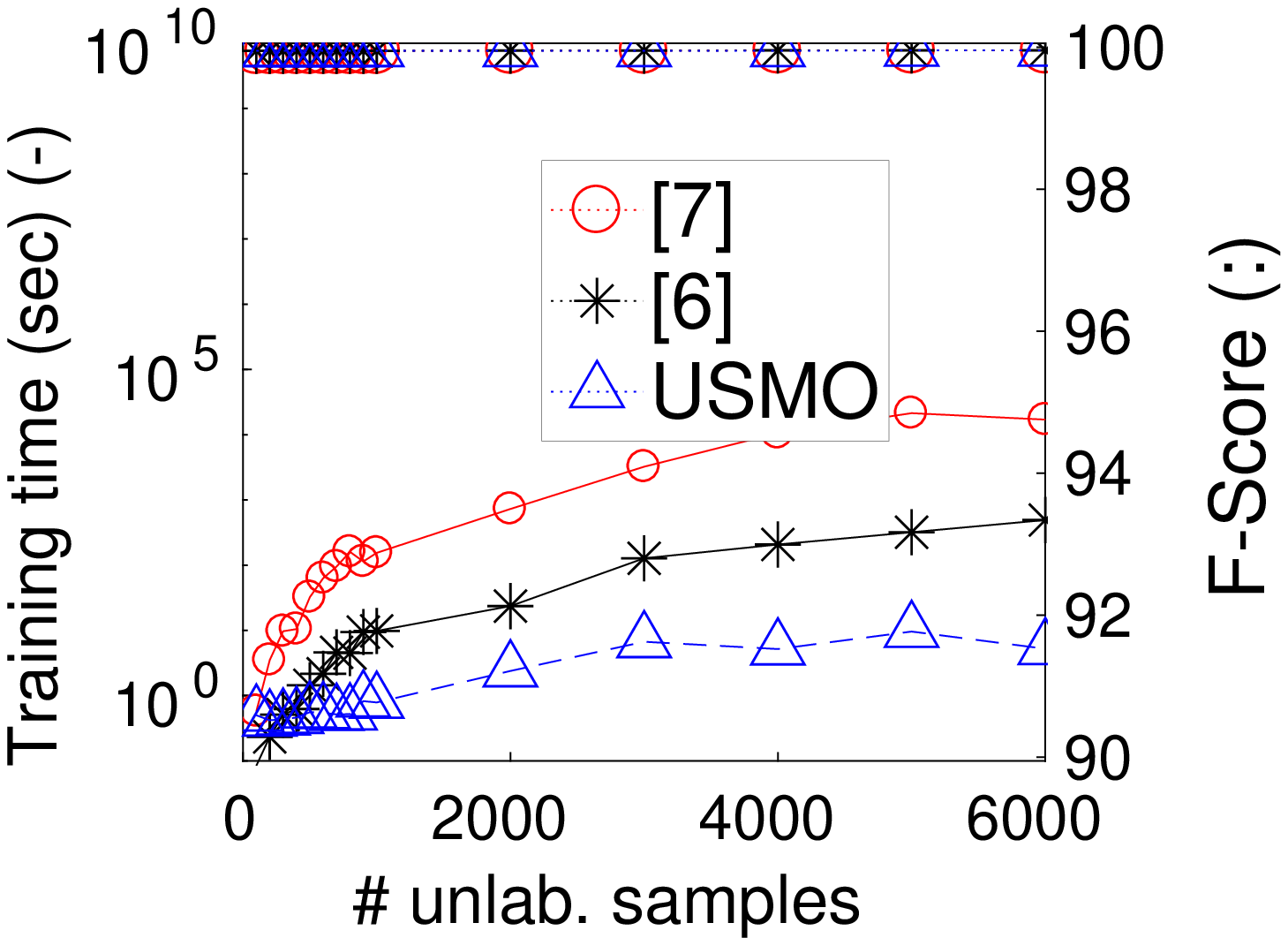}}
\subfloat[Statlog 3 vs. all]{\includegraphics[width=0.16\linewidth]{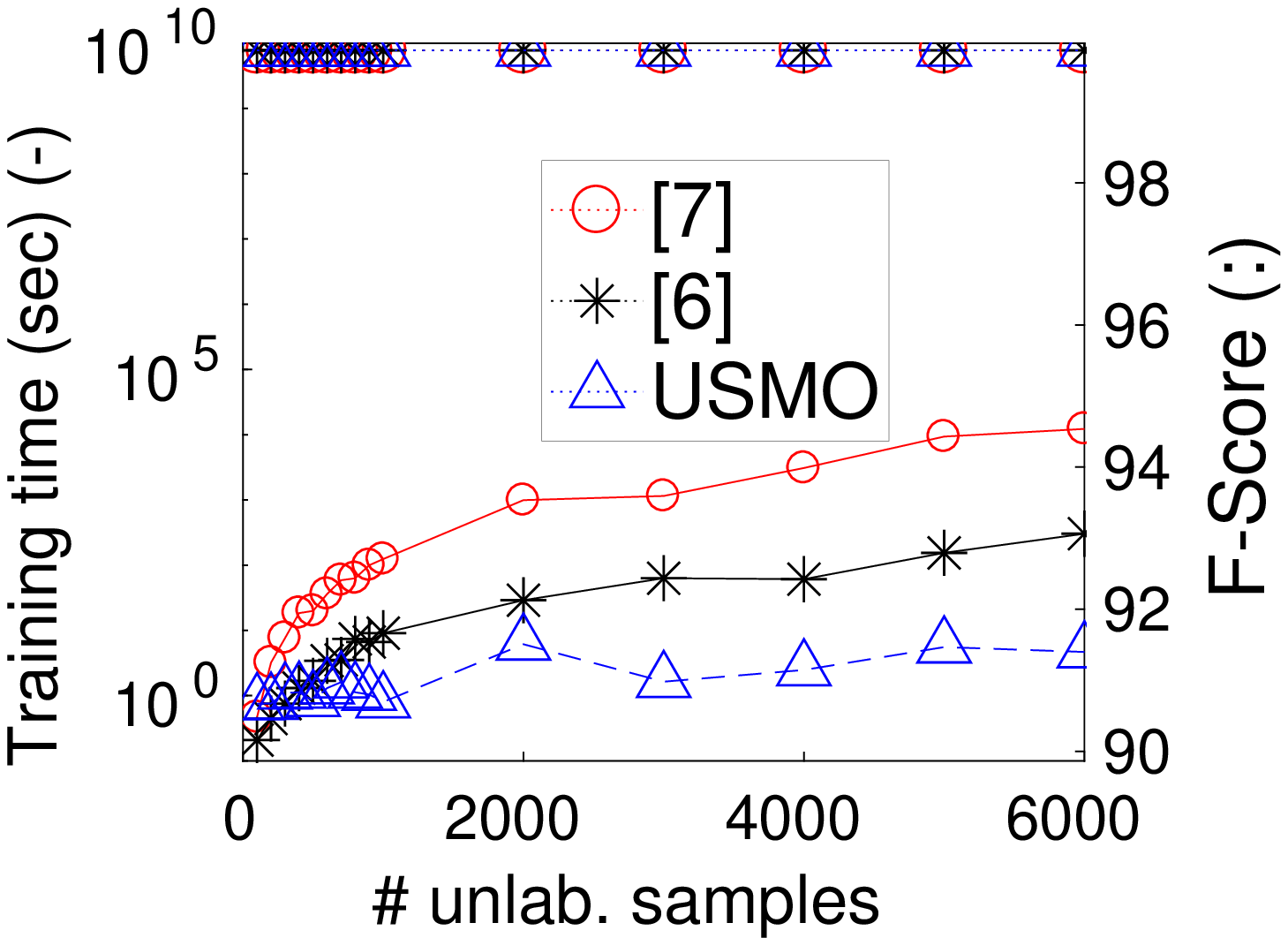}}
\subfloat[Statlog 4 vs. all]{\includegraphics[width=0.16\linewidth]{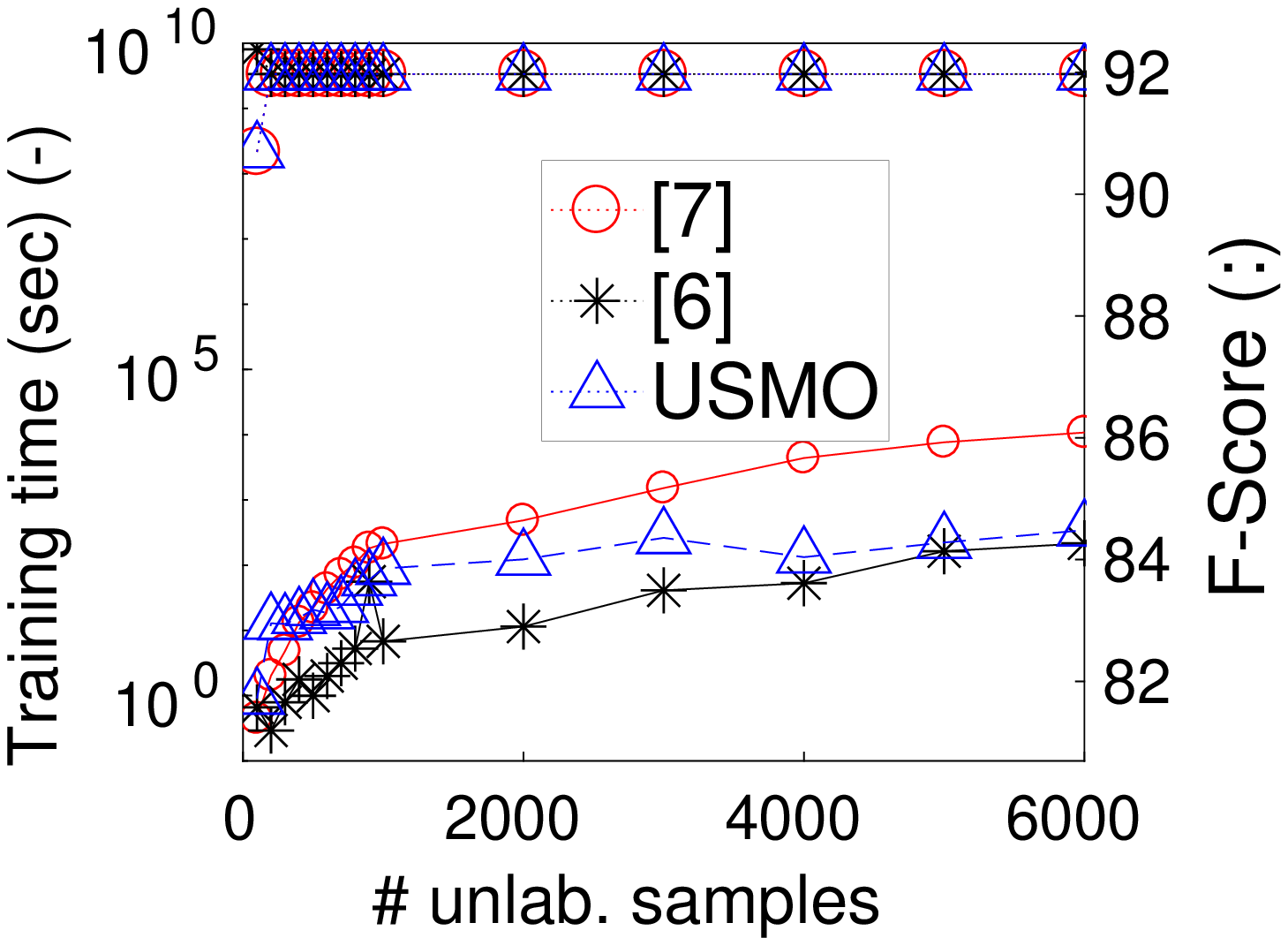}}
\subfloat[Statlog 5 vs. all]{\includegraphics[width=0.16\linewidth]{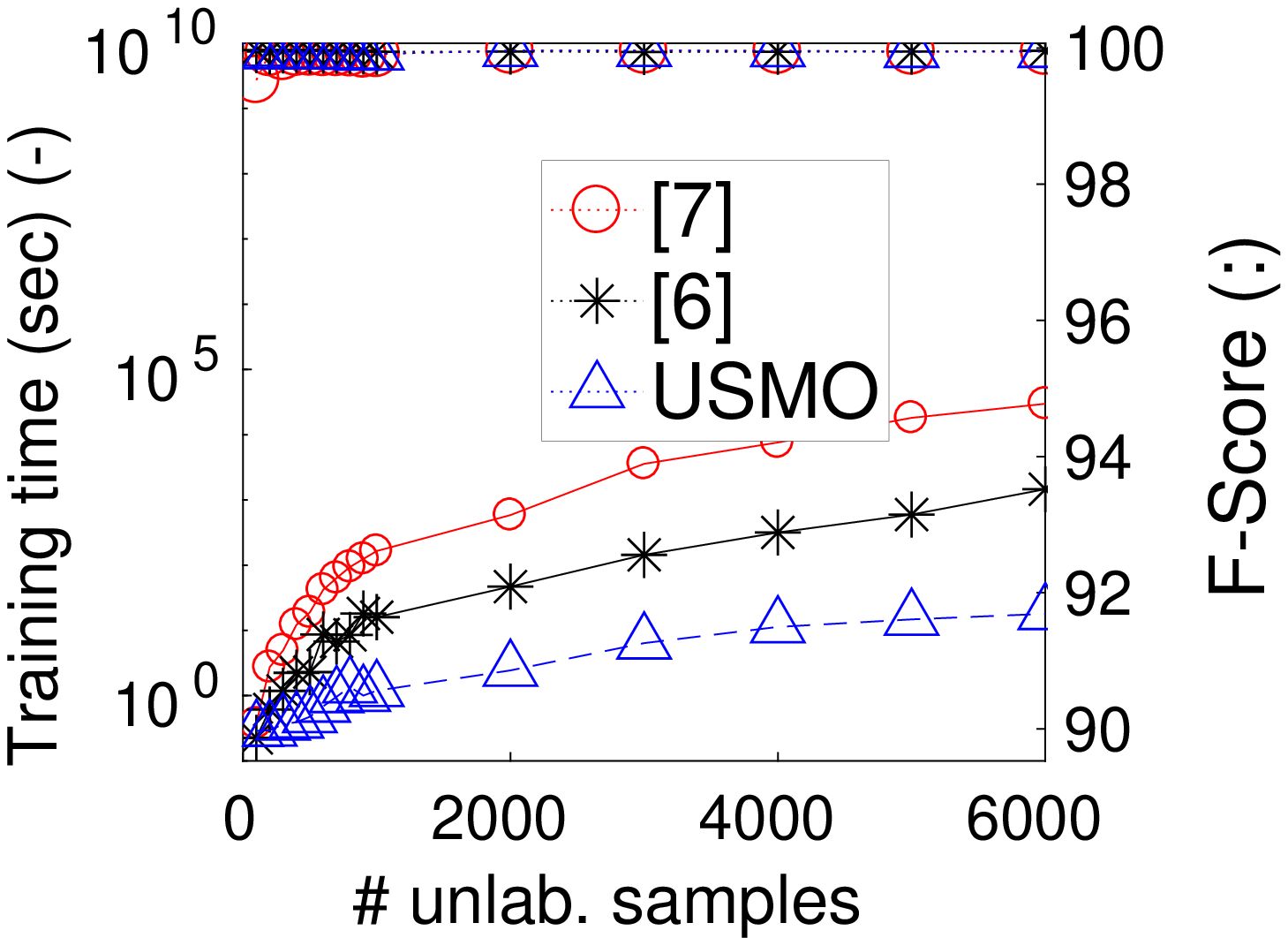}}
\subfloat[Statlog 6 vs. all]{\includegraphics[width=0.16\linewidth]{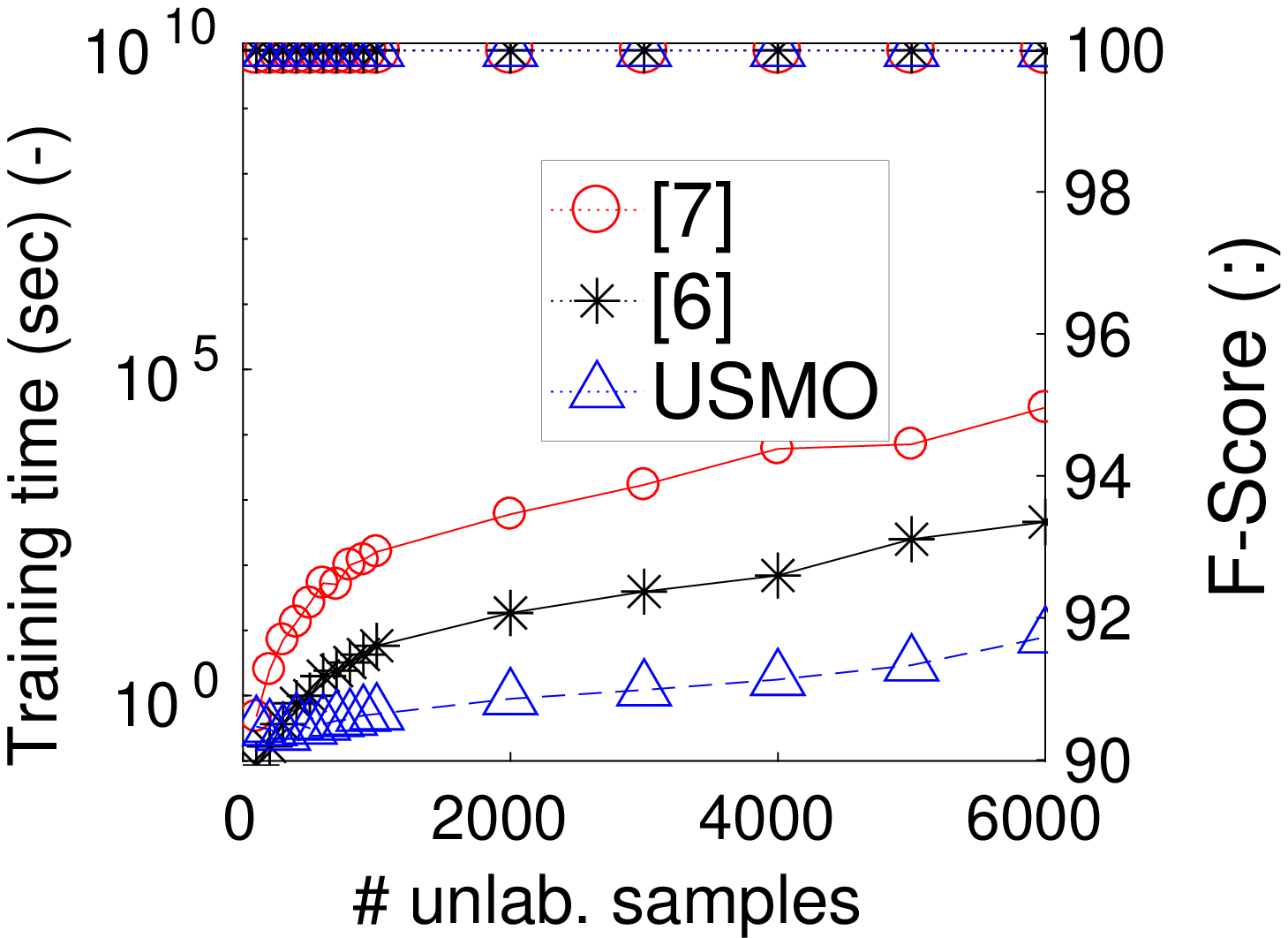}}\\
\subfloat[Statlog 7 vs. all]{\includegraphics[width=0.16\linewidth]{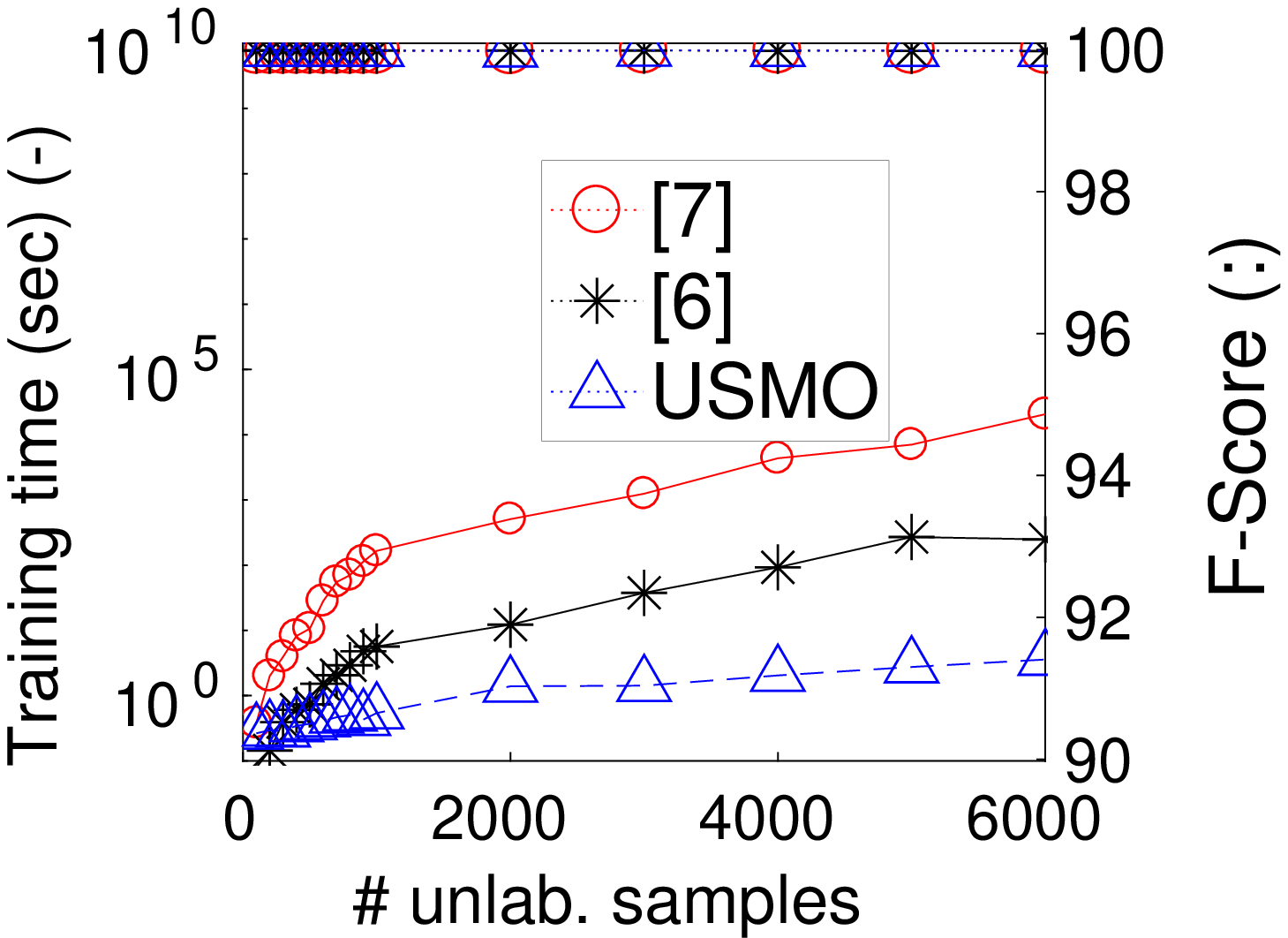}}
\subfloat[MNIST 0 vs. all]{\includegraphics[width=0.16\linewidth]{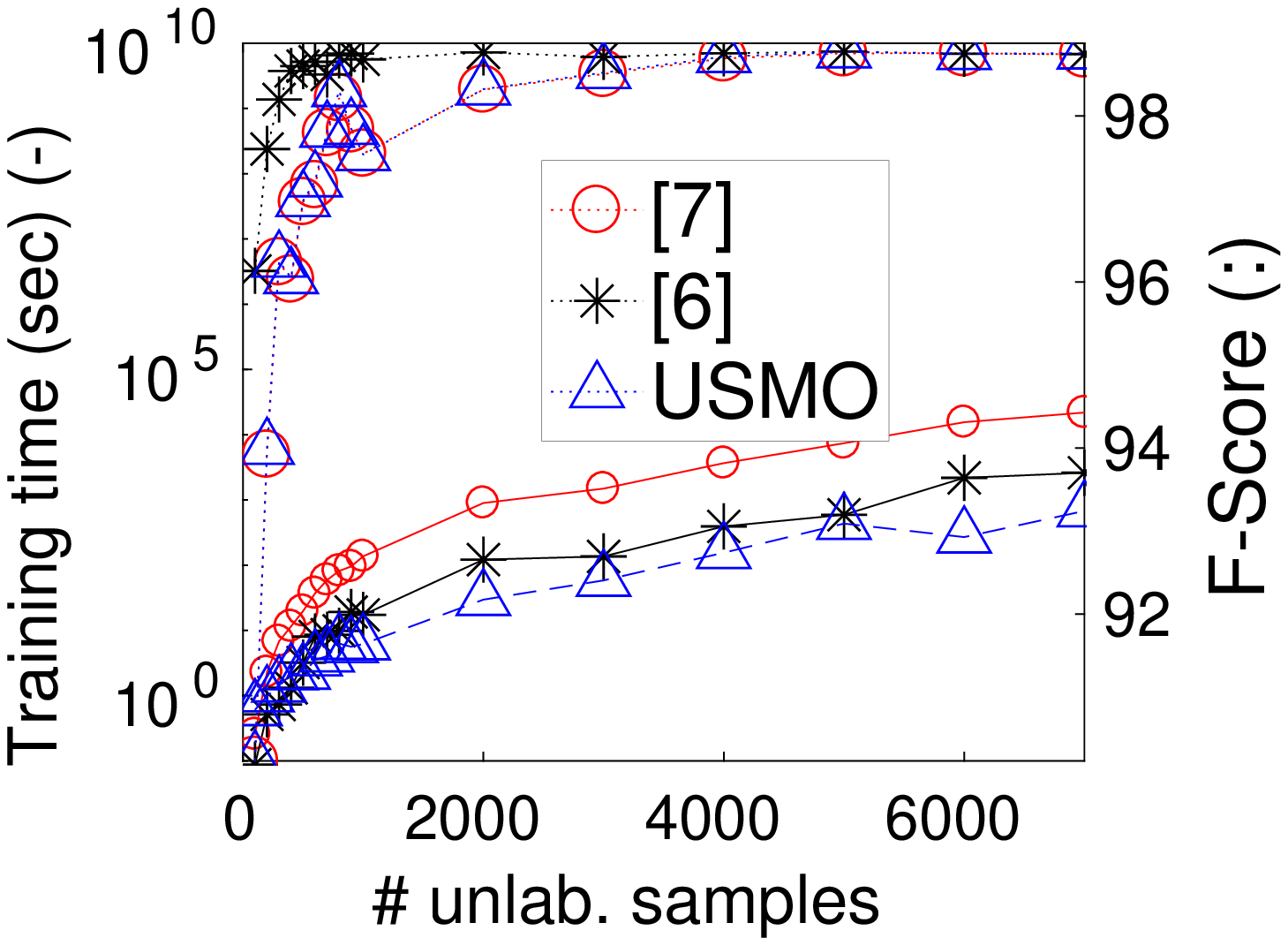}}
\subfloat[MNIST 1 vs. all]{\includegraphics[width=0.16\linewidth]{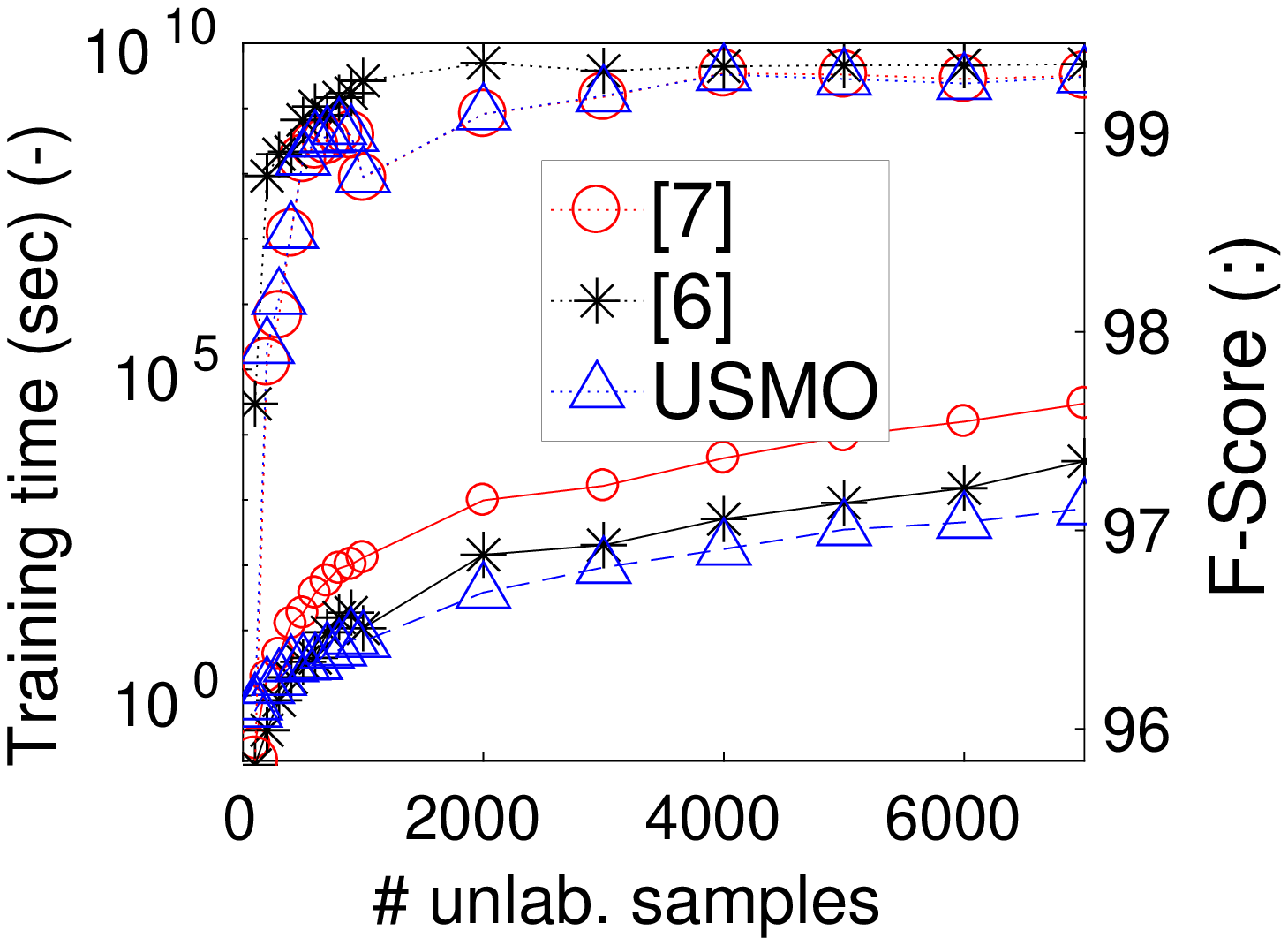}}
\subfloat[MNIST 2 vs. all]{\includegraphics[width=0.16\linewidth]{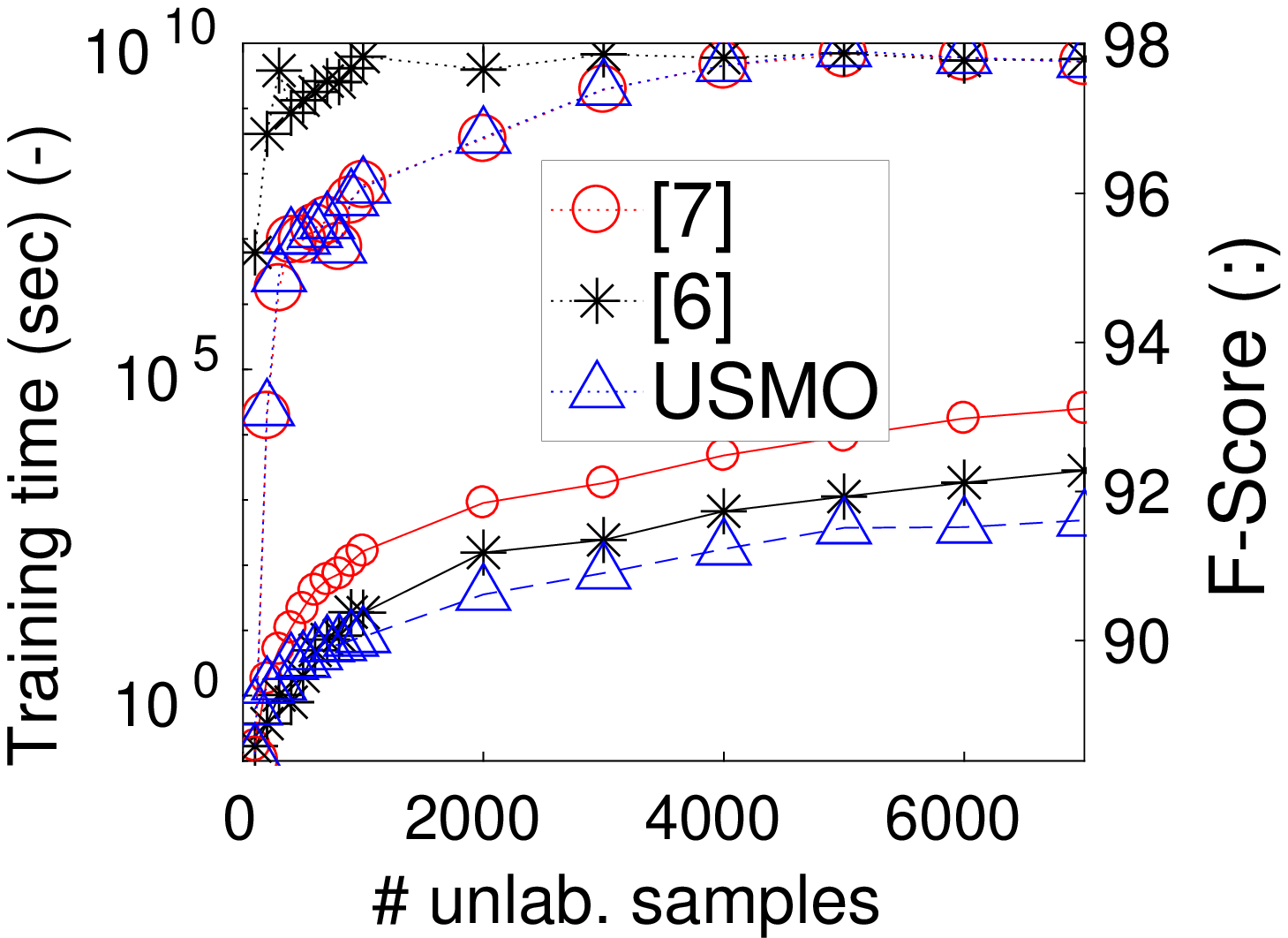}}
\subfloat[MNIST 3 vs. all]{\includegraphics[width=0.16\linewidth]{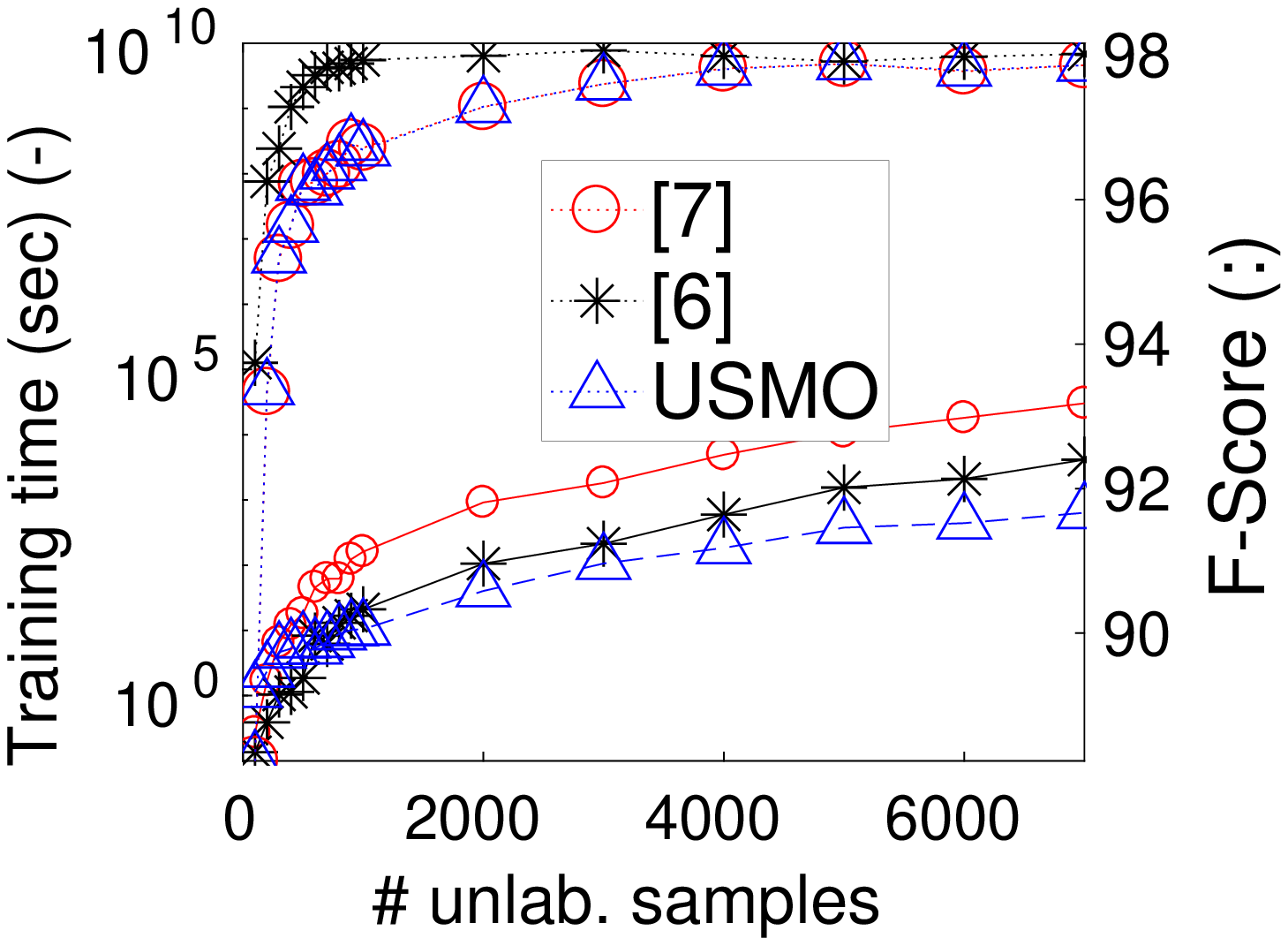}}
\subfloat[MNIST 4 vs. all]{\includegraphics[width=0.16\linewidth]{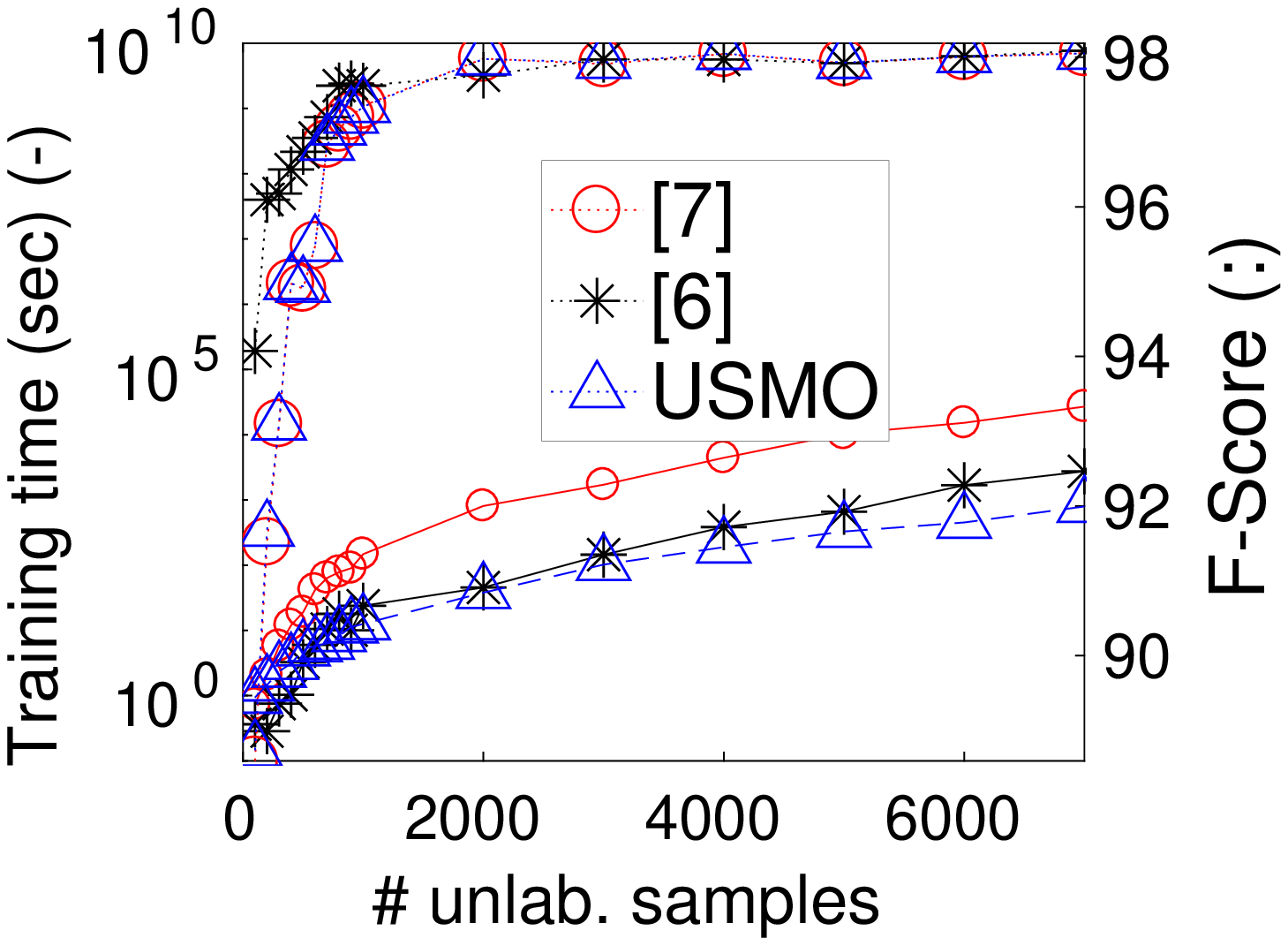}}\\
\subfloat[MNIST 5 vs. all]{\includegraphics[width=0.16\linewidth]{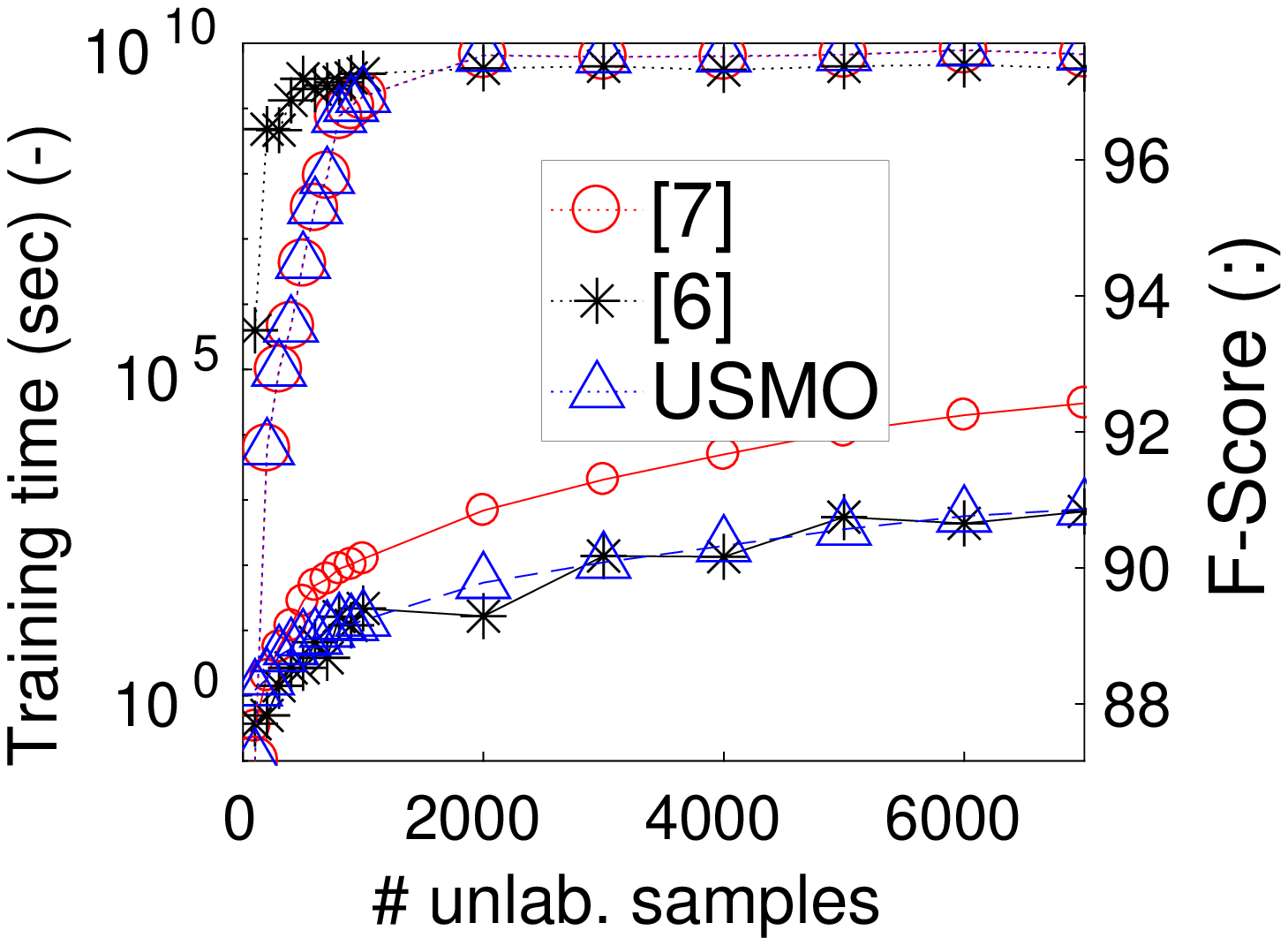}}
\subfloat[MNIST 6 vs. all]{\includegraphics[width=0.16\linewidth]{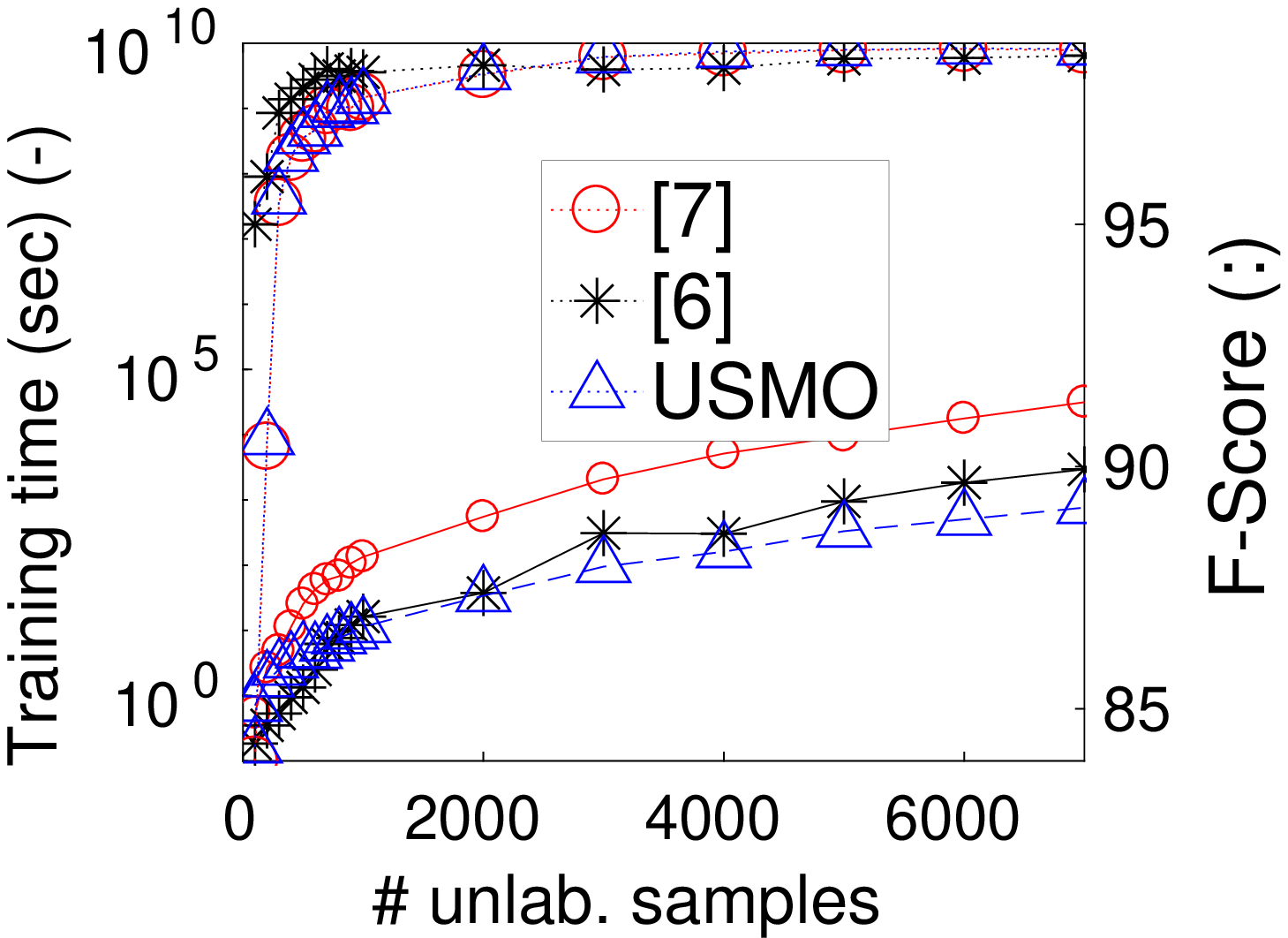}}
\subfloat[MNIST 7 vs. all]{\includegraphics[width=0.16\linewidth]{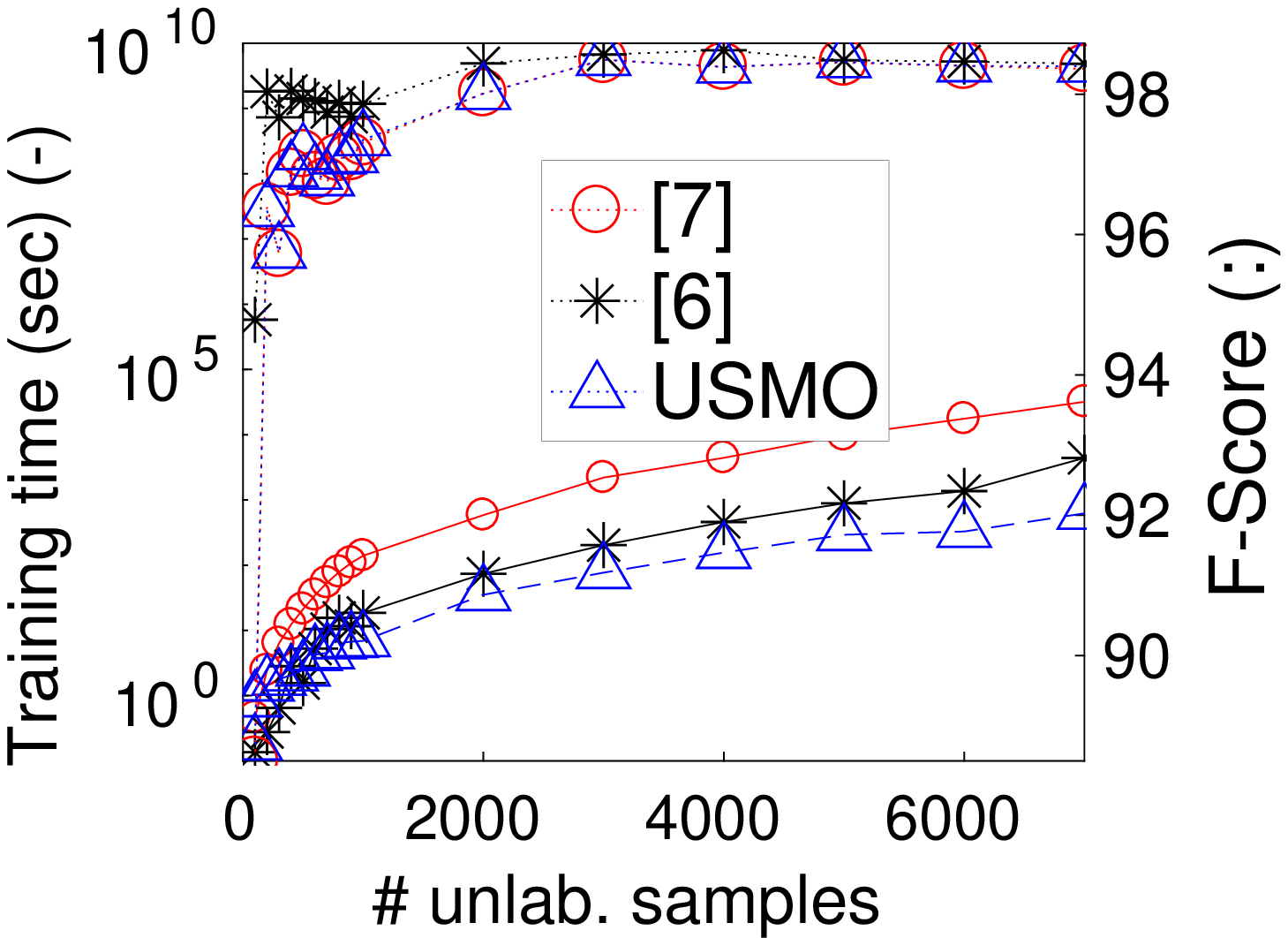}}
\subfloat[MNIST 8 vs. all]{\includegraphics[width=0.16\linewidth]{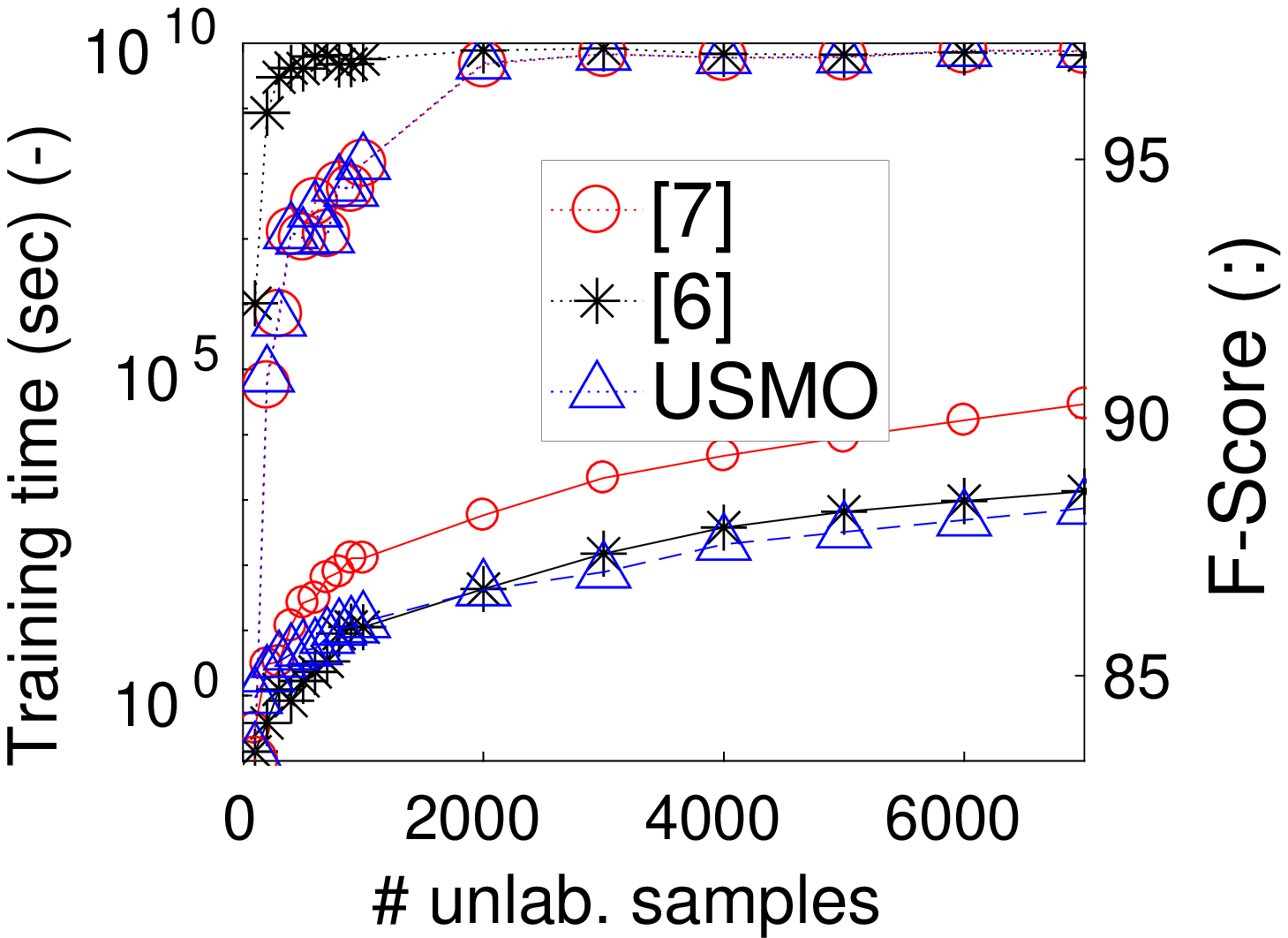}}
\subfloat[MNIST 9 vs. all]{\includegraphics[width=0.16\linewidth]{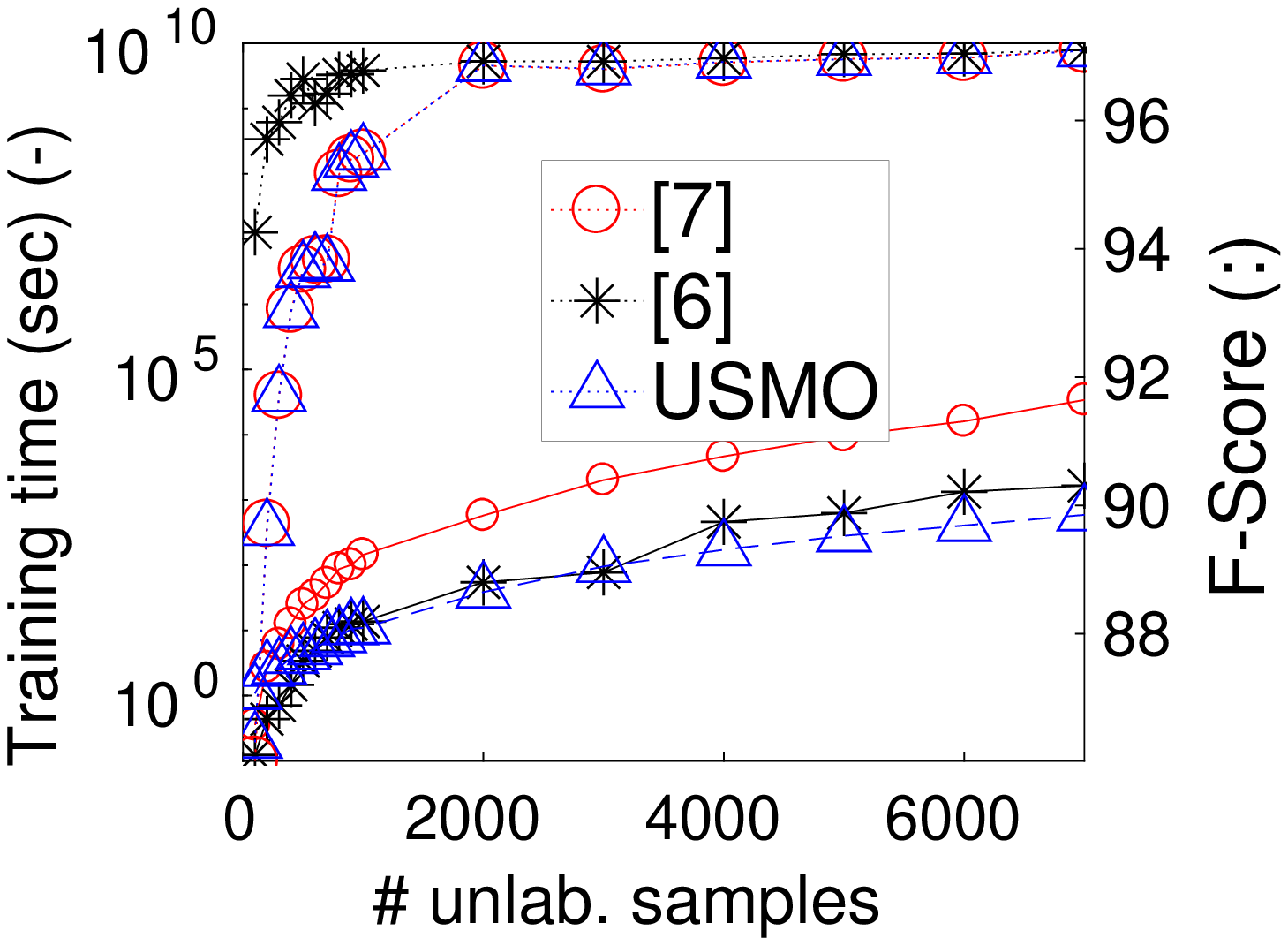}}
\caption{Comparative results on (a)-(g) Statlog (shuttle) and (h)-(q) MNIST datasets using the 
linear kernel ($\lambda = 0.01$). Each plot shows the training time against different number of 
unlabeled samples ($100$ positive samples) as well as the generalization performance on the test set.}
\label{fig:results_large1}
\end{figure*}

\begin{figure*}[h!]
\centering
\subfloat[Bank 1 vs. all]{\includegraphics[width=0.16\linewidth]{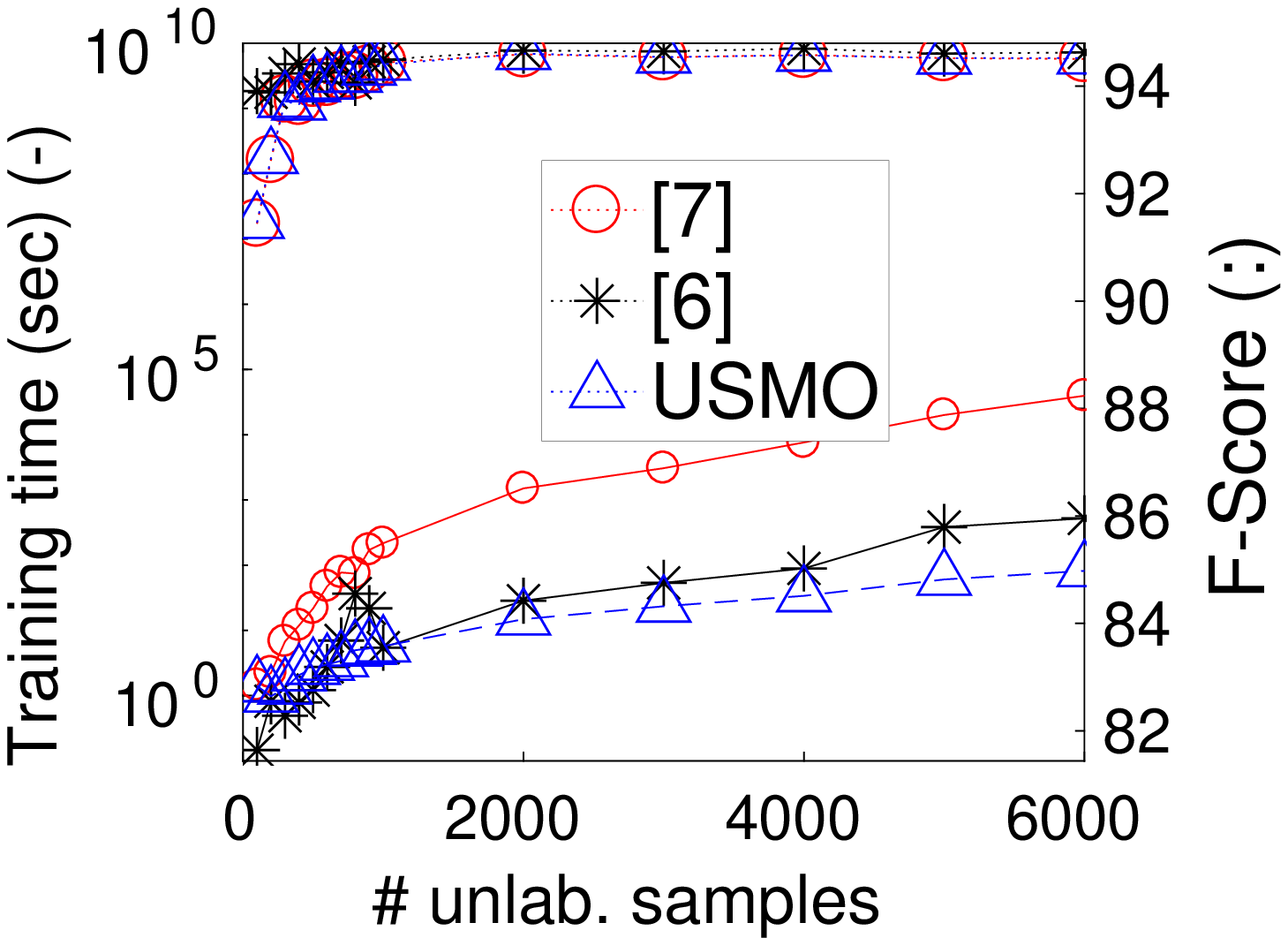}}
\subfloat[Adult 1 vs. all]{\includegraphics[width=0.16\linewidth]{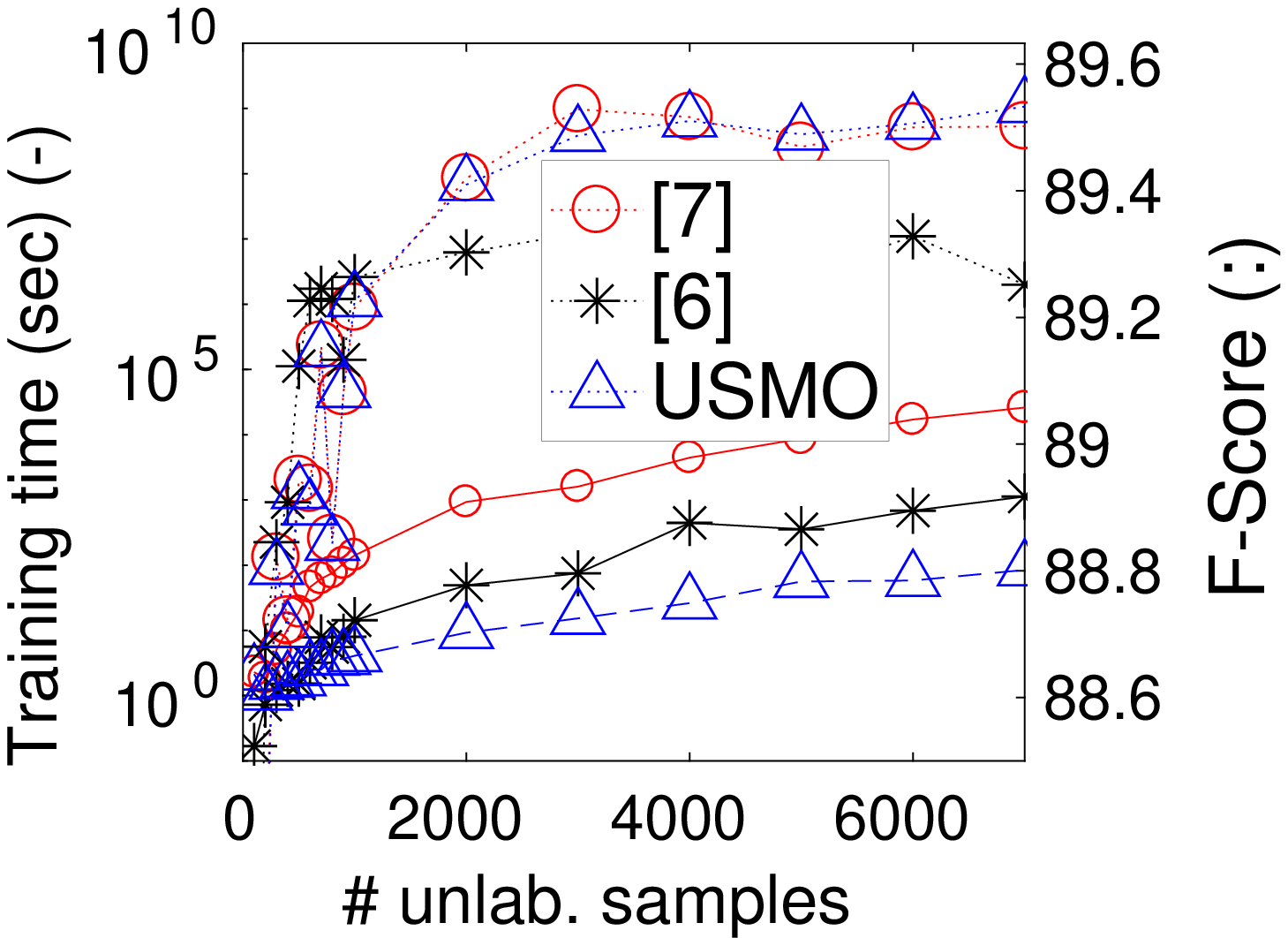}}
\subfloat[POKER 0 vs. all]{\includegraphics[width=0.16\linewidth]{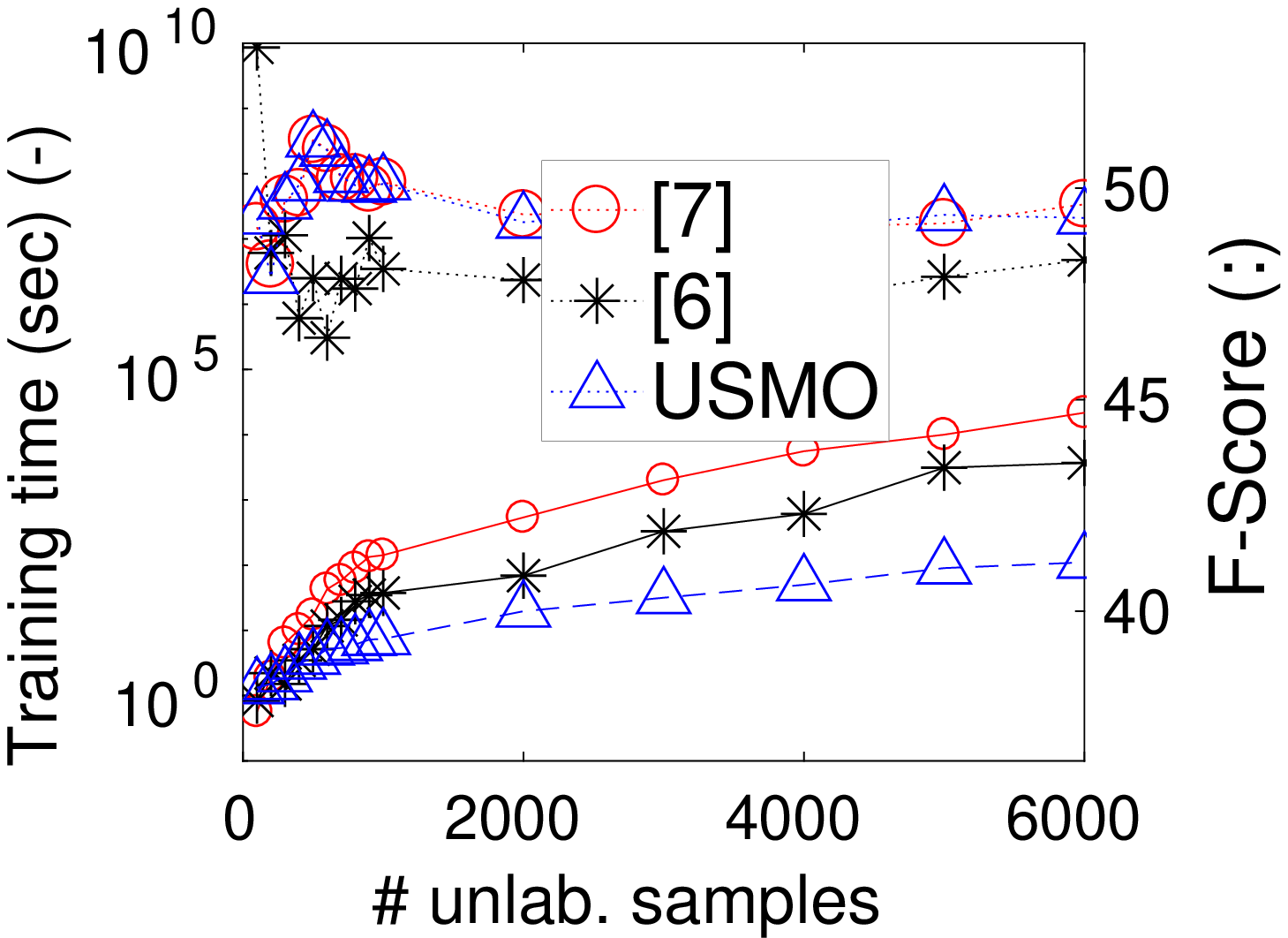}}
\subfloat[POKER 1 vs. all]{\includegraphics[width=0.16\linewidth]{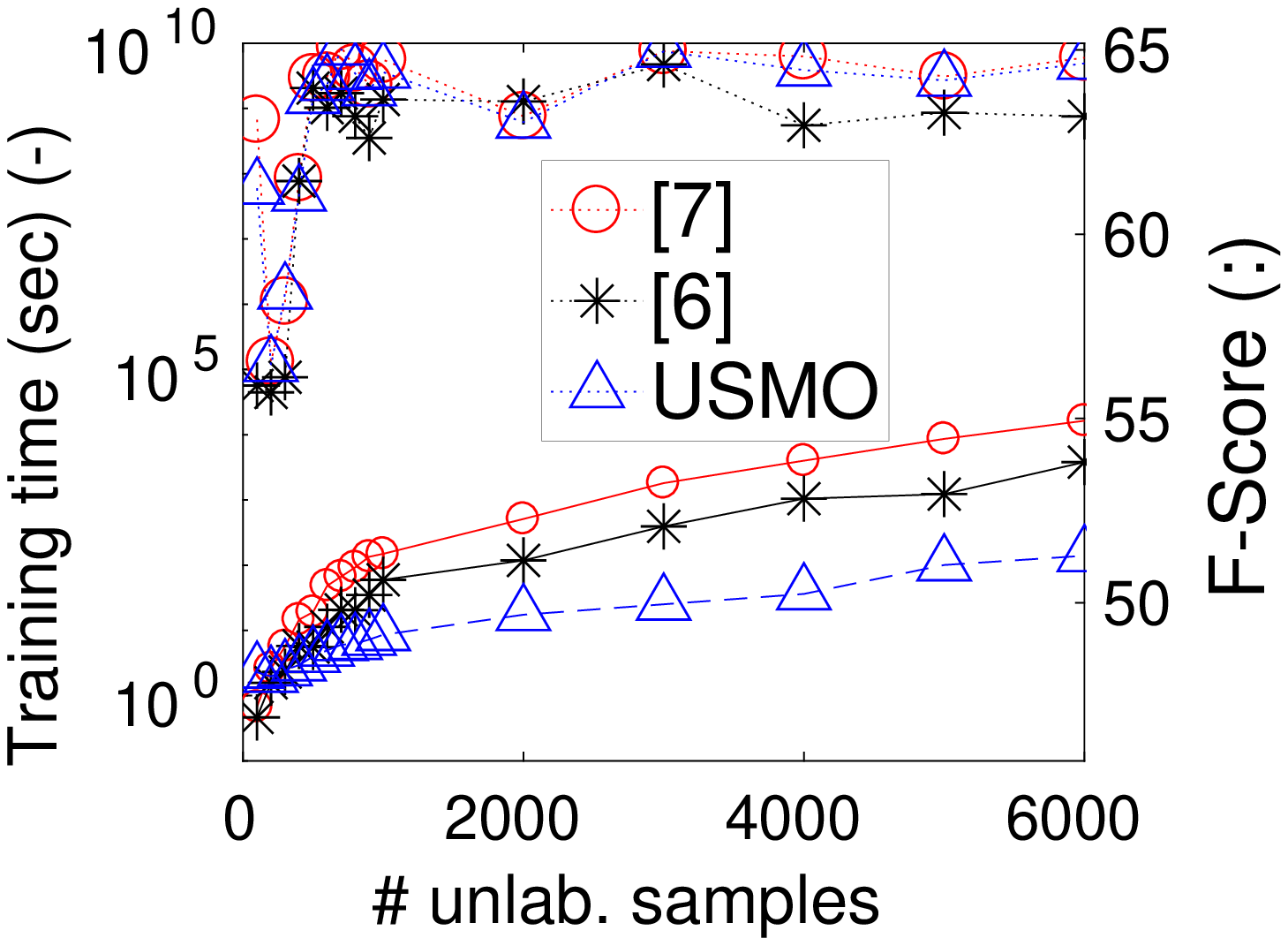}}
\subfloat[POKER 2 vs. all]{\includegraphics[width=0.16\linewidth]{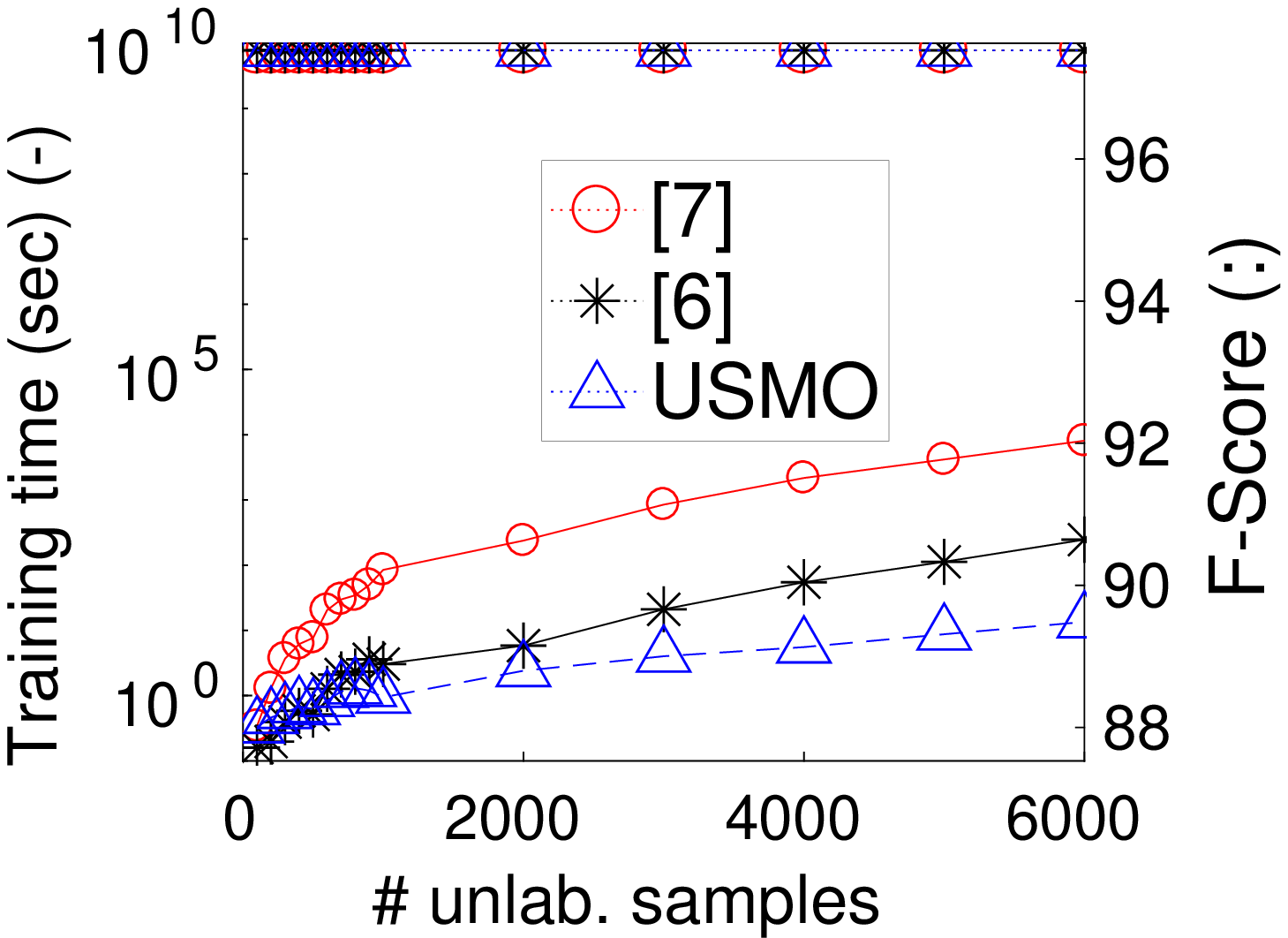}}
\subfloat[POKER 3 vs. all]{\includegraphics[width=0.16\linewidth]{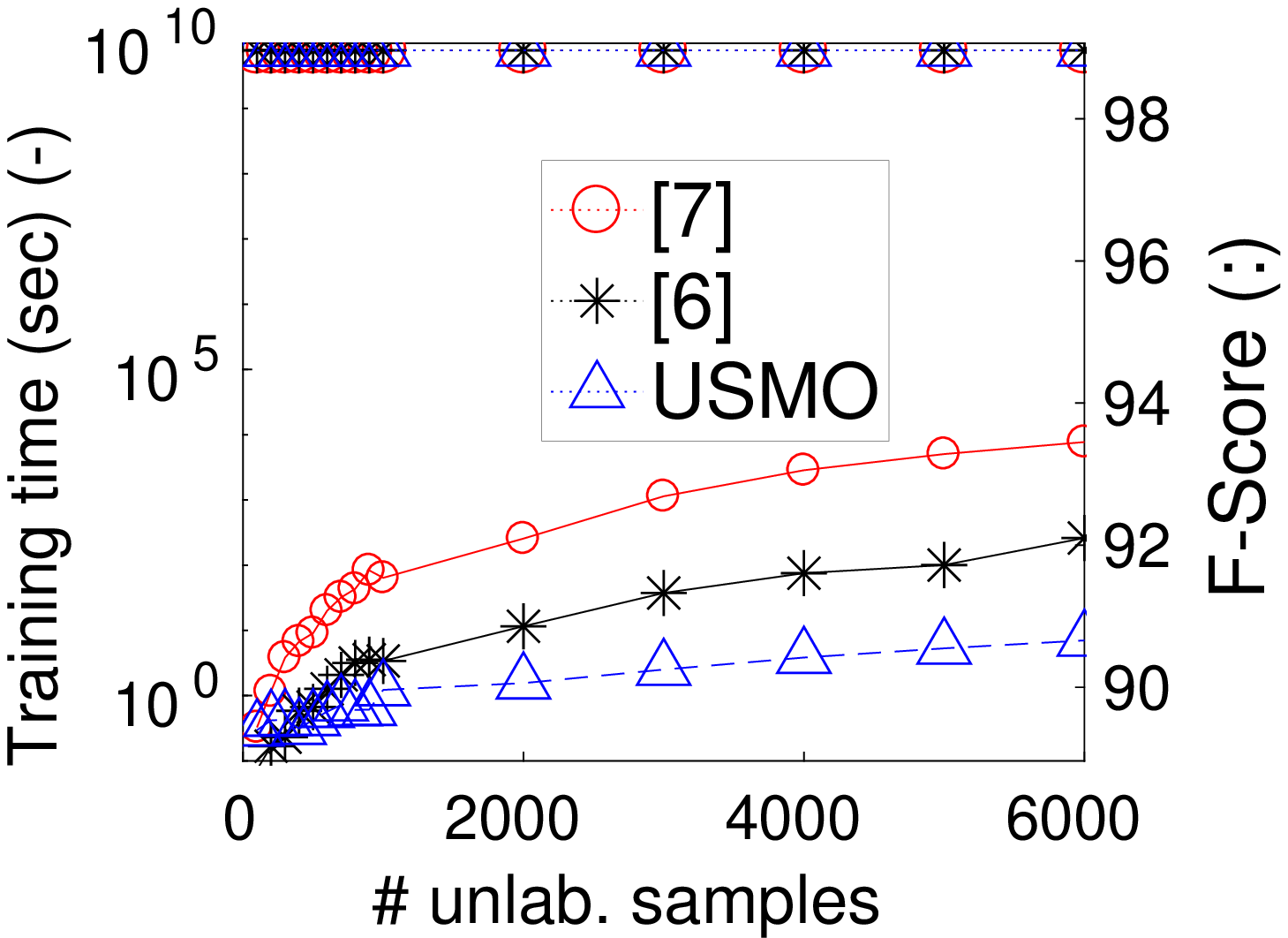}}\\
\subfloat[POKER 4 vs. all]{\includegraphics[width=0.16\linewidth]{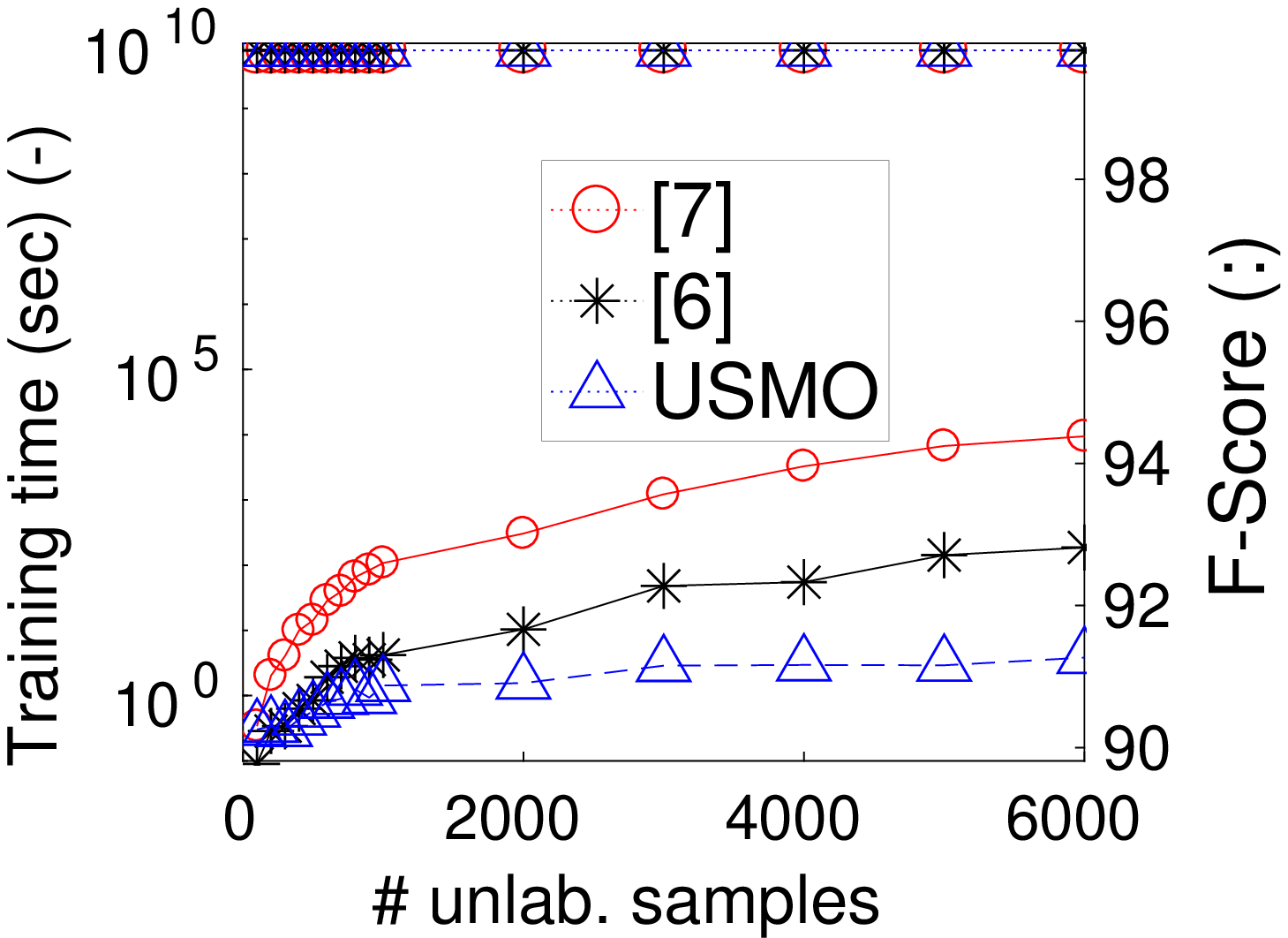}}
\subfloat[POKER 5 vs. all]{\includegraphics[width=0.16\linewidth]{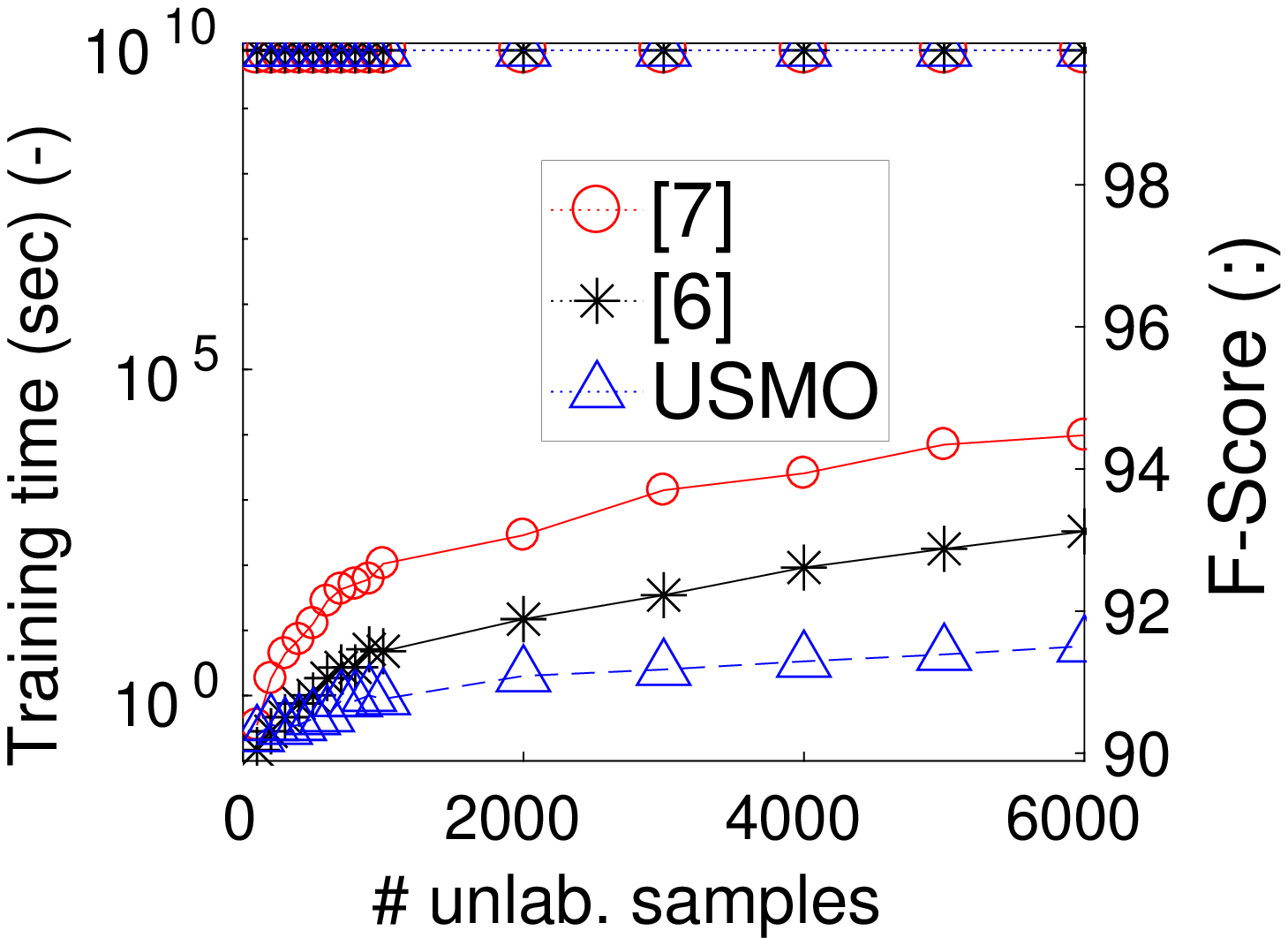}}
\subfloat[POKER 6 vs. all]{\includegraphics[width=0.16\linewidth]{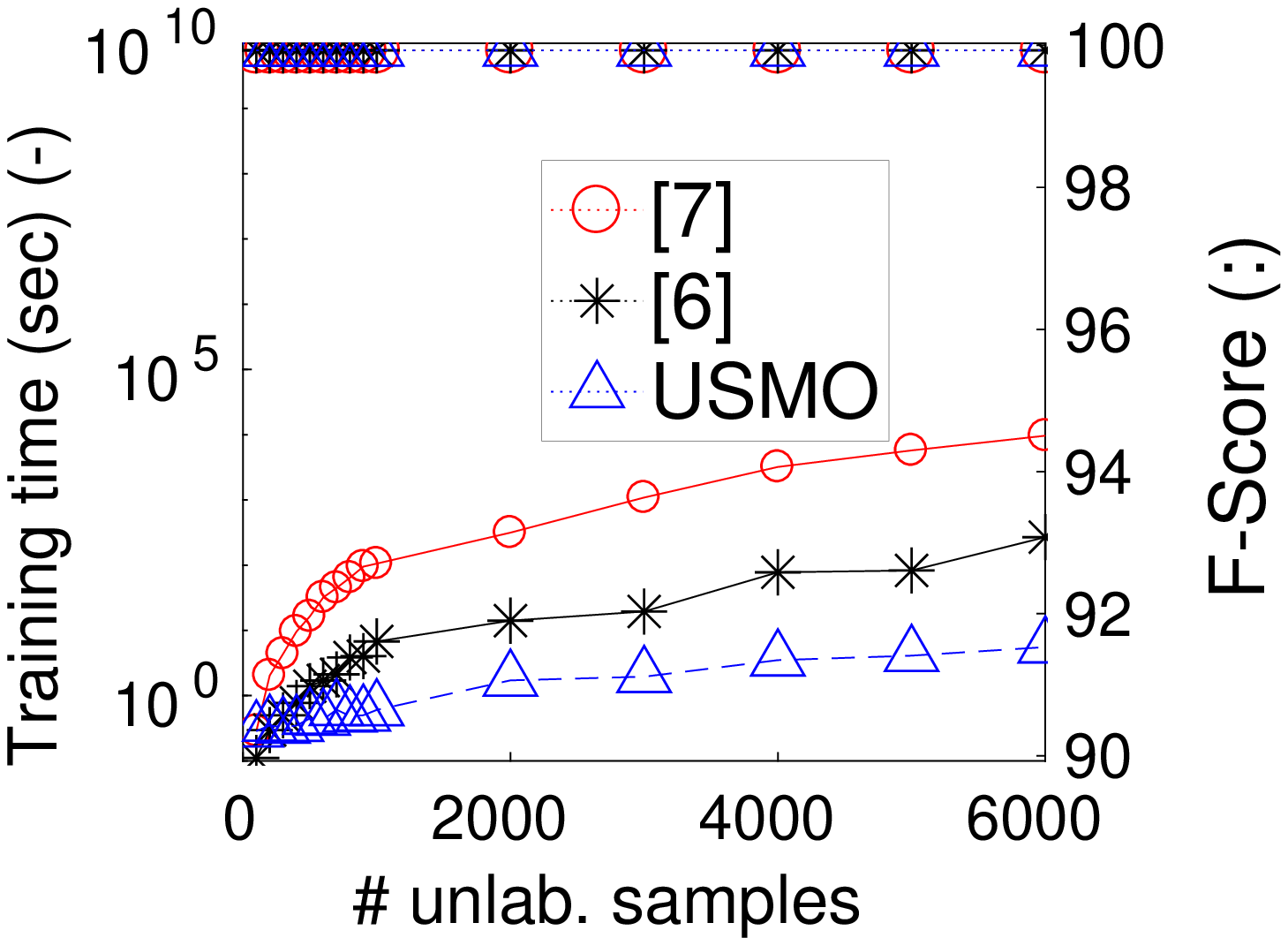}}
\subfloat[POKER 7 vs. all]{\includegraphics[width=0.16\linewidth]{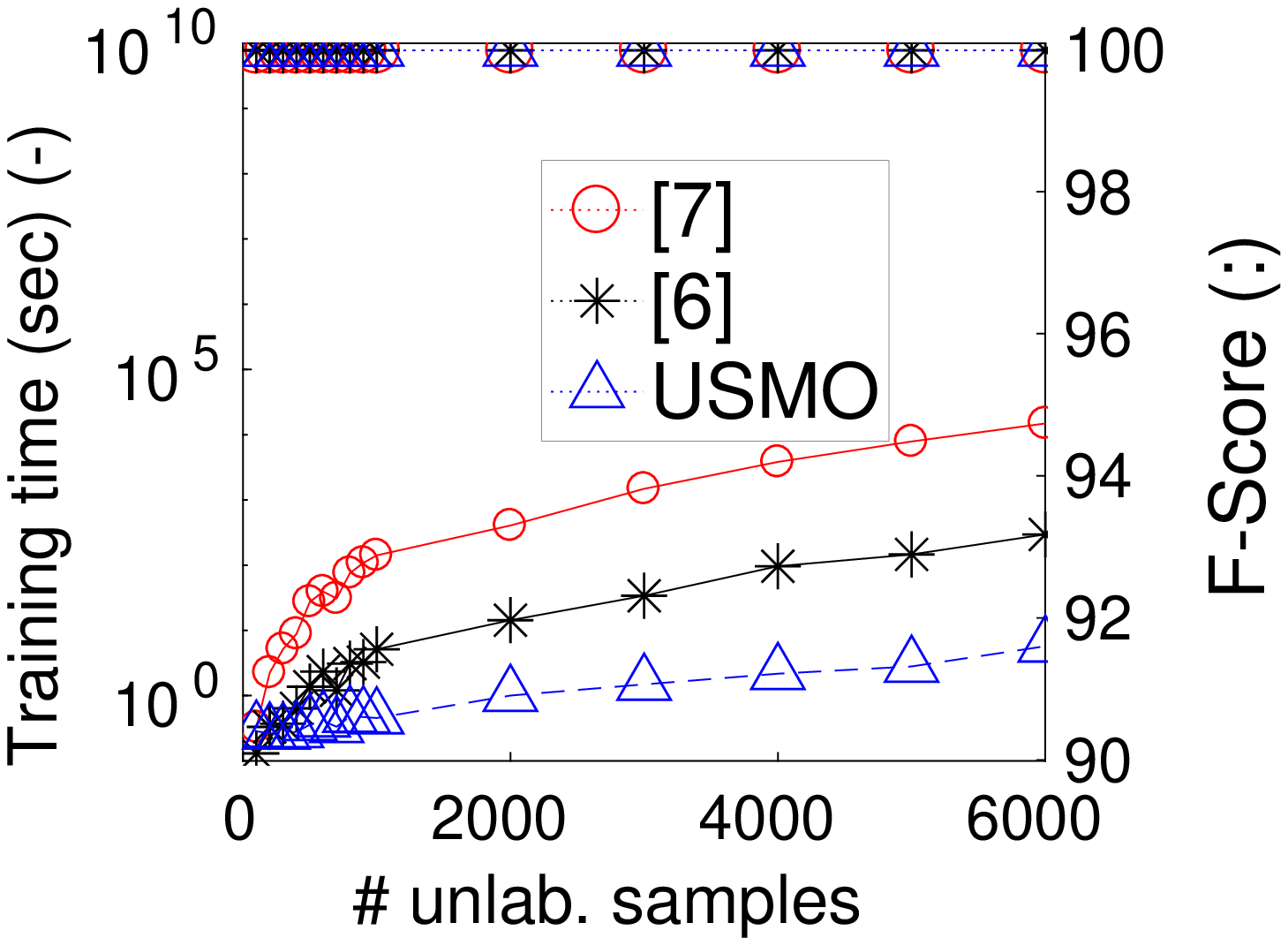}}
\subfloat[POKER 8 vs. all]{\includegraphics[width=0.16\linewidth]{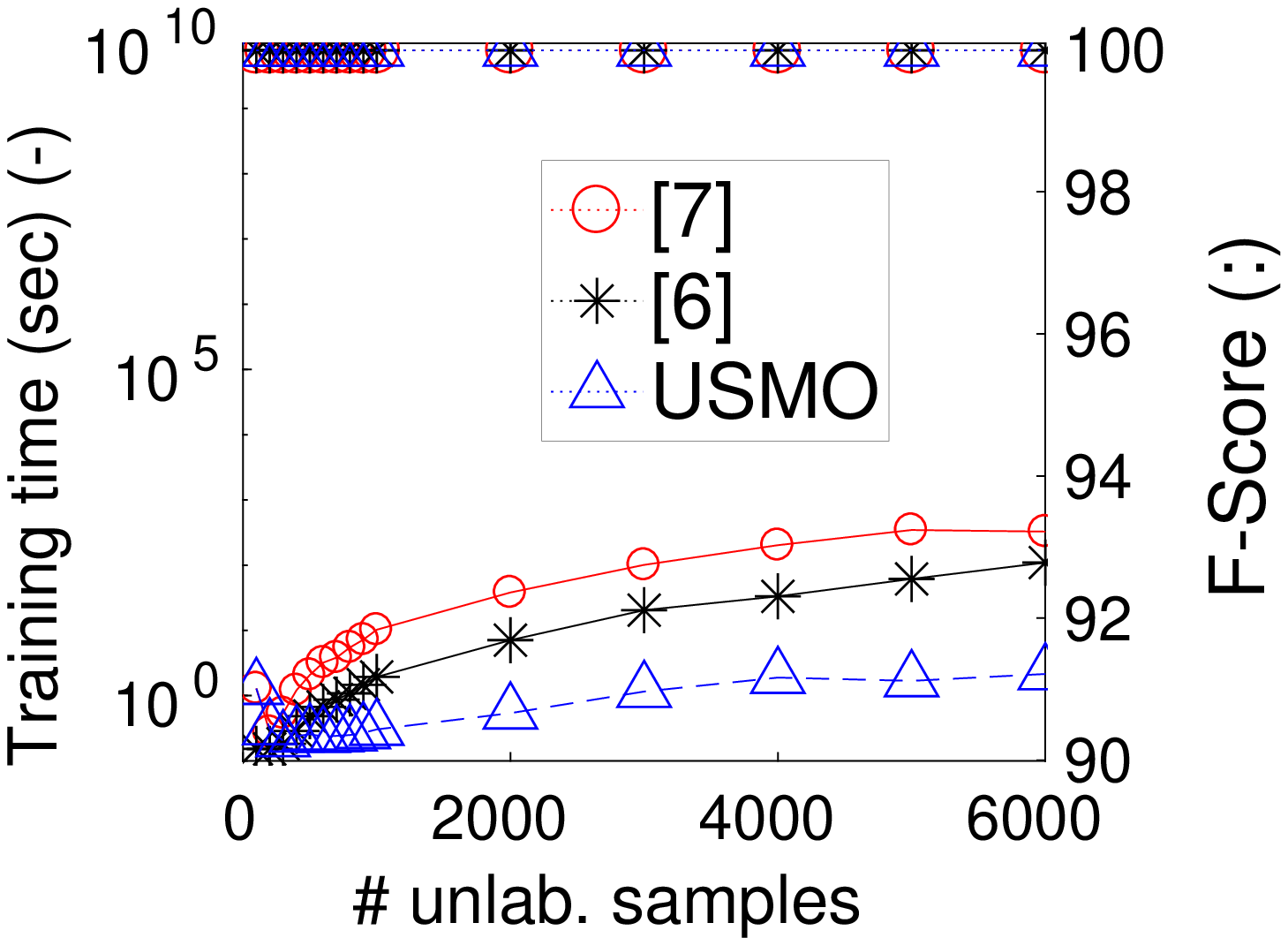}}
\subfloat[POKER 9 vs. all]{\includegraphics[width=0.16\linewidth]{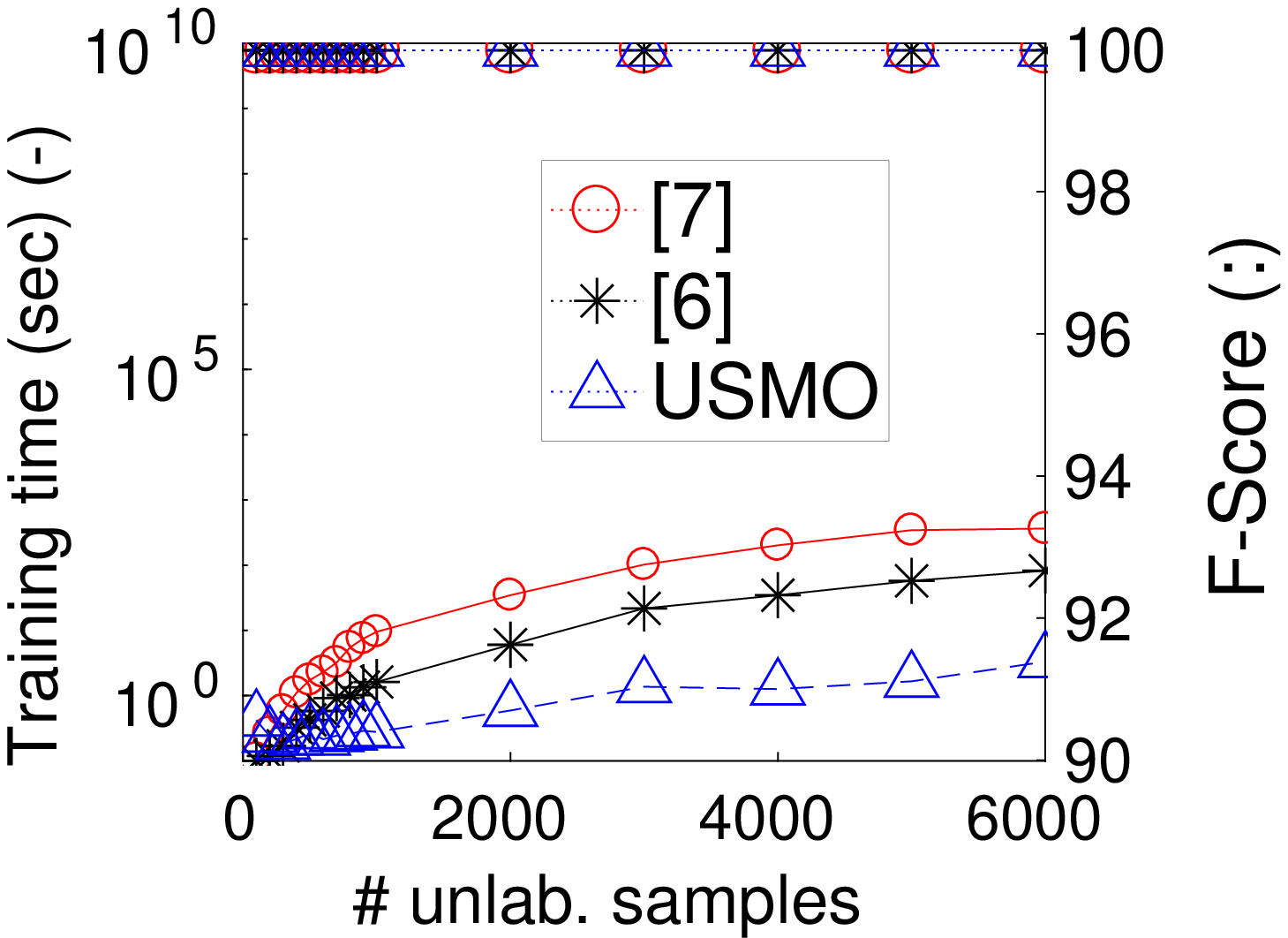}}
\caption{Comparative results on (a) Bank-marketing, (b) Adult and (c)-(l) Poker-hand datasets using 
the linear kernel ($\lambda = 0.01$). Each plot shows the training time against different number of 
unlabeled samples ($100$ positive samples) as well as the generalization performance on the test set.}
\label{fig:results_large2}
\end{figure*}

\begin{figure*}[h!]
\centering
\subfloat[Statlog 1 vs. all]{\includegraphics[width=0.16\linewidth]{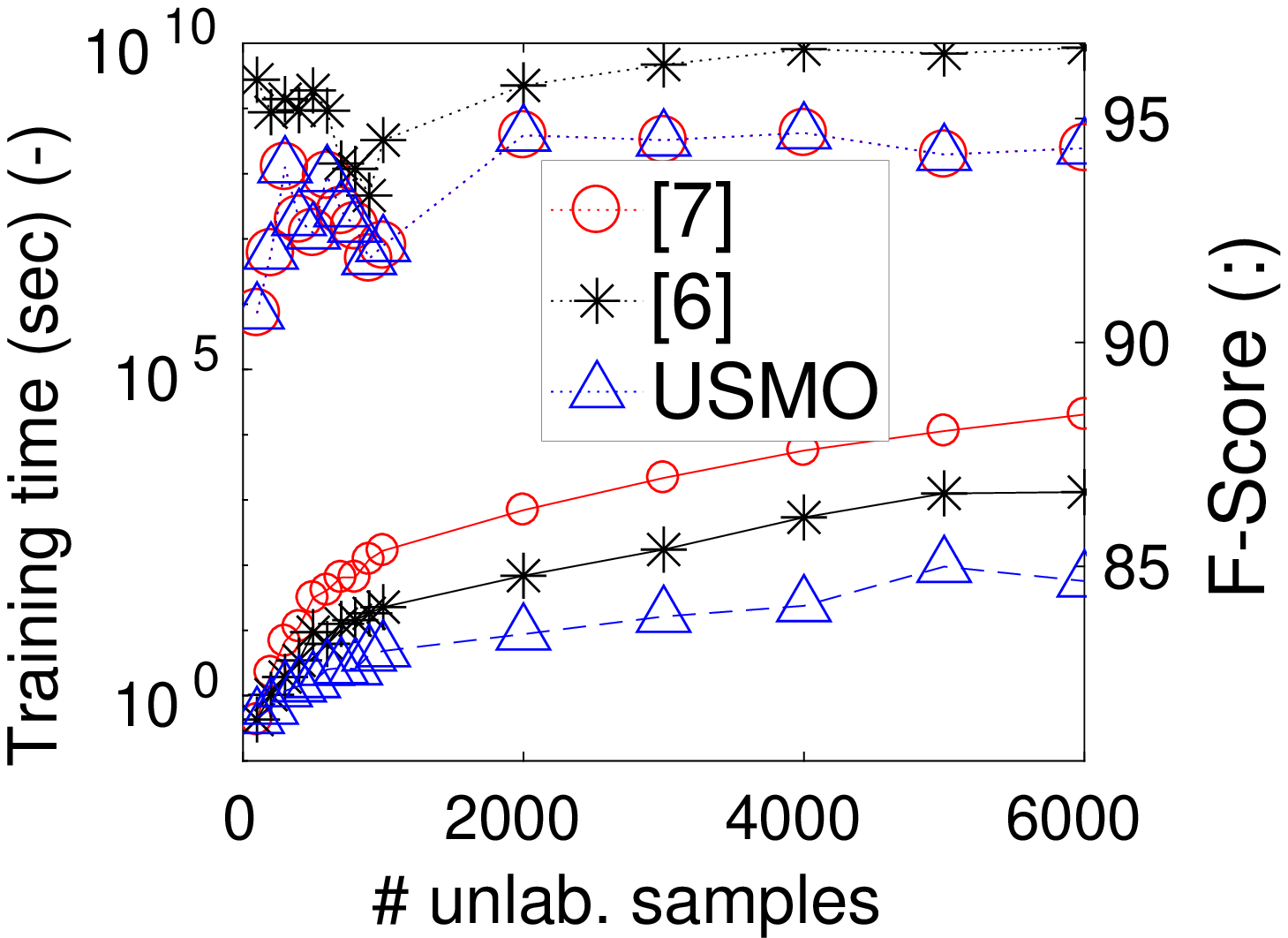}}
\subfloat[Statlog 2 vs. all]{\includegraphics[width=0.16\linewidth]{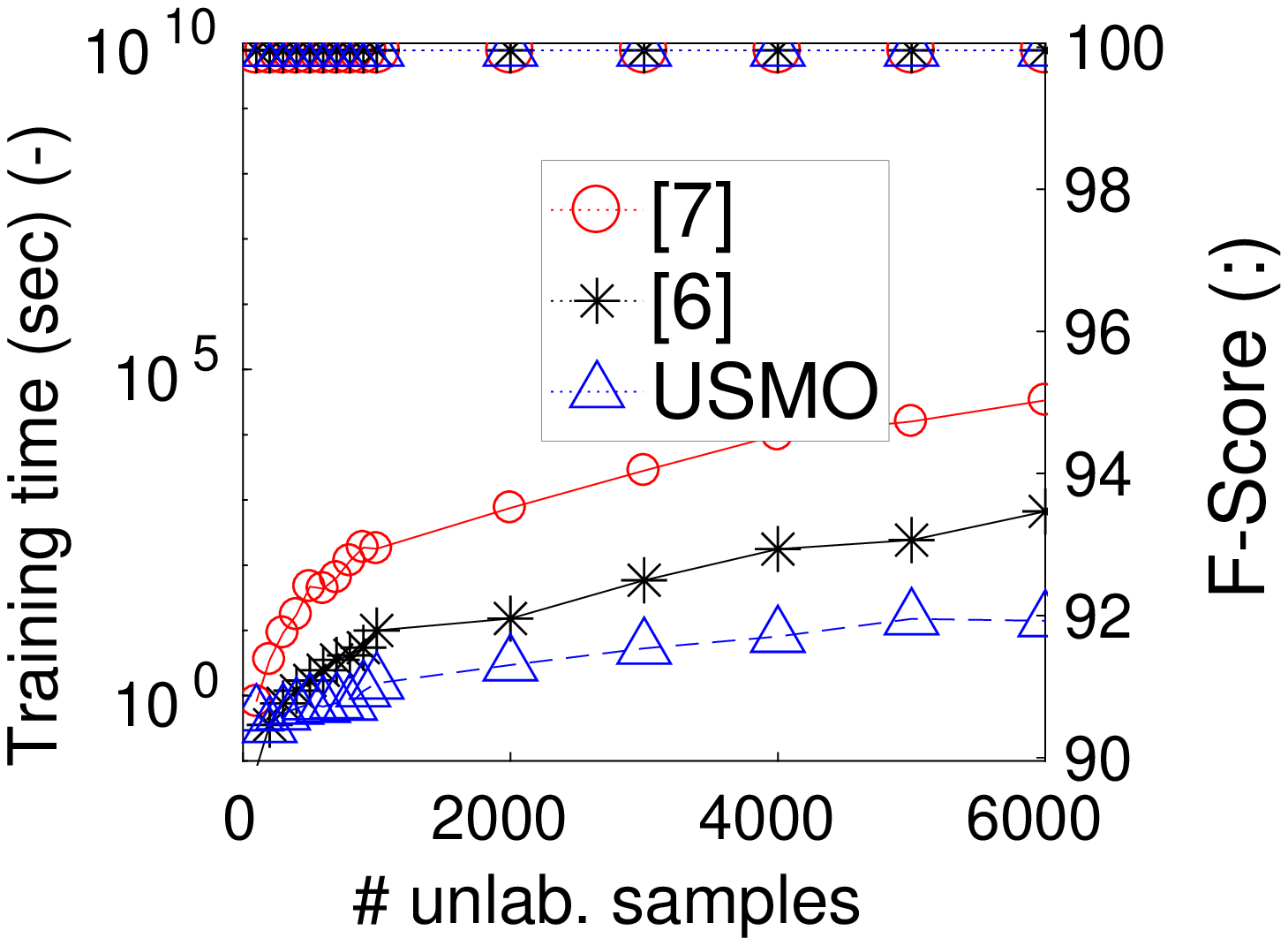}}
\subfloat[Statlog 3 vs. all]{\includegraphics[width=0.16\linewidth]{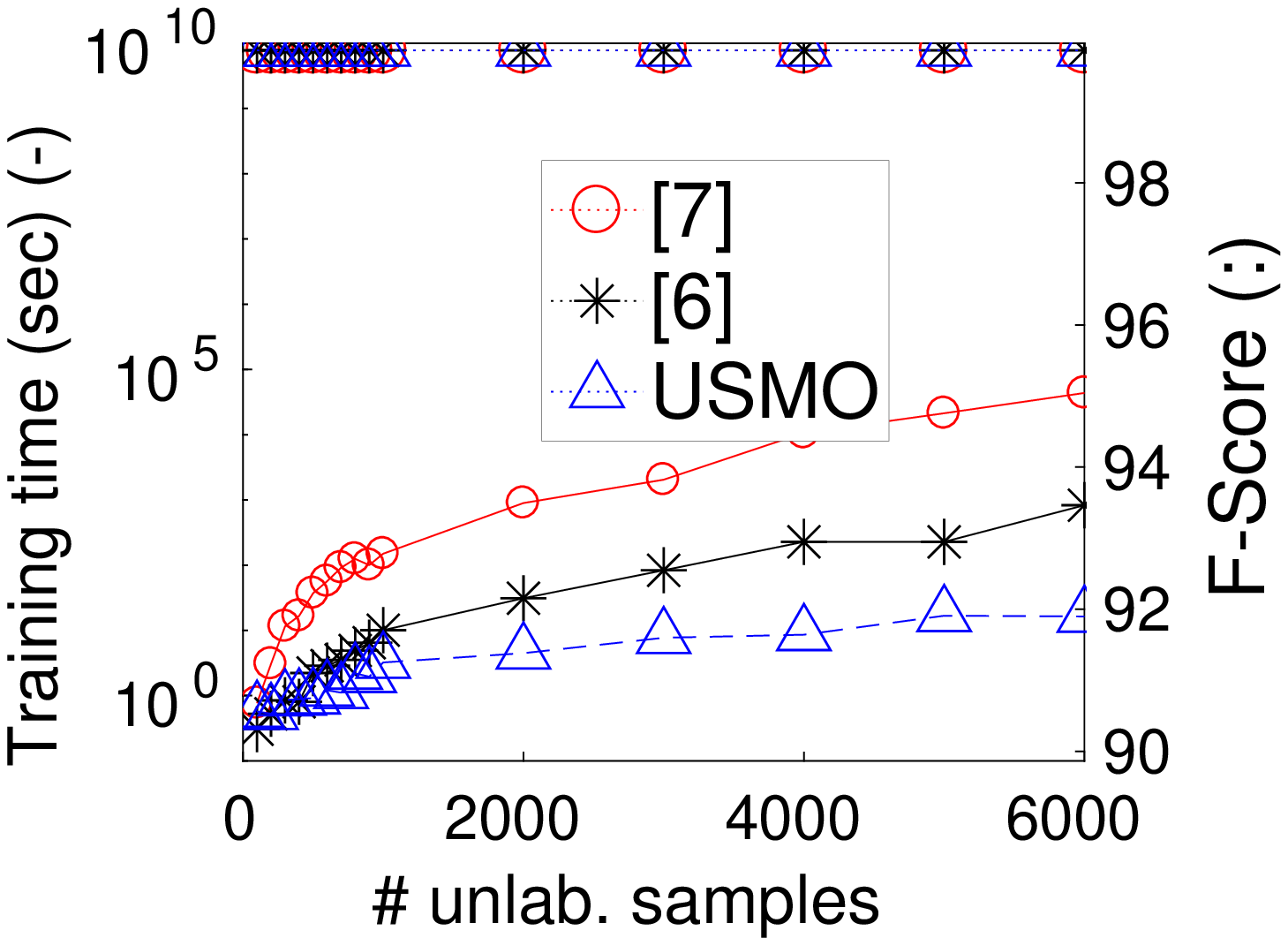}}
\subfloat[Statlog 4 vs. all]{\includegraphics[width=0.16\linewidth]{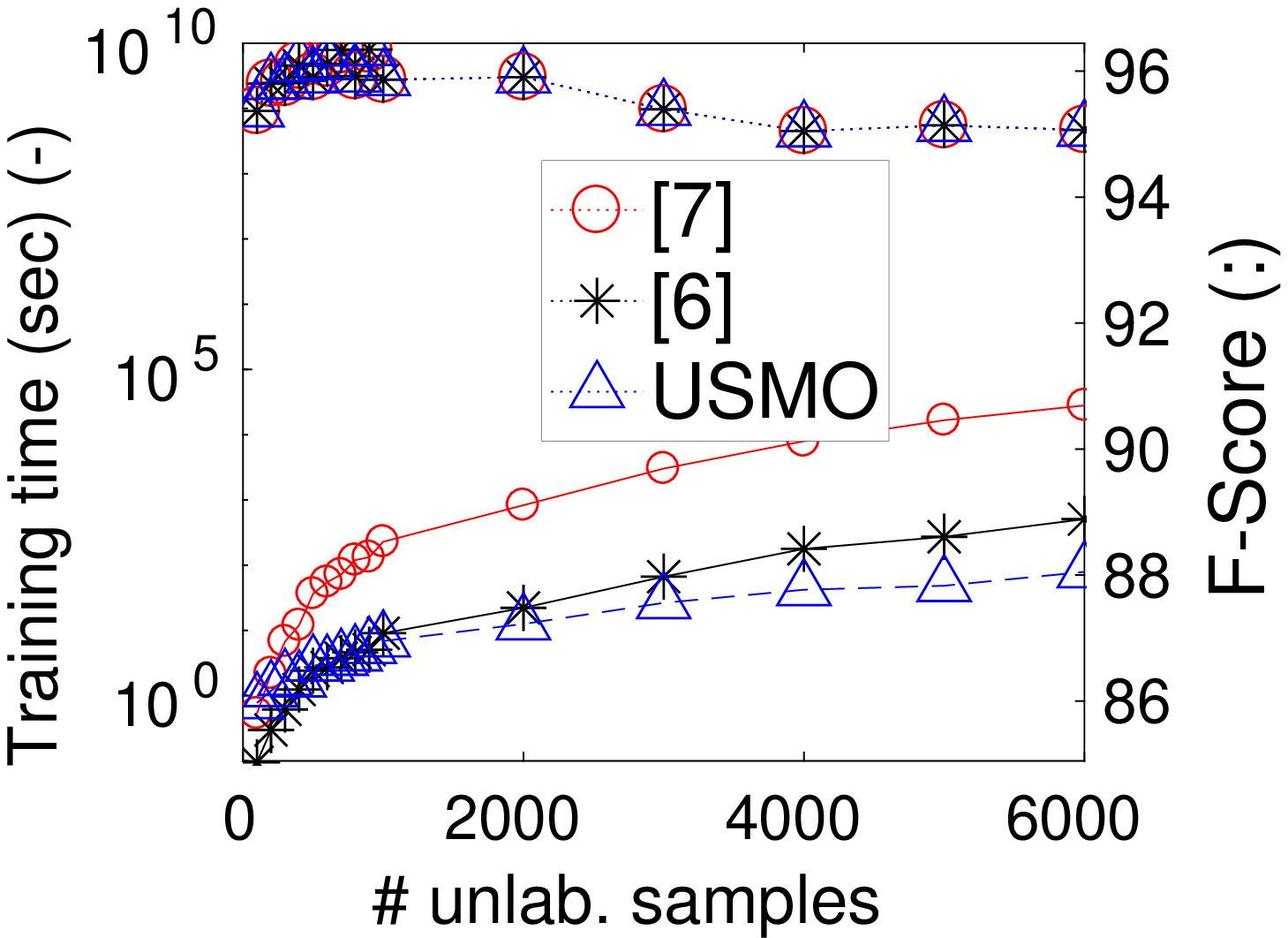}}
\subfloat[Statlog 5 vs. all]{\includegraphics[width=0.16\linewidth]{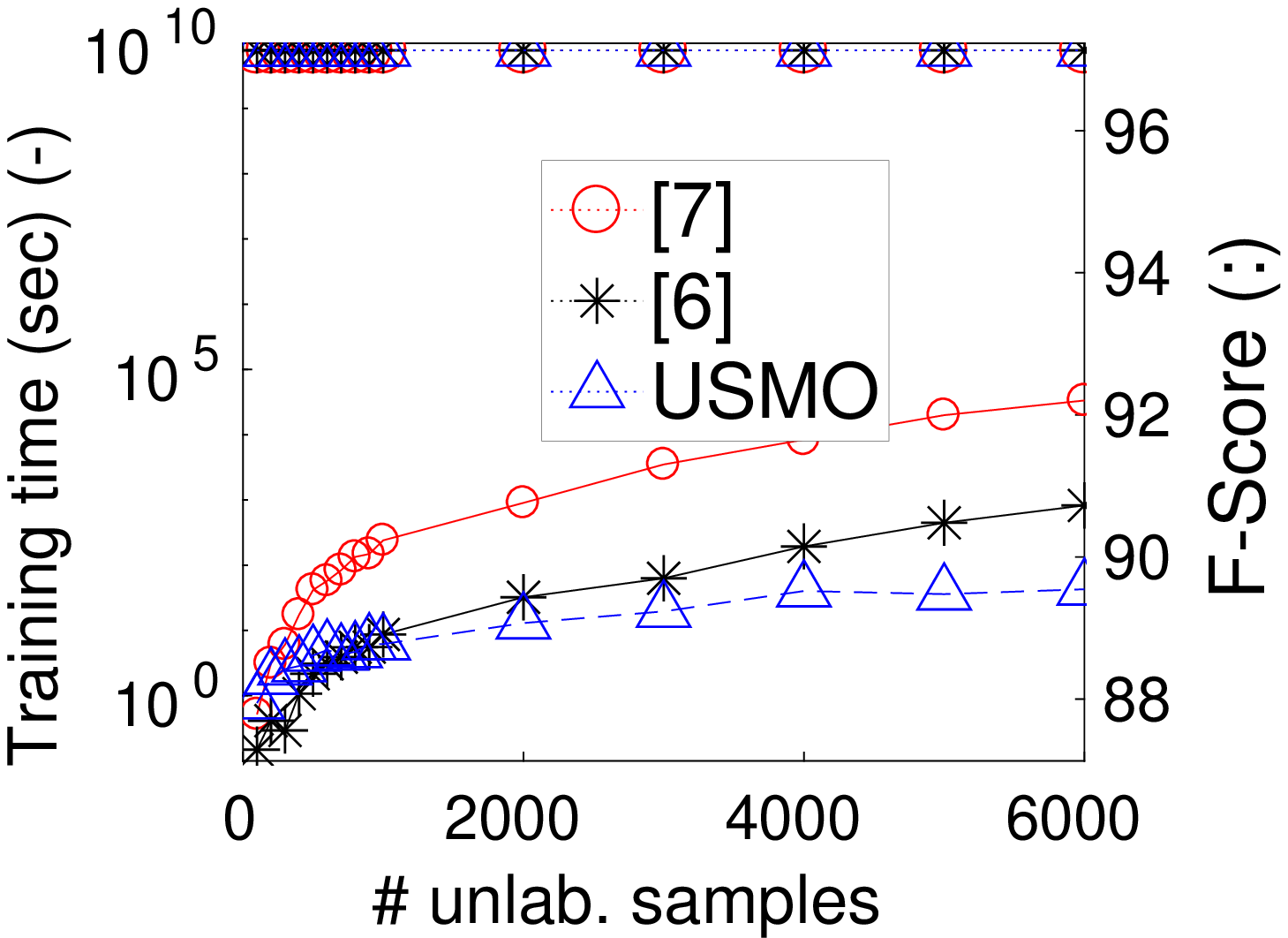}}
\subfloat[Statlog 6 vs. all]{\includegraphics[width=0.16\linewidth]{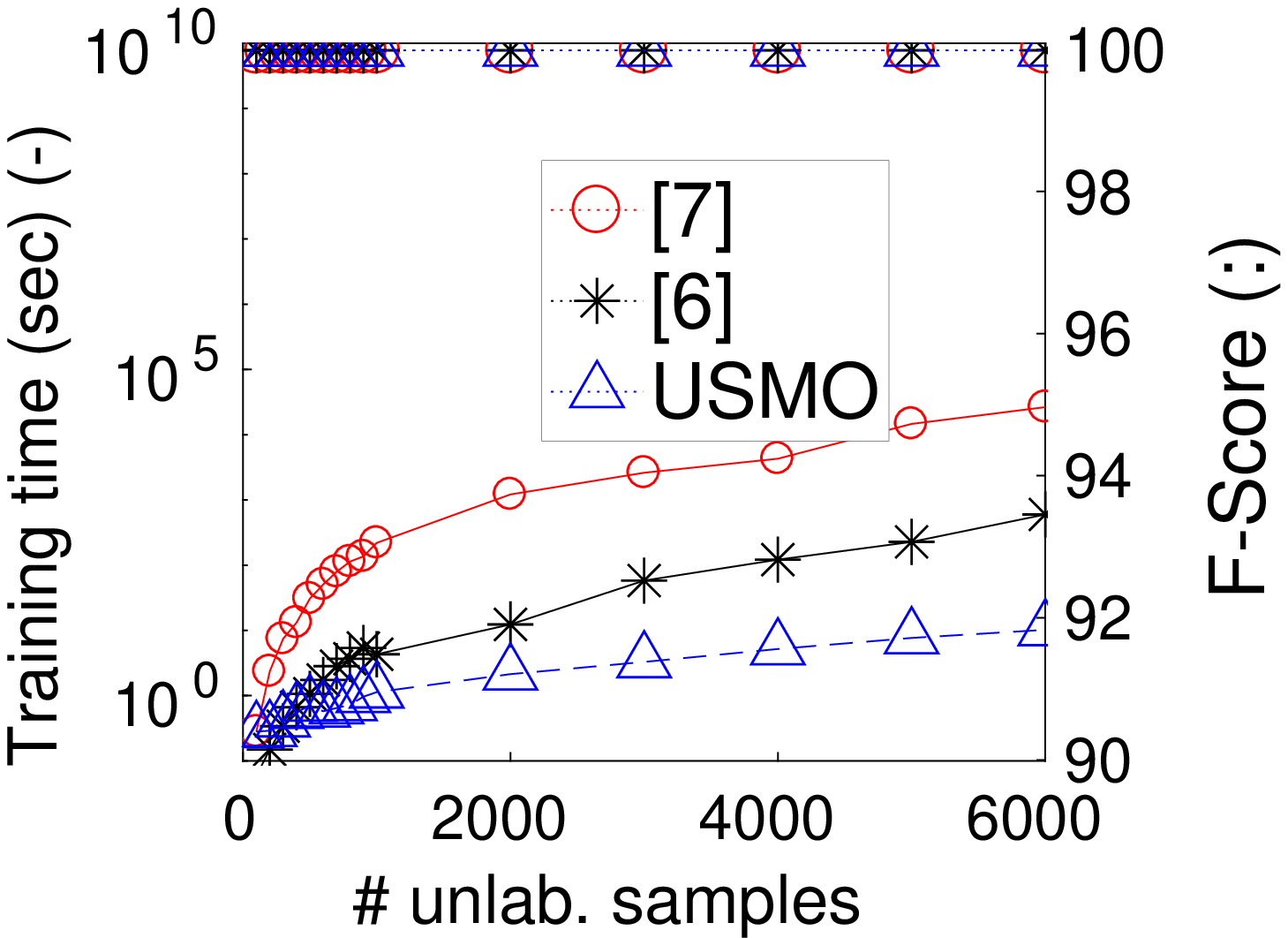}}\\
\subfloat[Statlog 7 vs. all]{\includegraphics[width=0.16\linewidth]{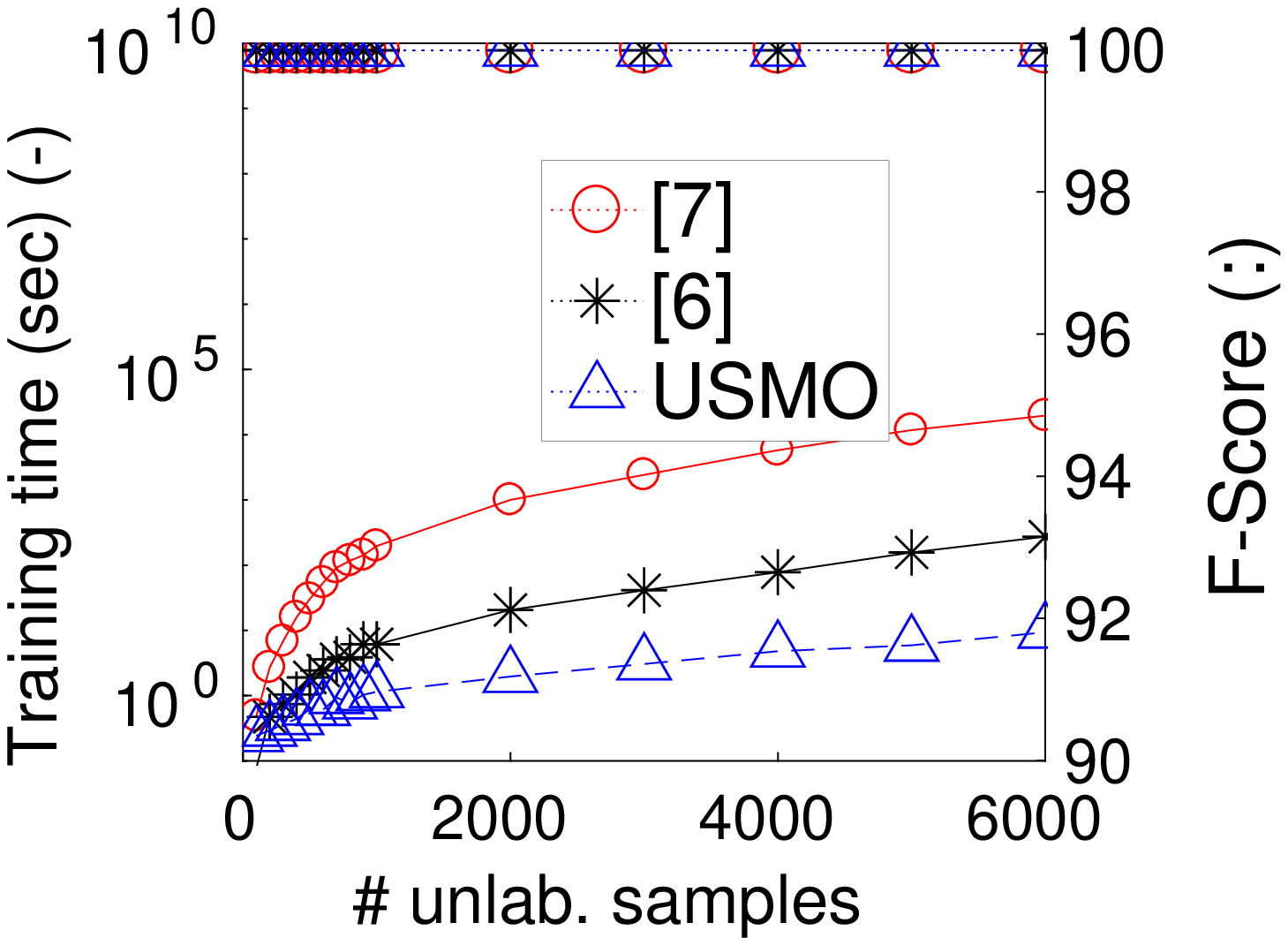}}
\subfloat[MNIST 0 vs. all]{\includegraphics[width=0.16\linewidth]{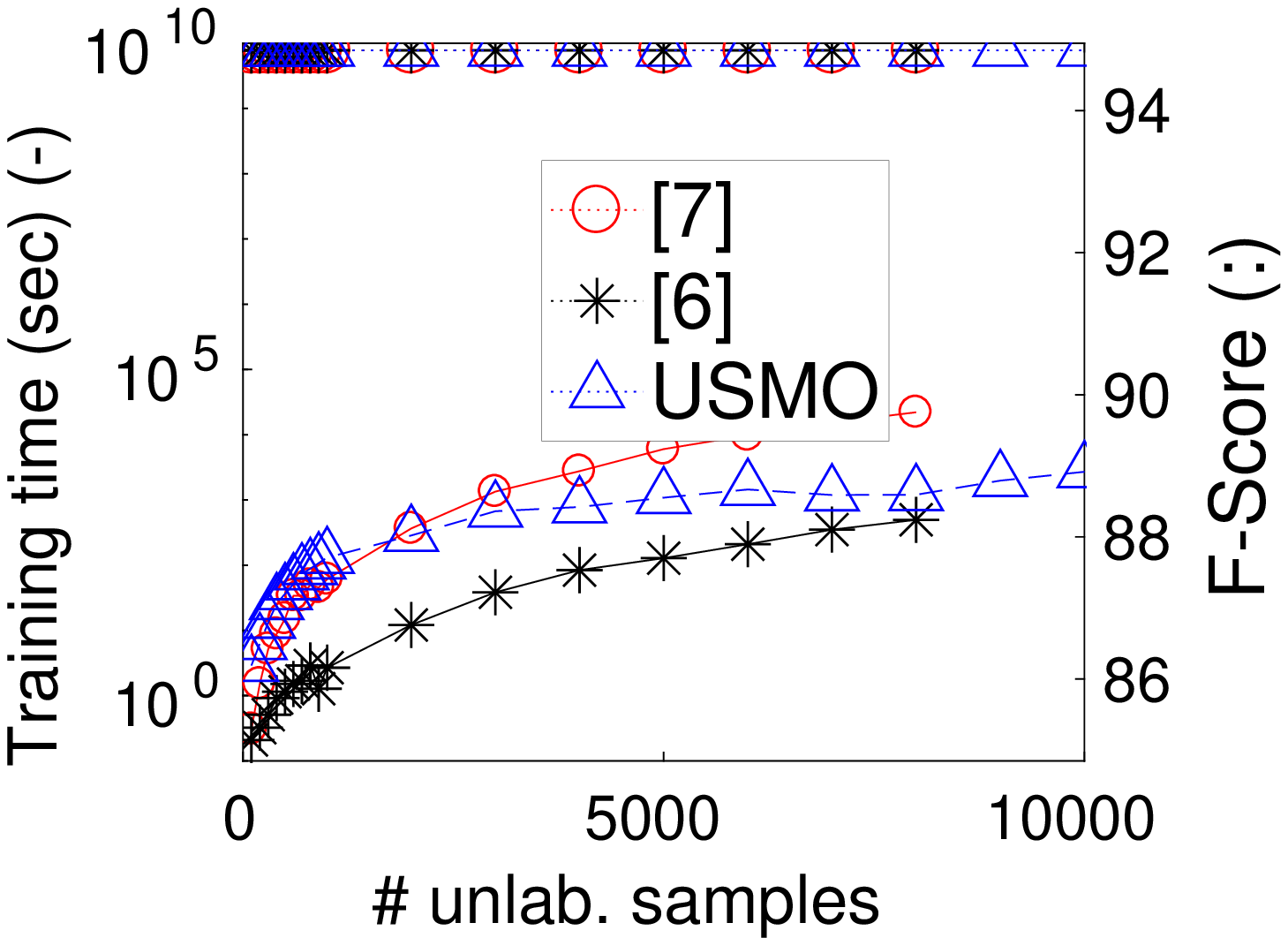}}
\subfloat[MNIST 1 vs. all]{\includegraphics[width=0.16\linewidth]{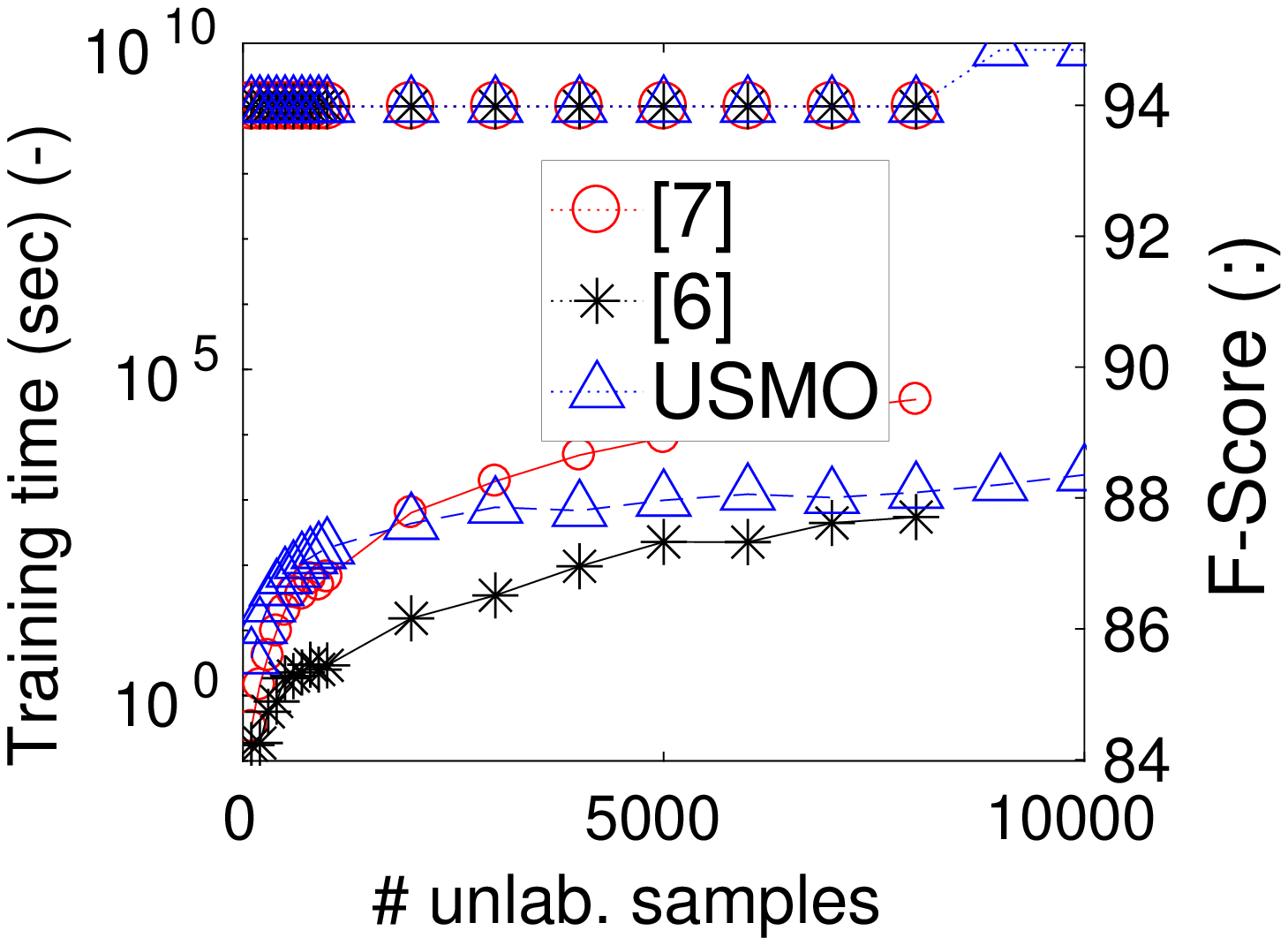}}
\subfloat[MNIST 2 vs. all]{\includegraphics[width=0.16\linewidth]{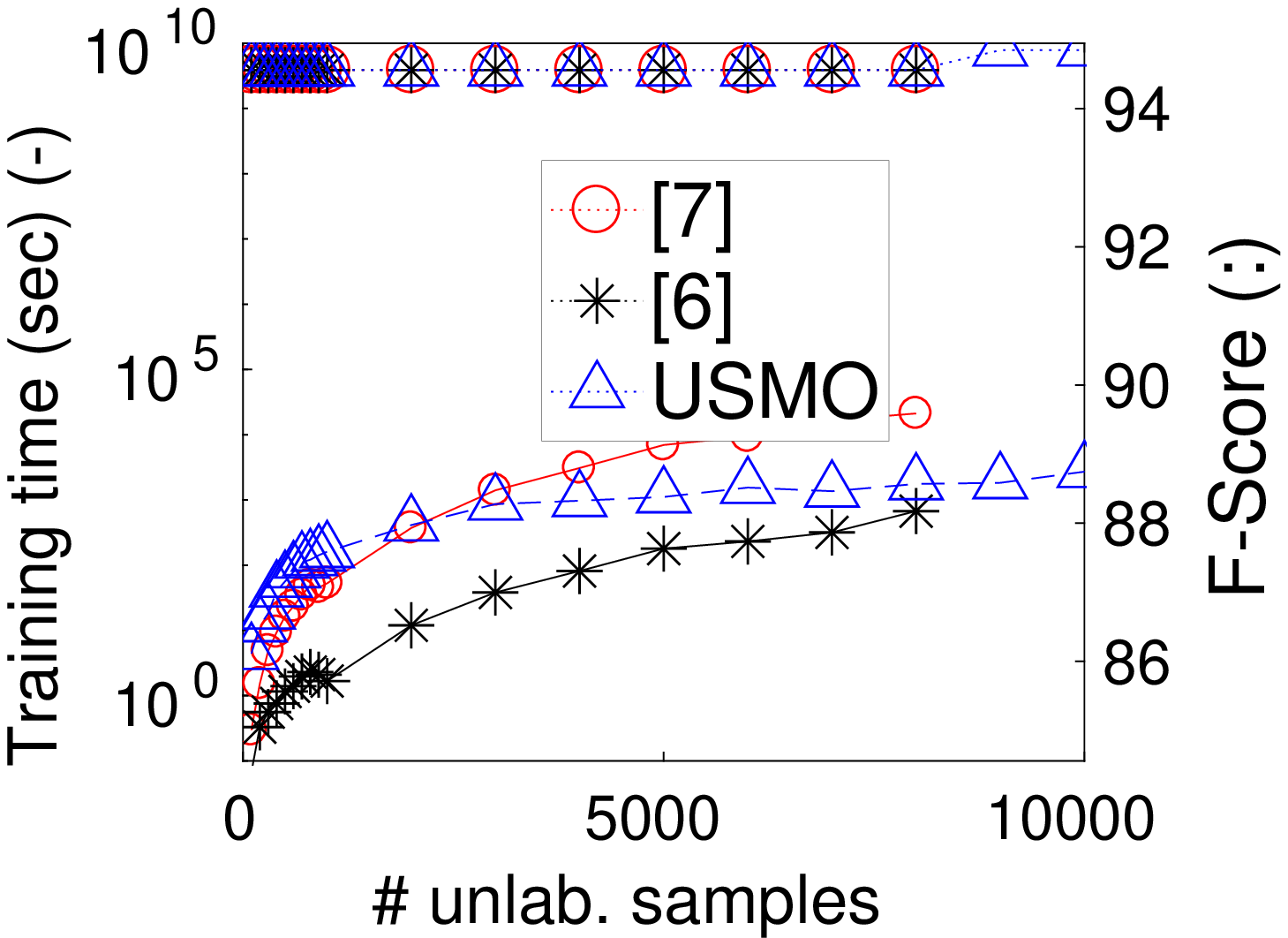}}
\subfloat[MNIST 3 vs. all]{\includegraphics[width=0.16\linewidth]{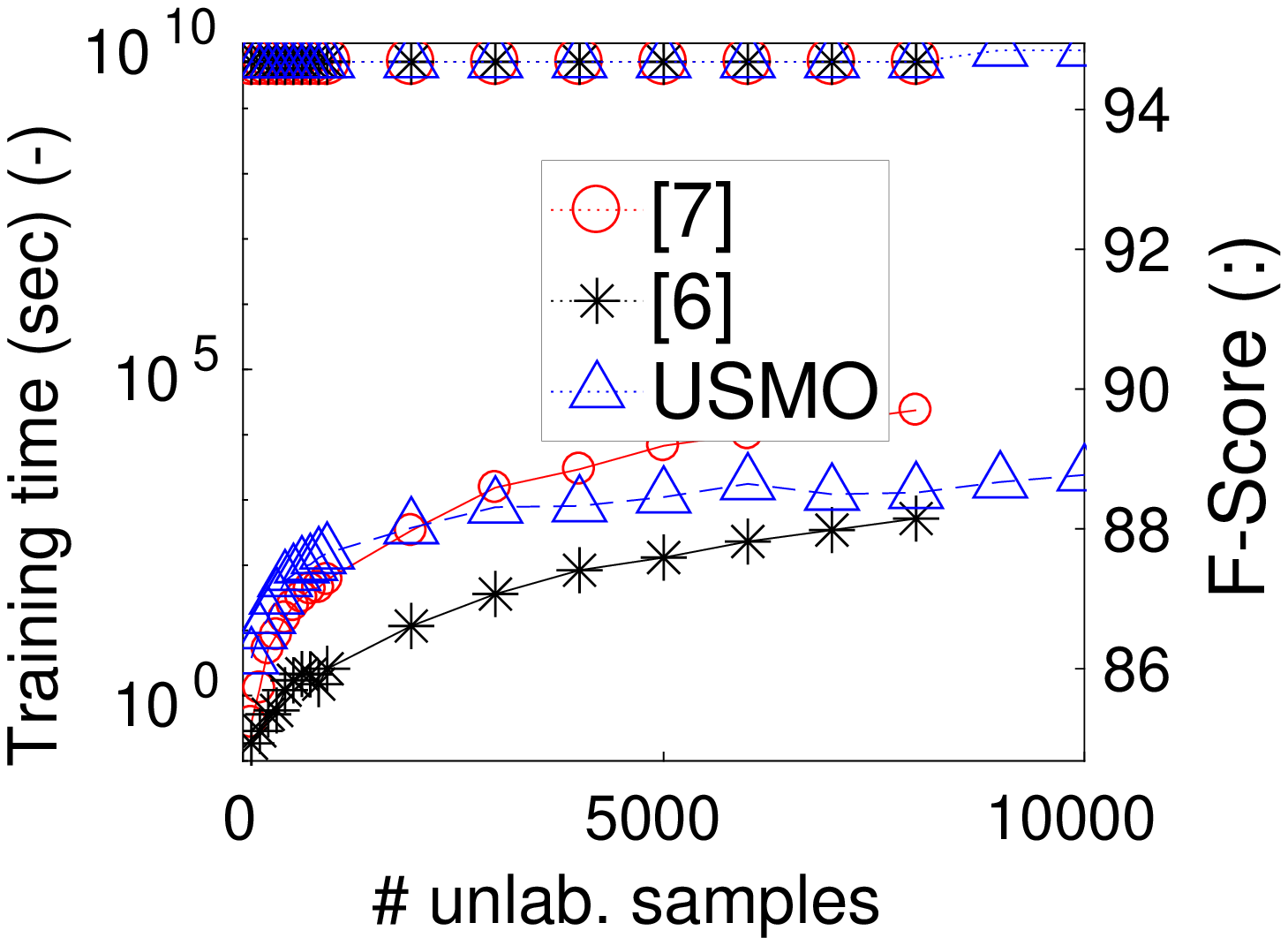}}
\subfloat[MNIST 4 vs. all]{\includegraphics[width=0.16\linewidth]{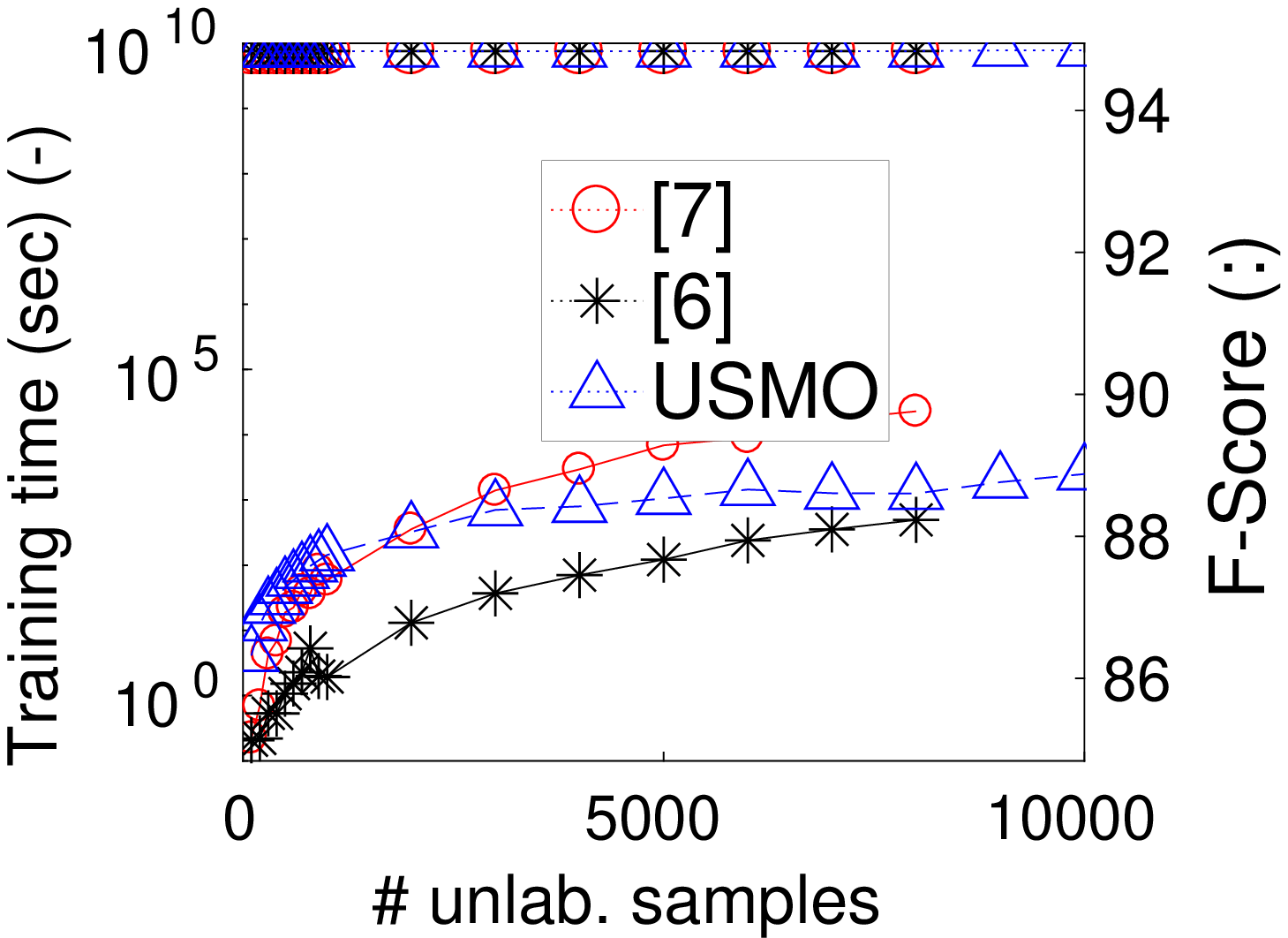}}\\
\subfloat[MNIST 5 vs. all]{\includegraphics[width=0.16\linewidth]{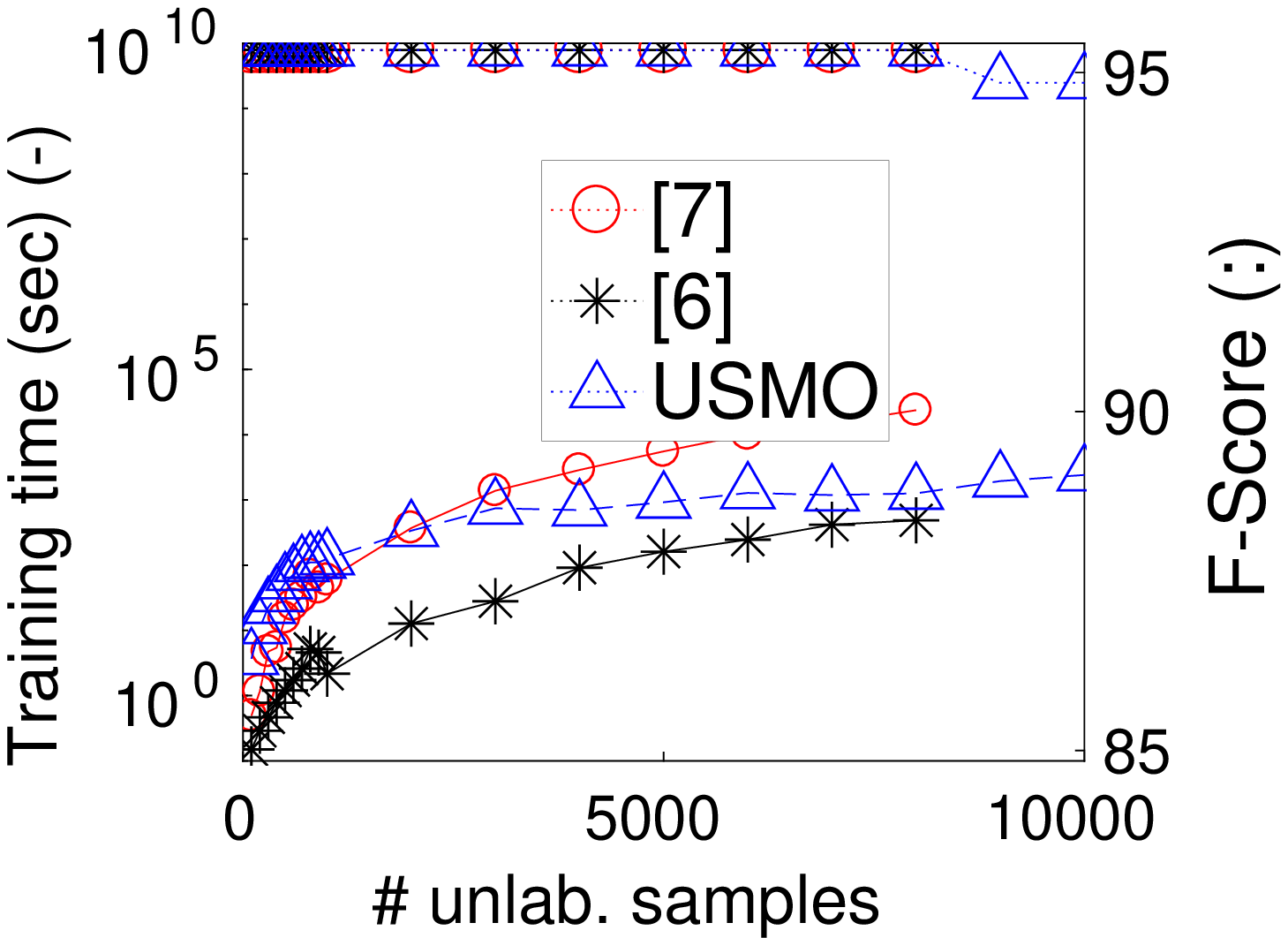}}
\subfloat[MNIST 6 vs. all]{\includegraphics[width=0.16\linewidth]{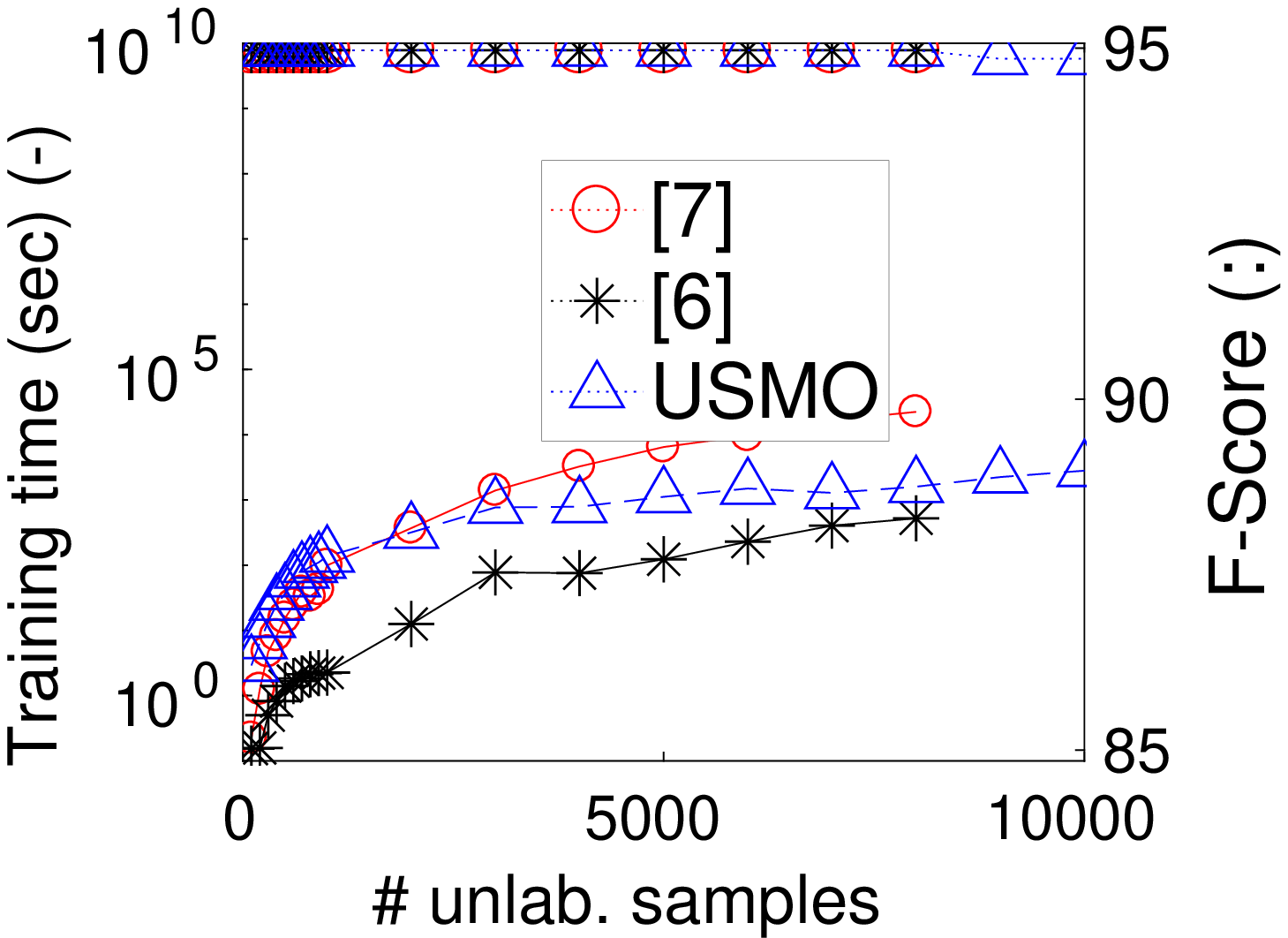}}
\subfloat[MNIST 7 vs. all]{\includegraphics[width=0.16\linewidth]{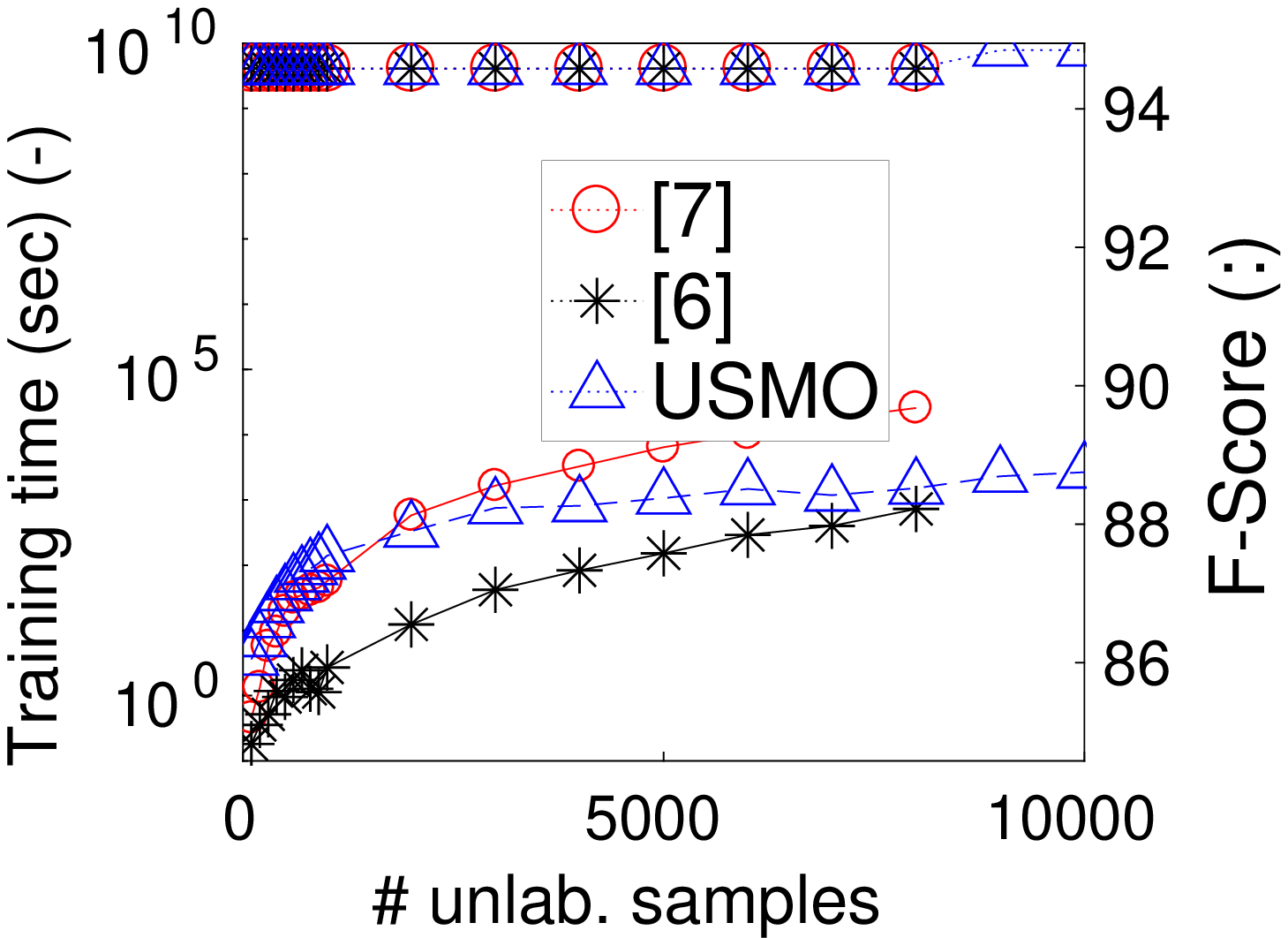}}
\subfloat[MNIST 8 vs. all]{\includegraphics[width=0.16\linewidth]{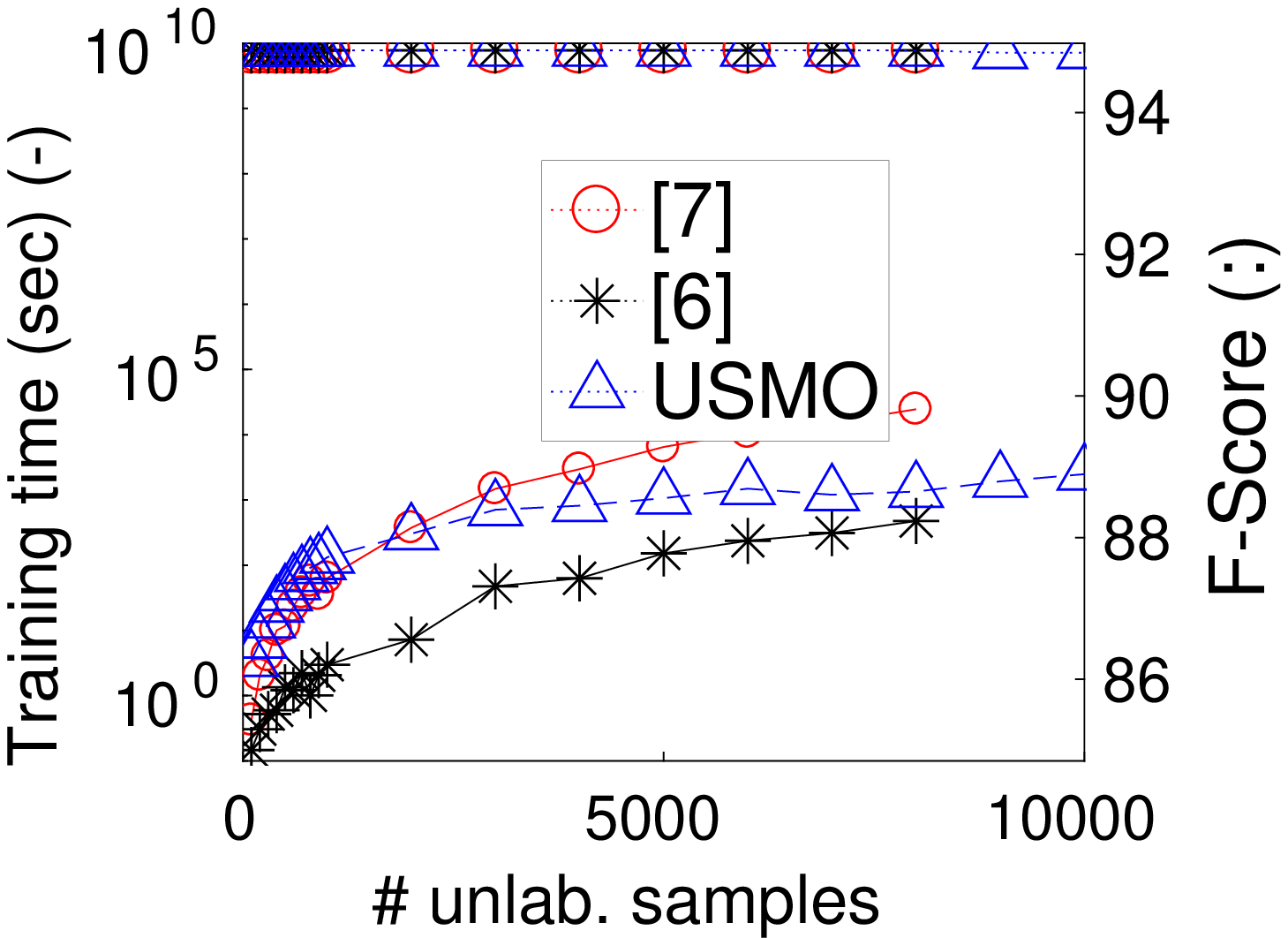}}
\subfloat[MNIST 9 vs. all]{\includegraphics[width=0.16\linewidth]{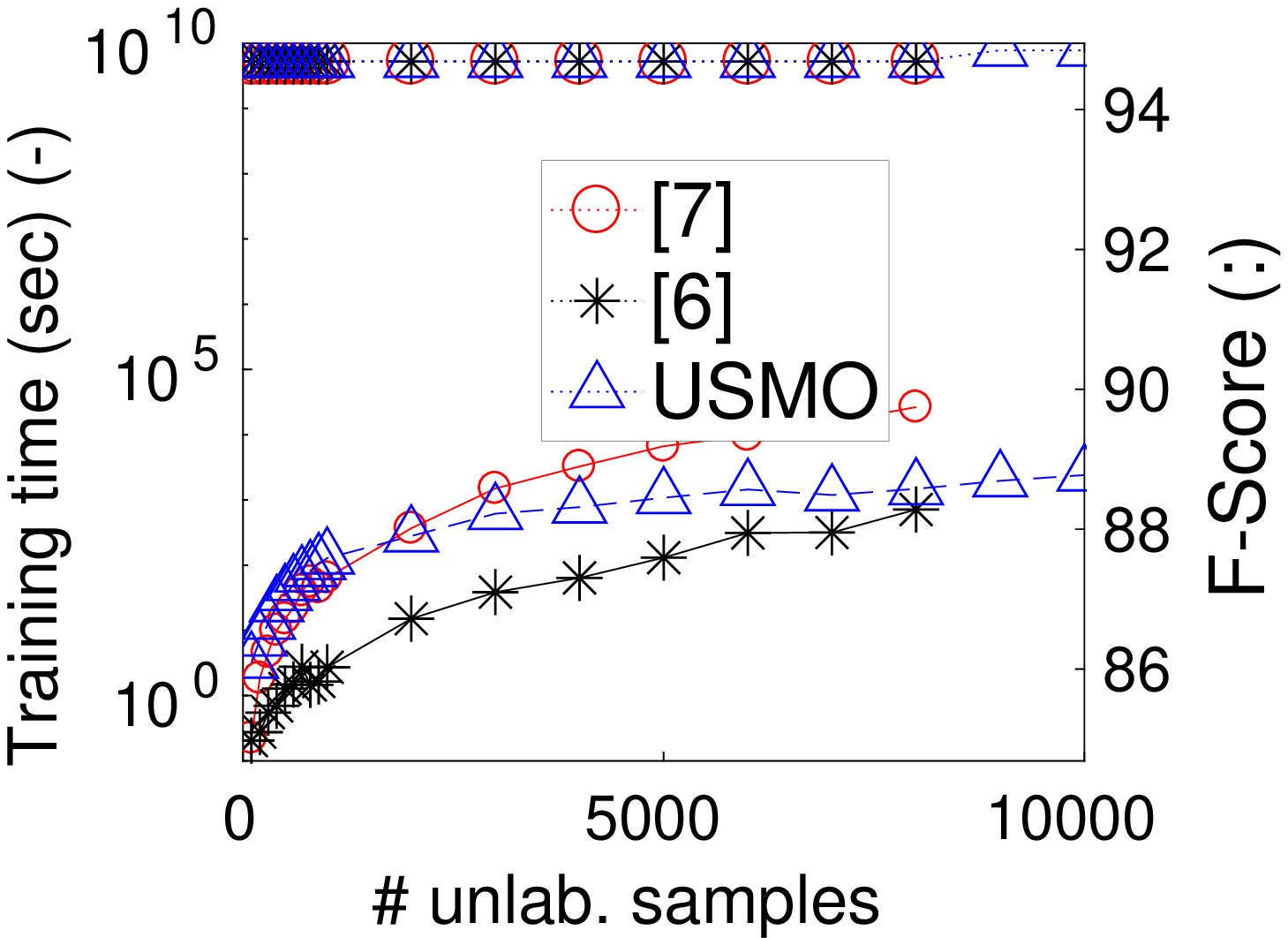}}
\caption{Comparative results on (a)-(g) Statlog (shuttle) and (h)-(q) MNIST datasets using the 
Gaussian kernel ($\lambda = 0.01$ and scale parameter equal to 1). 
Each plot shows the training time against different number of 
unlabeled samples ($100$ positive samples) as well as the generalization performance on the test set.}
\label{fig:results_large1_gaussian}
\end{figure*}

\begin{figure*}[h!]
\centering
\subfloat[Bank 1 vs. all]{\includegraphics[width=0.16\linewidth]{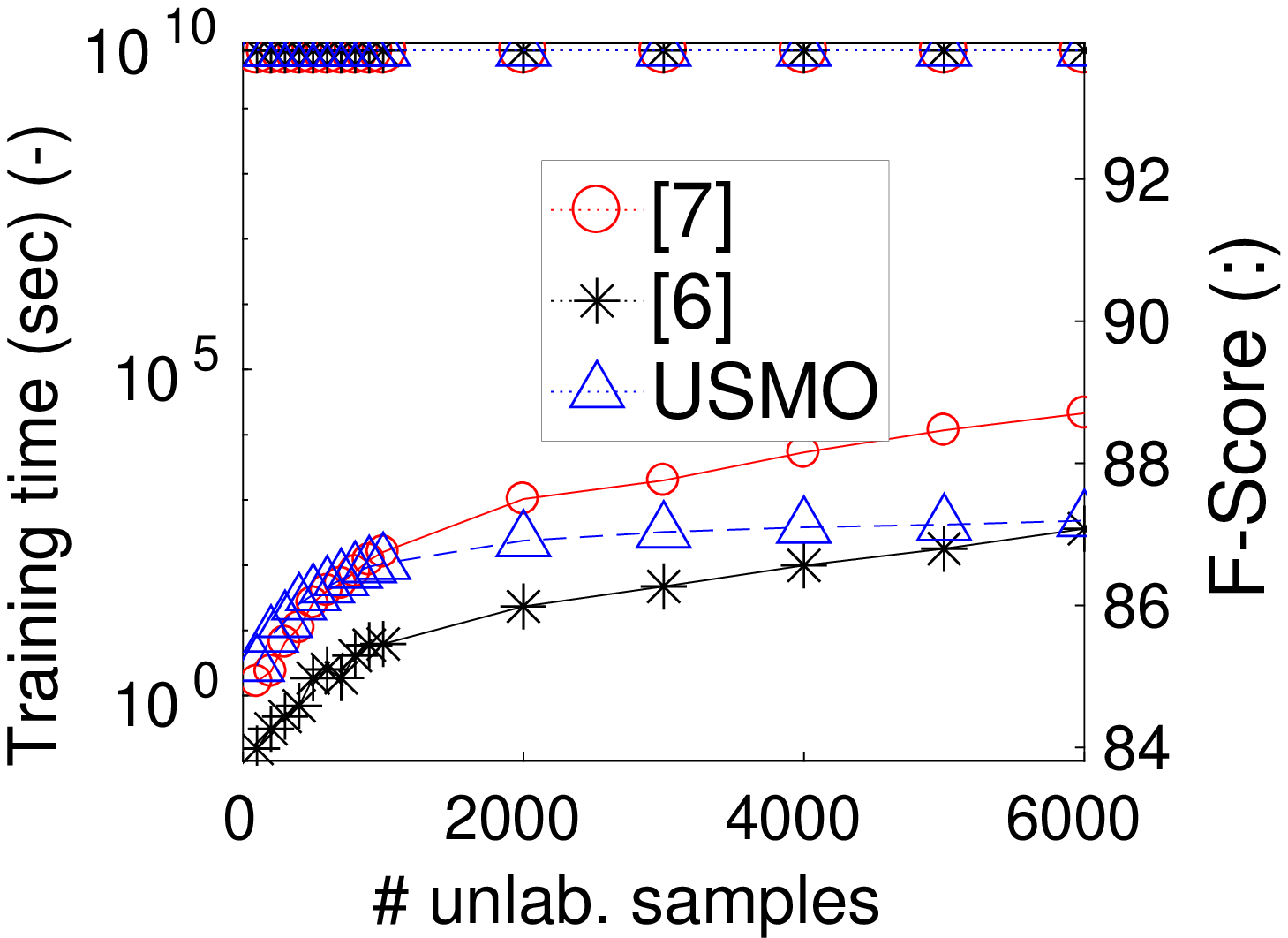}}
\subfloat[Adult 1 vs. all]{\includegraphics[width=0.16\linewidth]{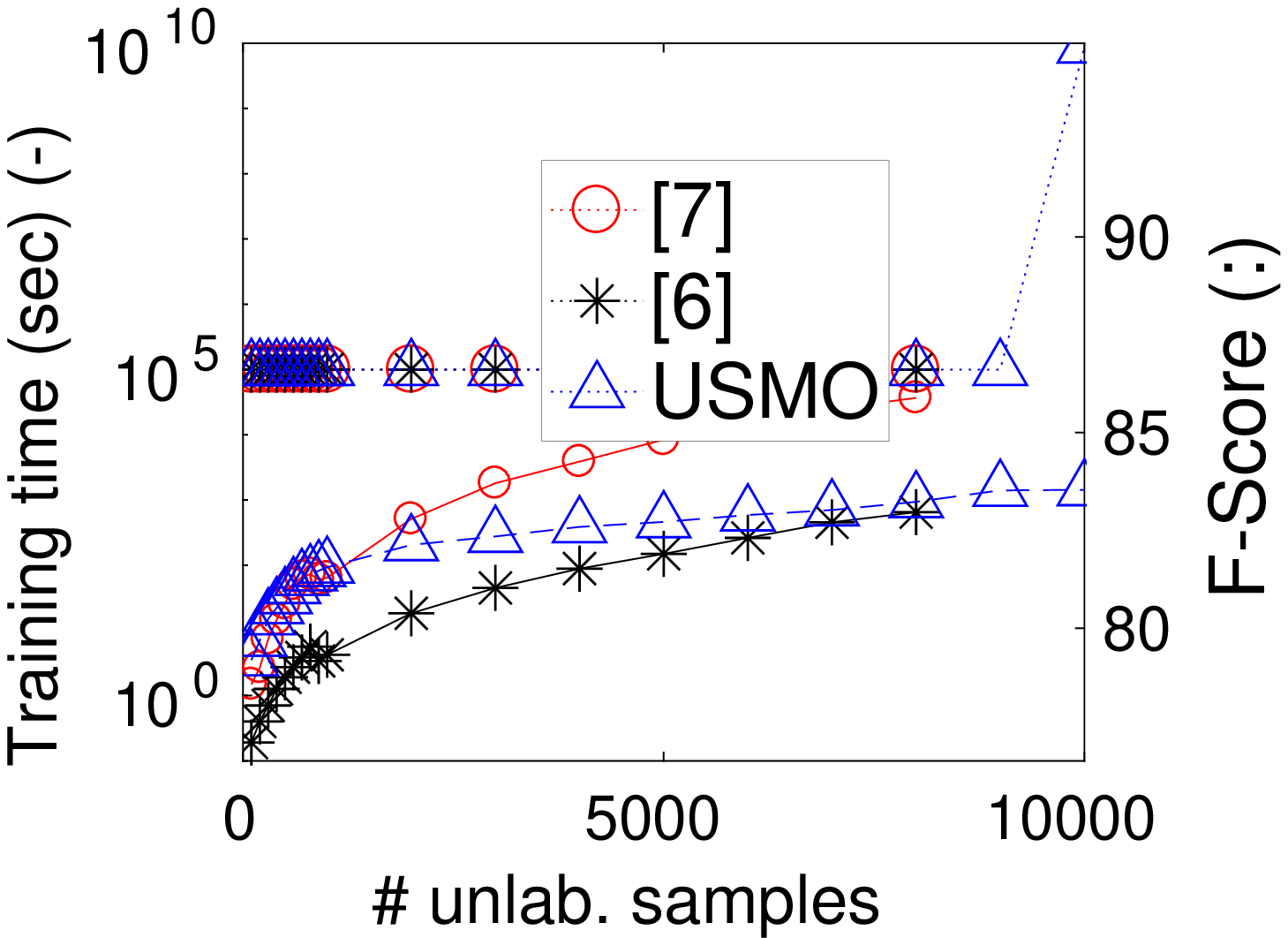}}
\subfloat[POKER 0 vs. all]{\includegraphics[width=0.16\linewidth]{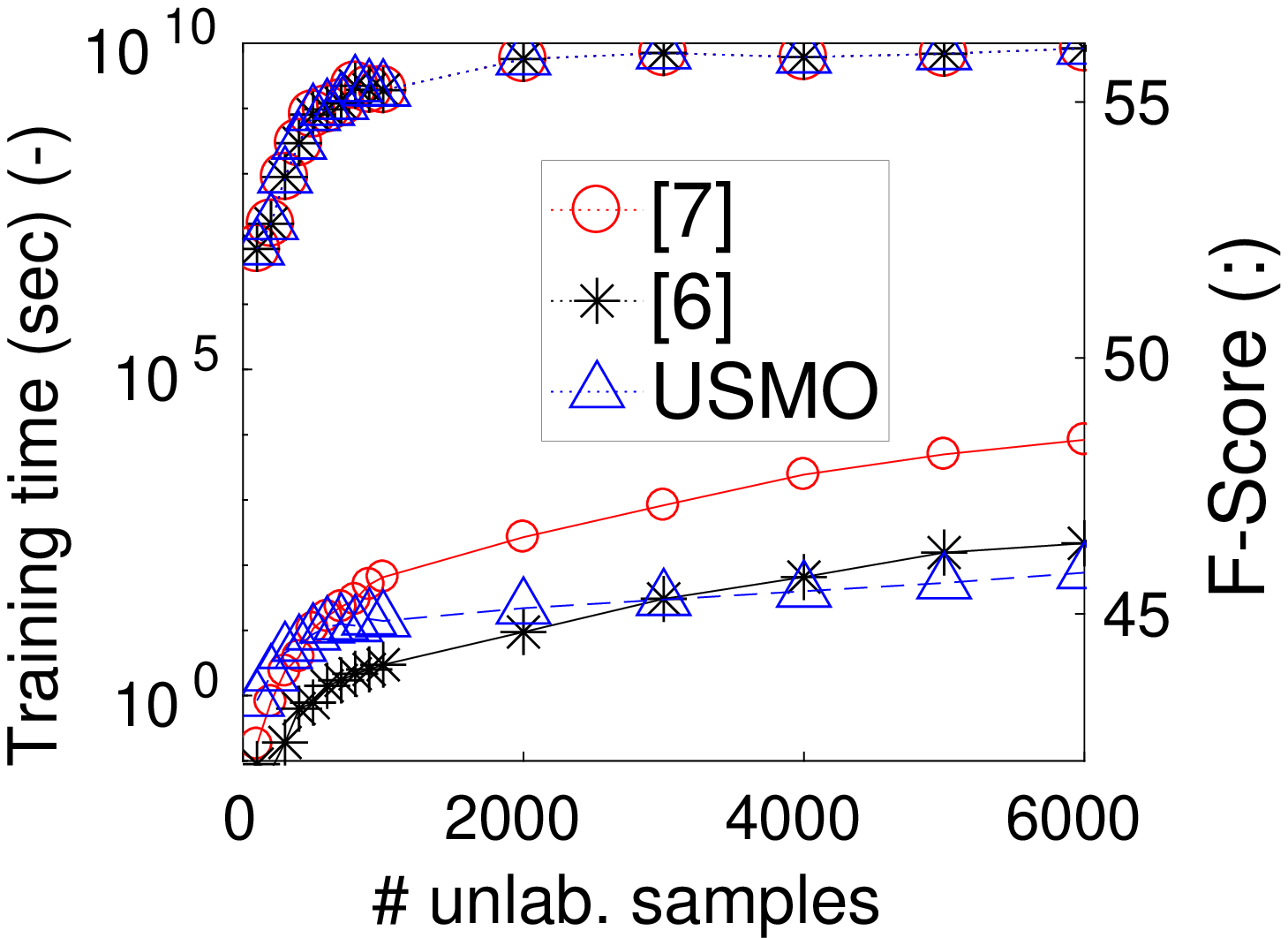}}
\subfloat[POKER 1 vs. all]{\includegraphics[width=0.16\linewidth]{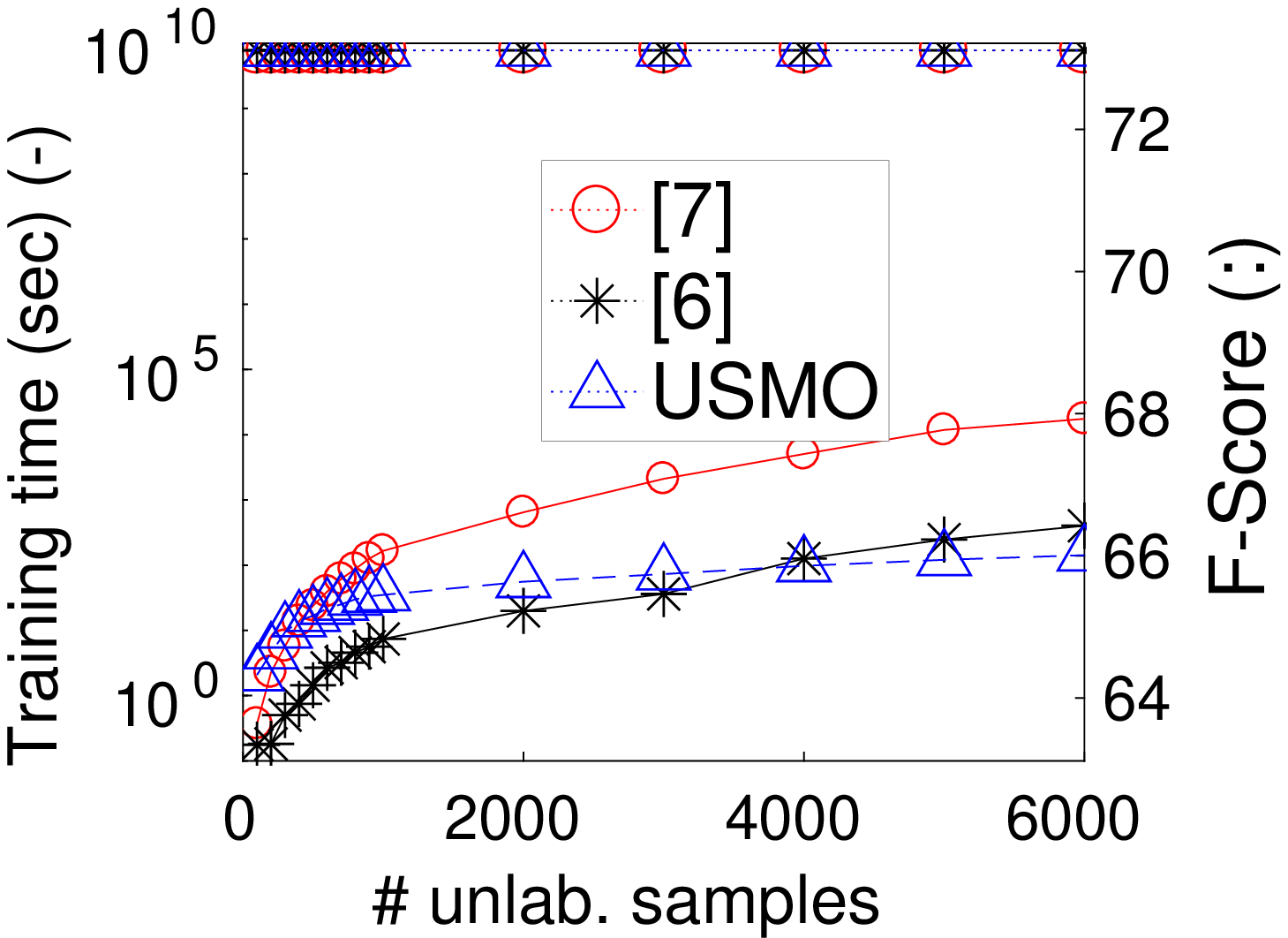}}
\subfloat[POKER 2 vs. all]{\includegraphics[width=0.16\linewidth]{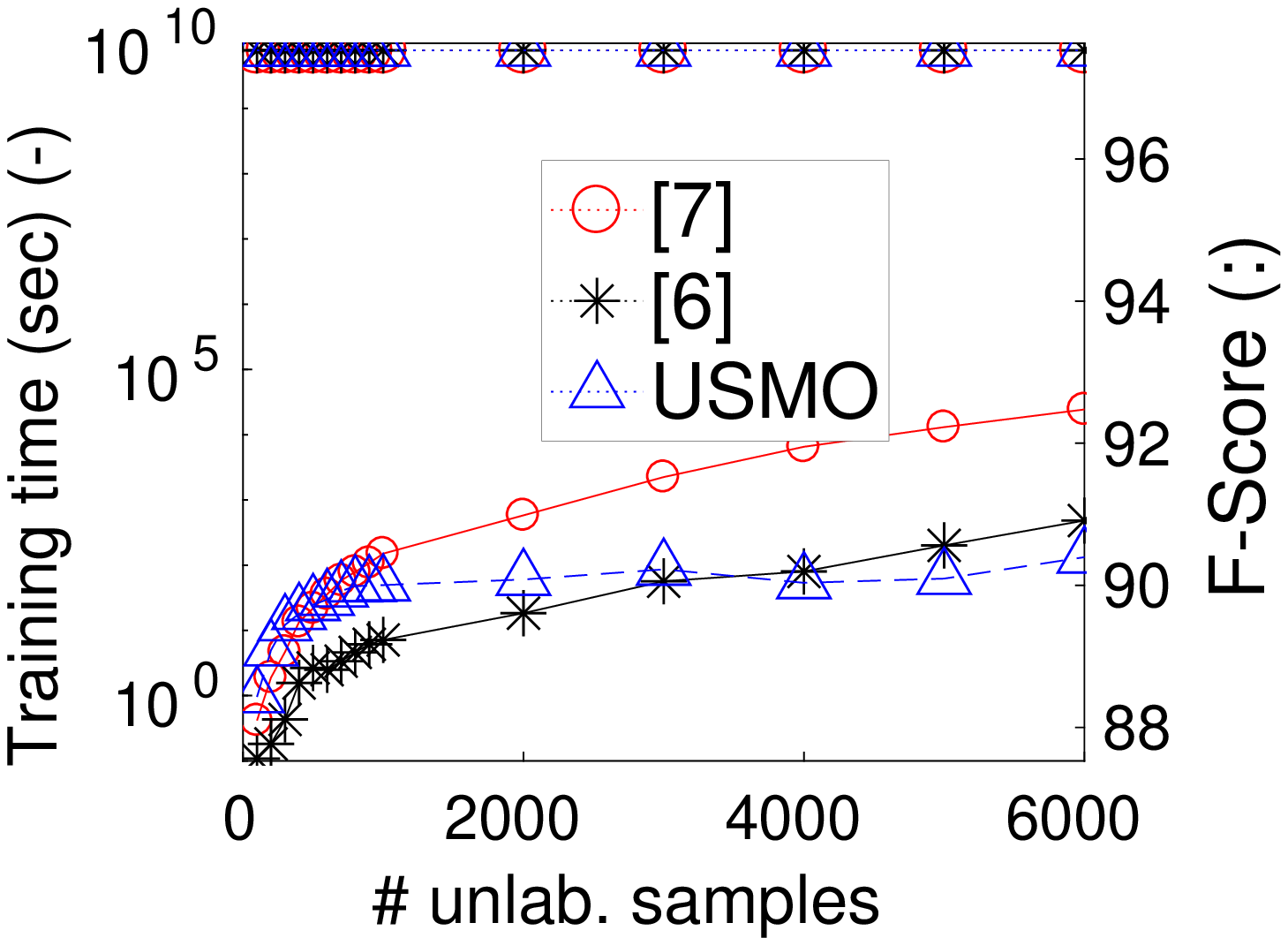}}
\subfloat[POKER 3 vs. all]{\includegraphics[width=0.16\linewidth]{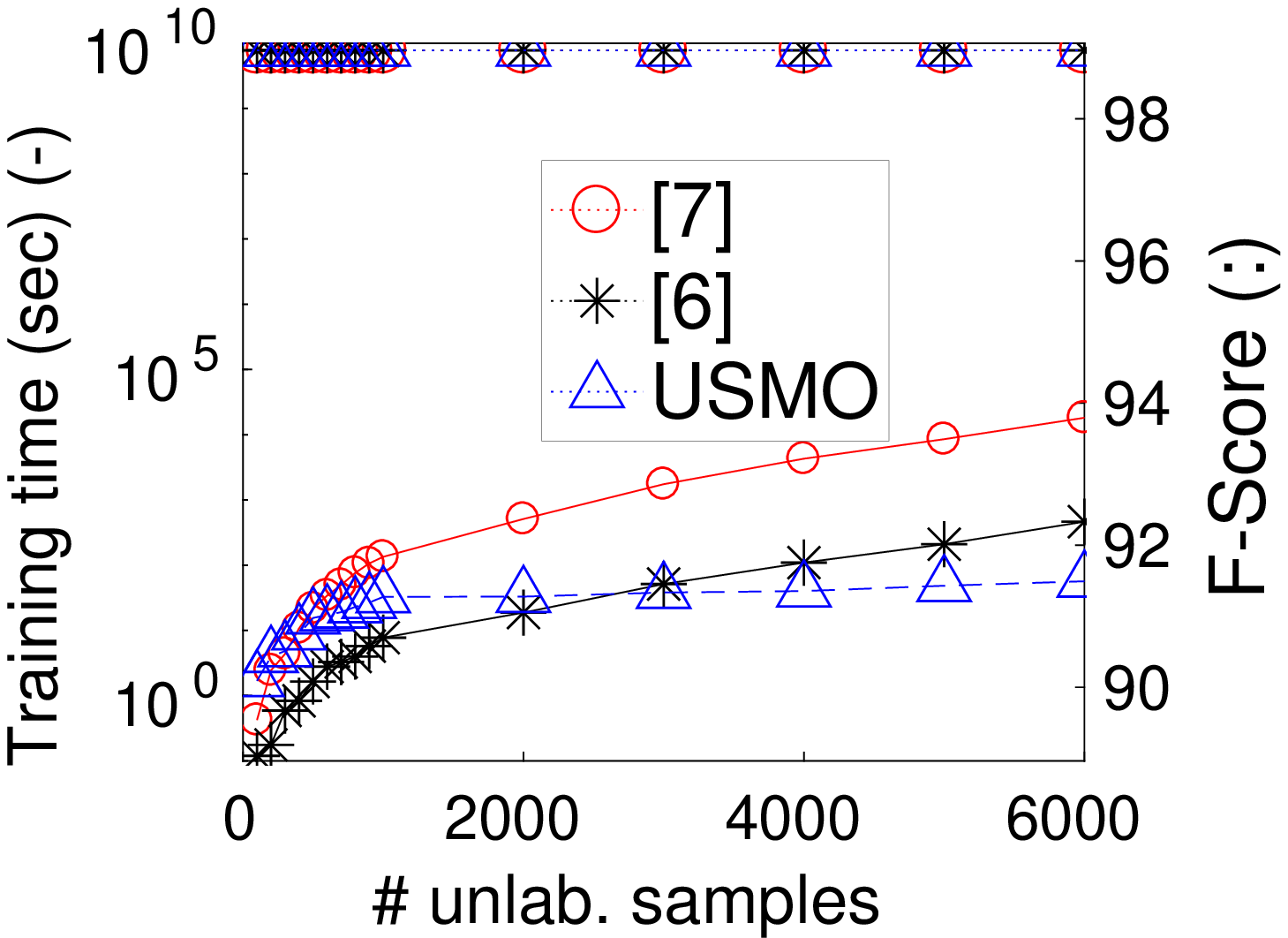}}\\
\subfloat[POKER 4 vs. all]{\includegraphics[width=0.16\linewidth]{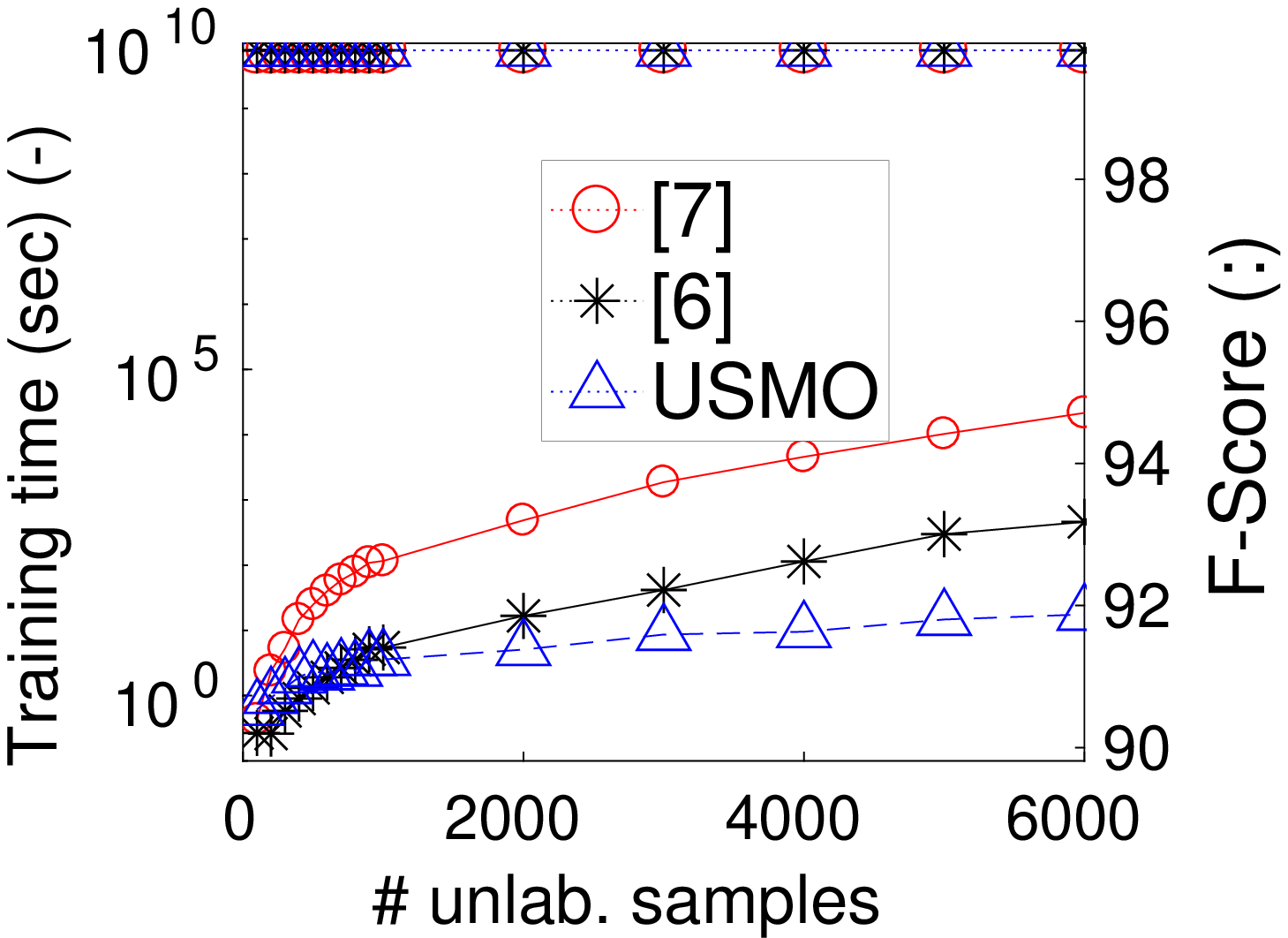}}
\subfloat[POKER 5 vs. all]{\includegraphics[width=0.16\linewidth]{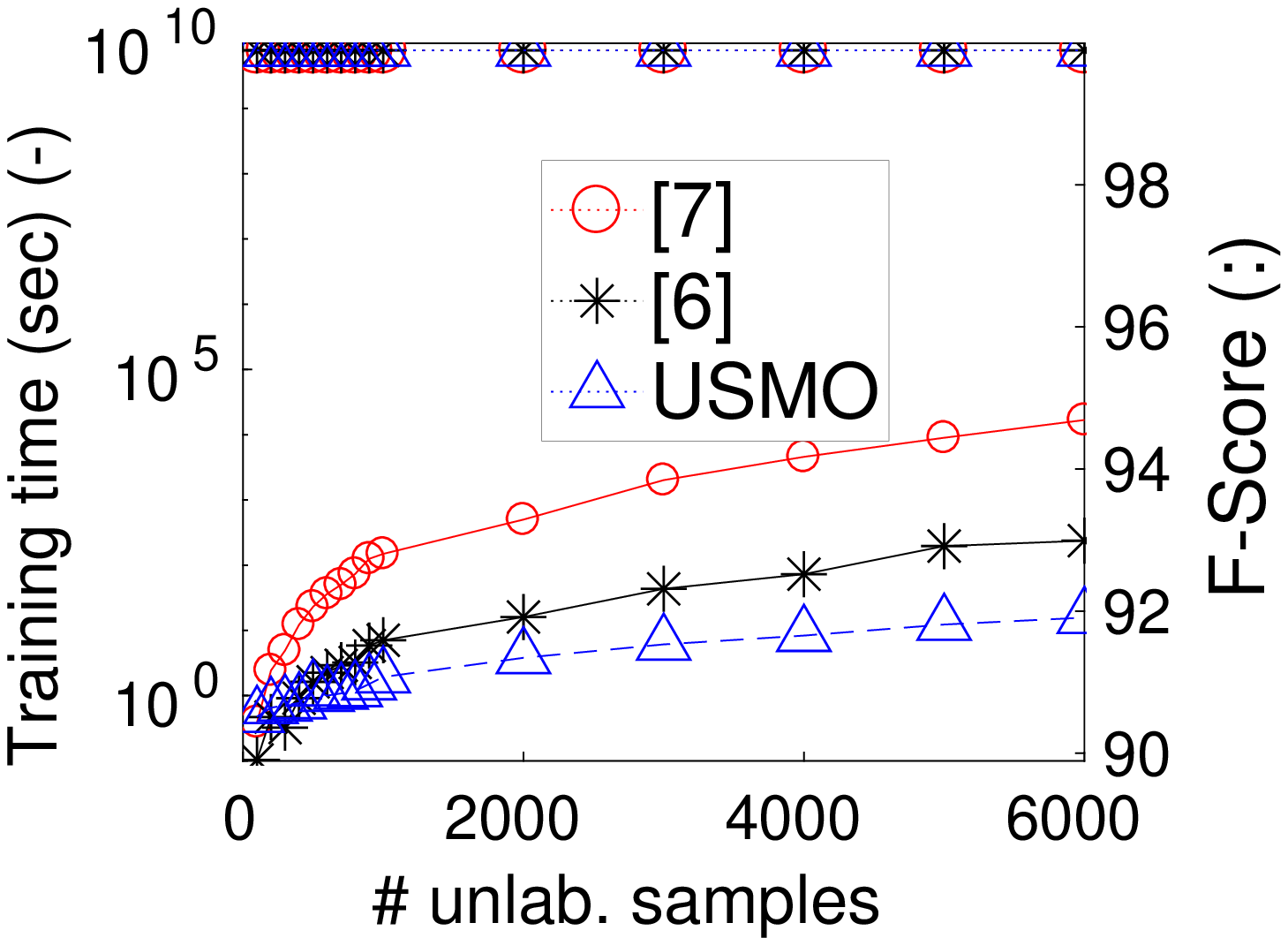}}
\subfloat[POKER 6 vs. all]{\includegraphics[width=0.16\linewidth]{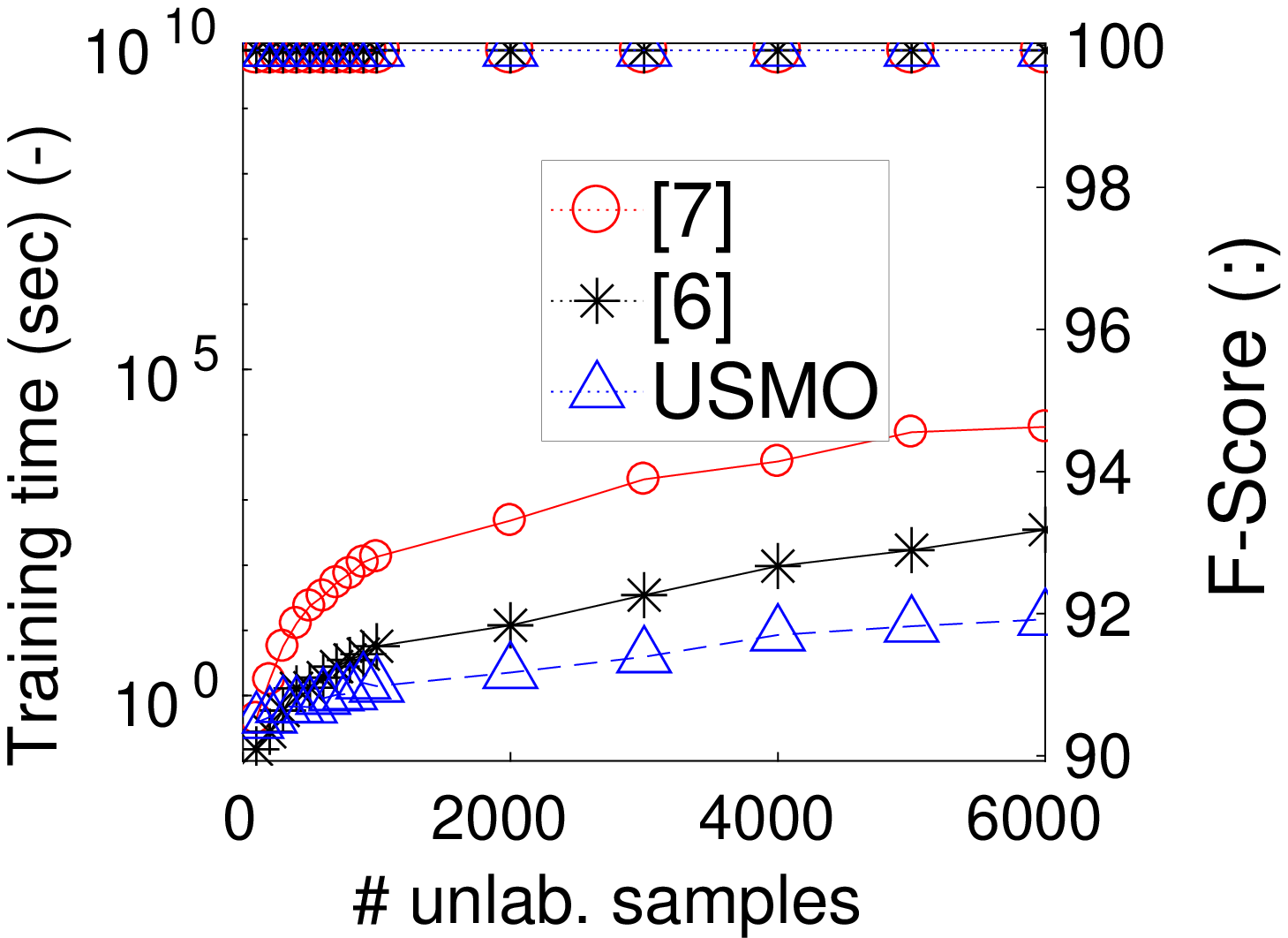}}
\subfloat[POKER 7 vs. all]{\includegraphics[width=0.16\linewidth]{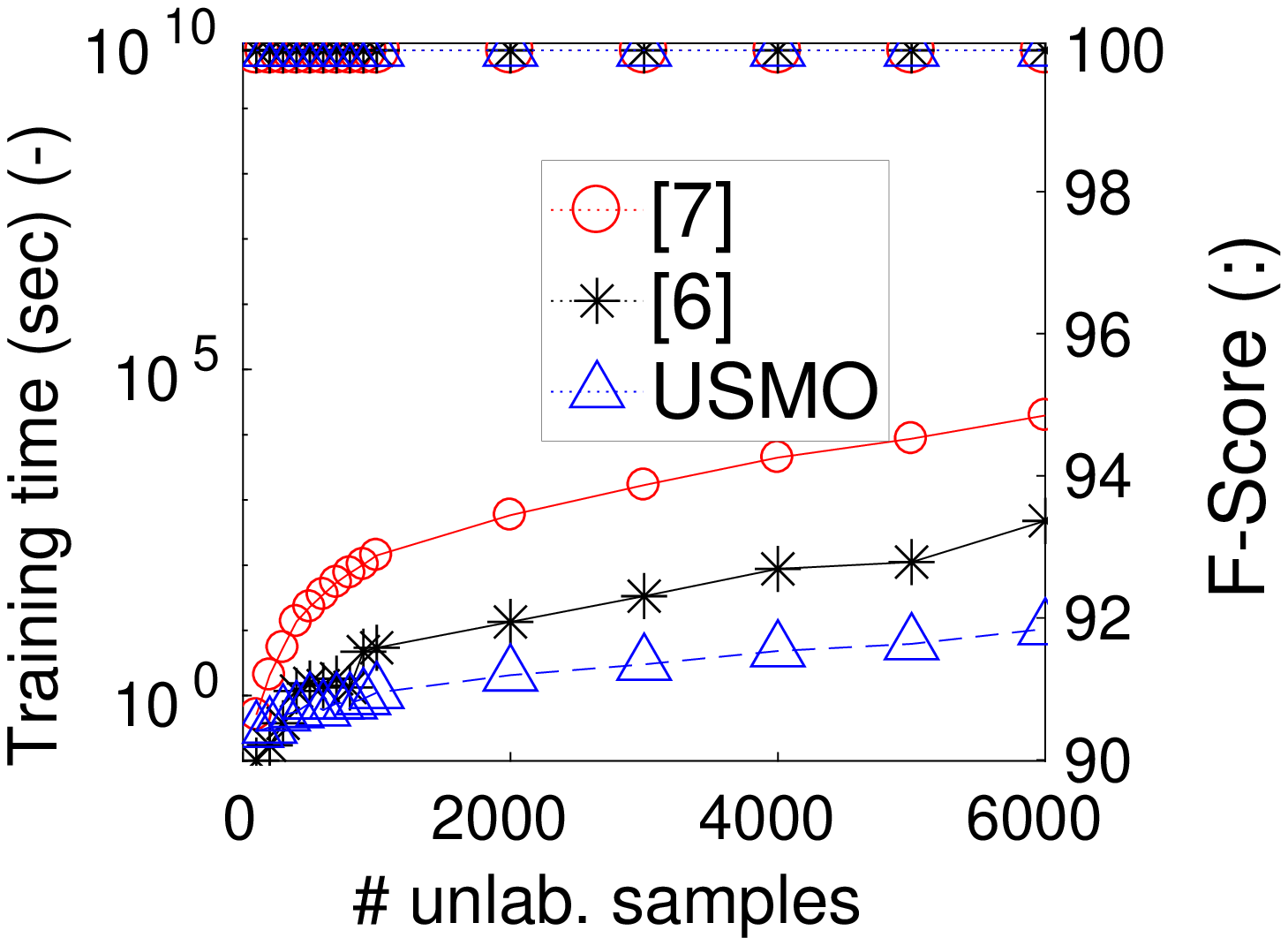}}
\subfloat[POKER 8 vs. all]{\includegraphics[width=0.16\linewidth]{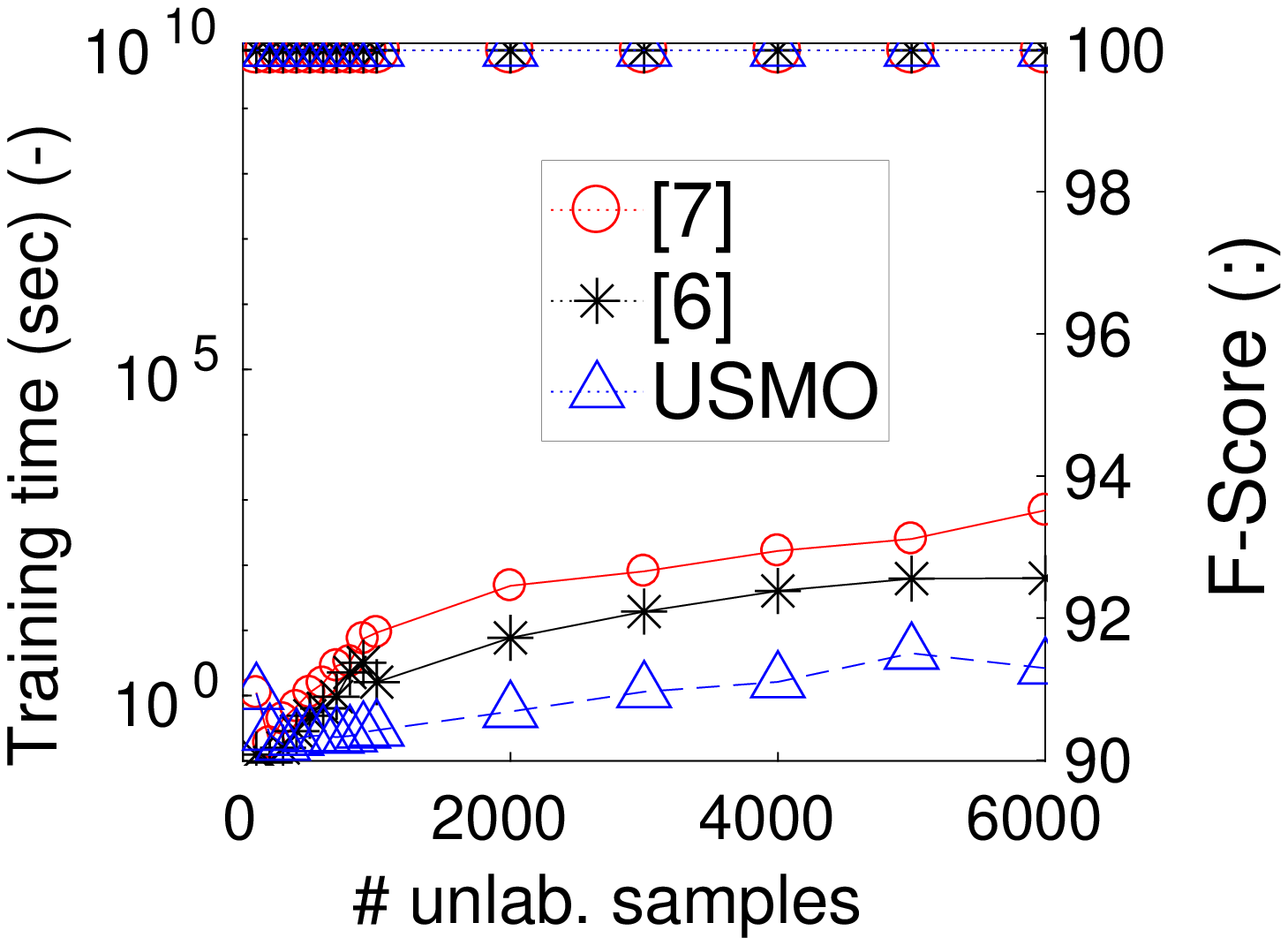}}
\subfloat[POKER 9 vs. all]{\includegraphics[width=0.16\linewidth]{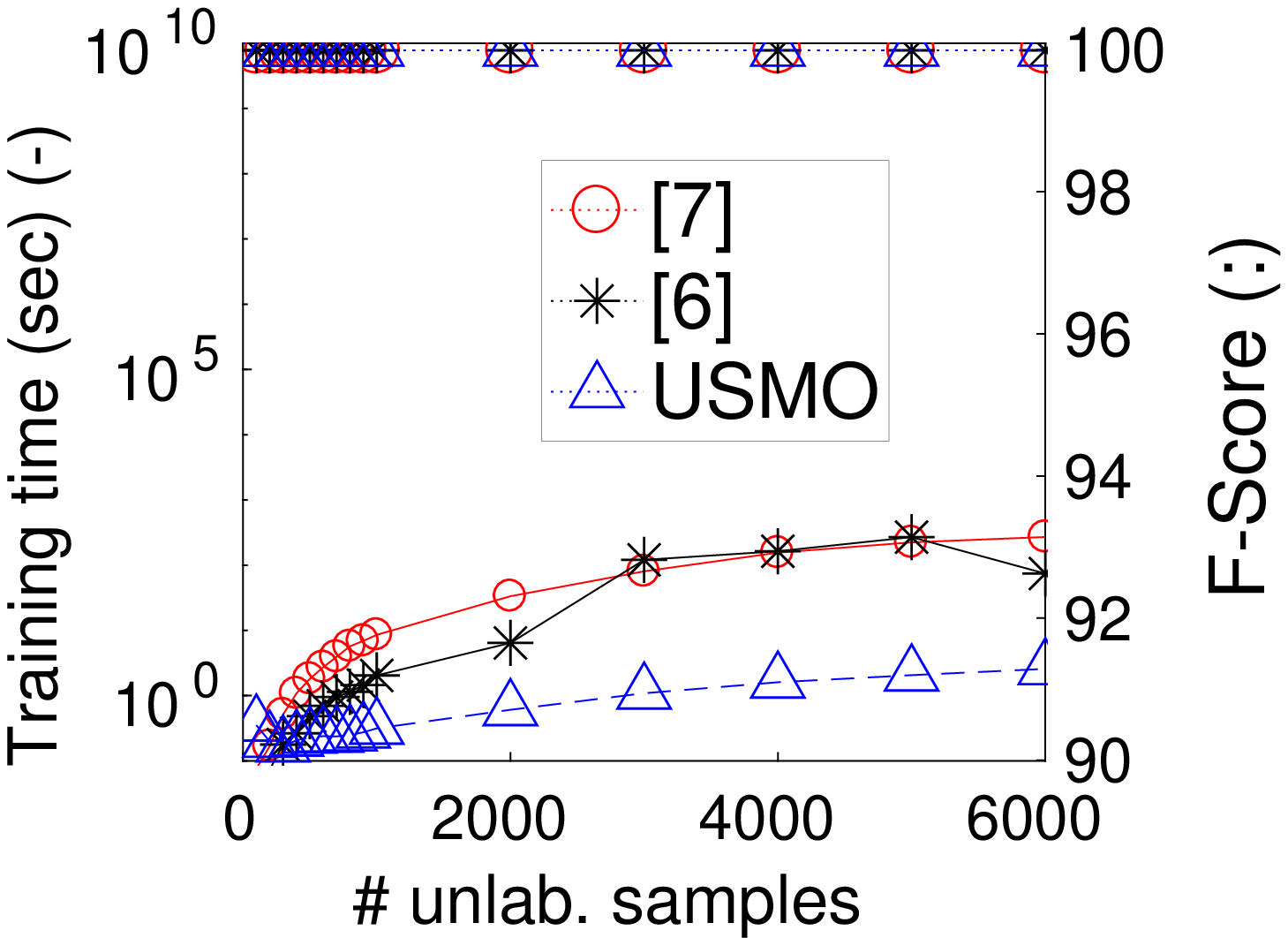}}
\caption{Comparative results on (a) Bank-marketing, (b) Adult and (c)-(l) Poker-hand datasets 
using the Gaussian kernel ($\lambda = 0.01$ and scale parameter equal to 1). Each plot shows 
the training time against different number of 
unlabeled samples ($100$ positive samples) as well as the generalization performance on the test set.}
\label{fig:results_large2_gaussian}
\end{figure*}

Secondly, we investigate \textbf{the complexity of USMO with respect 
to~\cite{plessis2014analysis,plessis2015convex}}. As to the storage requirements, USMO 
behaves linearly instead of quadratically as ~\cite{plessis2014analysis,plessis2015convex}. 
Concerning the computational complexity, it can be easily found
that each iteration has, in the worst case (i.e., an iteration over
the whole unlabeled dataset), a complexity $O(|D_n|)$. As to the
number of iterations, it is difficult to determine a theoretical
limit and it has been experimentally observed over a large and
variate set of tests that it is possible to establish a linear upper
bound with very low slope (less than 40 iterations for 6000
samples). Therefore, we can state that a quadratic dependence
represents a very conservative upper limit for the complexity
of USMO. In particular, 
we measured the processing time of all methods for an increasing number of unlabeled samples 
and with different kernel functions. Figures~\ref{fig:results_large1} and~\ref{fig:results_large2} 
show elapsed time and generalization performance with the linear kernel, 
while Figures~\ref{fig:results_large1_gaussian} and~\ref{fig:results_large2_gaussian} show 
the results achieved with a Gaussian kernel. In most cases, and especially for linear kernel, USMO outperforms all competitors. For Gaussian kernel and few unlabeled samples USMO may require higher computation than ramp loss~\cite{plessis2014analysis}, however, its performance consistently increases with the number of unlabeled samples and its lower storage requirements allow using it also when other methods run out of memory (see results for MNIST in Figure~\ref{fig:results_large1_gaussian}).
%
%
%